\documentclass{article}

\usepackage[final, nonatbib]{neurips_2021}

\usepackage[utf8]{inputenc} 
\usepackage[T1]{fontenc}    
\usepackage{hyperref}       
\usepackage{url}            
\usepackage{booktabs}       
\usepackage{amsfonts}       
\usepackage{nicefrac}       
\usepackage{microtype}      
\usepackage[usenames,dvipsnames]{xcolor}
\usepackage{amsmath, amsthm, amssymb, bm, bbm}

\usepackage{amsmath,amsfonts,bm}

\def\Figref#1{Figure~\ref{#1}}

\def\Secref#1{Section~\ref{#1}}

\def\eqref#1{equation~\ref{#1}}
\def\Eqref#1{Eq.~\ref{#1}}

\def\1{\bm{1}}

\newcommand{\normal}{\mathcal{N}}
\newcommand{\bld}[1]{\boldsymbol{#1}}

\def\rf{{f}}

\def\rvf{{\mathbf{f}}}

\def\rvk{{\mathbf{k}}}

\def\rvx{{\mathbf{x}}}
\def\rvy{{\mathbf{y}}}

\def\rmB{{\mathbf{B}}}

\def\rmK{{\mathbf{K}}}

\def\rmX{{\mathbf{X}}}

\DeclareMathAlphabet{\mathsfit}{\encodingdefault}{\sfdefault}{m}{sl}
\SetMathAlphabet{\mathsfit}{bold}{\encodingdefault}{\sfdefault}{bx}{n}

\def\sR{{\mathbb{R}}}

\newcommand{\E}{\mathbb{E}}

\newcommand{\bK}{\boldsymbol{K}}

\newcommand{\KL}{KL}

\newtheorem{theorem}{Theorem}

\usepackage{multirow}
\usepackage{subfigure}
\usepackage{graphicx}
\usepackage{caption}
\usepackage{soul}
\usepackage{wrapfig}
\usepackage{algorithm}
\usepackage{algpseudocode}

\title{Personalized Federated Learning with \\ Gaussian Processes}

\author{%
  Idan Achituve \\
  Bar-Ilan University, Israel \\
  \texttt{idan.achituve@biu.ac.il}
   \And
   Aviv Shamsian \\
   Bar-Ilan University, Israel \\
   \texttt{aviv.shamsian@biu.ac.il}
   \And
   Aviv Navon \\
   Bar-Ilan University, Israel \\
   \texttt{aviv.navon@biu.ac.il}\\
   \AND
   Gal Chechik \\
   Bar-Ilan University, Israel\\
   NVIDIA, Isreal \\
   \texttt{gal.chechik@biu.ac.il}
   \And
   Ethan Fetaya \\
   Bar-Ilan University, Israel\\
   \texttt{ethan.fetaya@biu.ac.il}
}

\newcommand{\pg}{P\'olya-Gamma }
\newcommand{\K}{\mathbf{K}}
\newcommand{\Knn}{\mathbf{K}_{NN}}
\newcommand{\Kmm}{\mathbf{K}_{MM}}
\newcommand{\Kmn}{\mathbf{K}_{MN}}
\newcommand{\Knm}{\mathbf{K}_{NM}}

\newcommand{\KmmInv}{\mathbf{K}_{MM}^{-1}}
\newcommand{\rvks}{\mathbf{\rvk}^*}
\newcommand{\rvfbar}{\bar{\bld{\mathbf{f}}}}
\newcommand{\rvmu}{\bld{\mathbf{\mu}}}
\newcommand{\rvomega}{\bld{\mathbf{\omega}}}
\newcommand{\rvkappa}{\bld{\mathbf{\kappa}}}
\newcommand{\rmOmega}{\mathbf{\Omega}}
\newcommand{\rmOmegaInv}{\mathbf{\Omega}^{-1}}
\newcommand{\rmLambda}{\mathbf{\Lambda}}
\newcommand{\rmLambdaInv}{\mathbf{\Lambda}^{-1}}
\newcommand{\rmSigma}{\mathbf{\Sigma}}

\newcommand{\NNP}{\mathbf{\theta}}
\newcommand{\rmXbar}{\bar{\mathbf{\mathbf{X}}}}
\newcommand{\rvxbar}{\bar{\mathbf{\mathbf{x}}}}
\newcommand{\rvybar}{\bar{\mathbf{\mathbf{y}}}}
\newcommand{\rmQInv}{\mathbf{B}^{-1}}
\newcommand{\rmKInv}{\mathbf{K}^{-1}}

\newcommand{\tblref}[1]{Table~\ref{#1}}

\begin{document}

\maketitle

\begin{abstract}
Federated learning aims to learn a global model that performs well on client devices with limited cross-client communication. Personalized federated learning (PFL) further extends this setup to handle data heterogeneity between clients by learning personalized models. A key challenge in this setting is to learn effectively across clients even though each client has unique data that is often limited in size. Here we present \textit{pFedGP}, a solution to PFL that is based on Gaussian processes (GPs) with deep kernel learning. GPs are highly expressive models that work well in the low data regime due to their Bayesian nature.
However, applying GPs to PFL raises multiple challenges. Mainly, GPs performance depends heavily on access to a good kernel function, and learning a kernel requires a large training set. Therefore, we propose learning a shared kernel function across all clients, parameterized by a neural network, with a personal GP classifier for each client. We further extend pFedGP to include inducing points using two novel methods, the first helps to improve generalization in the low data regime and the second reduces the computational cost. We derive a PAC-Bayes generalization bound on novel clients and empirically show that it gives non-vacuous guarantees. Extensive experiments on standard PFL benchmarks with CIFAR-10, CIFAR-100, and CINIC-10, and on a new setup of learning under input noise show that pFedGP achieves well-calibrated predictions while significantly outperforming baseline methods, reaching up to 21\% in accuracy gain.

\end{abstract}

\section{Introduction}
In recent years, there is a growing interest in applying learning in decentralized systems under the setup of federated learning (FL) \cite{konevcny2016federated, mcmahan2017communication, shokri2015privacy}. In FL, a server node stores a global model and connects to multiple end-devices (``clients"), which have private data that cannot be shared. The goal is to learn the global model in a communication-efficient manner.  However, learning a single shared model across all clients may perform poorly when the data distribution varies significantly across clients. \textit{Personalized Federated Learning} (PFL) \cite{smith2017federated} addresses this challenge by jointly learning a personalized model for each client. While significant progress had been made in recent years, leading approaches still struggle in realistic scenarios. First, when the amount of data per client is limited, even though this is one of the original motivations behind federated learning \cite{arivazhagan2019federated, mcmahan2017communication, t2020personalized}. Second, when the input distribution shifts between clients, which is often the case, as clients use different devices and sensors. Last, when we require well-calibrated predictions, which is an important demand from medical and other safety-critical applications.  

Here, we show how \textit{Gaussian Processes} (GPs) with deep kernel learning (DKL) \cite{gordon16_DKL} is an effective alternative for handling these challenges. GPs have good predictive performance in a wide range of dataset sizes \cite{achituve2021gp_icml, wilson2016stochastic}, they are robust to input noise \cite{villacampa2021multi}, can adapt to shifts in the data distribution \cite{maddox2021fast}, and provide well-calibrated predictions \cite{snell2020bayesian}. While regression tasks are more natural for GPs, here we focus on classification tasks for consistency with common benchmarks and learning procedures in the field; however, our approach is also applicable to regression tasks.

Consider a naive approach that fits a separate GP classifier to each client based on its personal data. Its performance heavily depends on the quality of the kernel, and standard kernels tend to work poorly in domains such as images. A popular solution to this problem is to use deep kernel learning (DKL) \cite{gordon16_DKL}, where a kernel is applied to features outputted by a neural network (NN). Unfortunately, GPs with DKL can strongly overfit, often even worse than standard NNs \cite{OberRW2021}, and thus negate the main benefit of using a GP. We solve this issue by jointly learning a shared kernel function across clients. As the kernel captures similarities between inputs, a single kernel should work well across clients, while using a separate GP per client will give the required flexibility for personalization.

We adapt a GP classifier recently proposed in \cite{achituve2021gp_icml} which uses the \pg augmentation \cite{polya_gamma} in a tree-structure model to the federated setting. We term our method \textit{pFedGP}. We extend pFedGP by tailoring two inducing points (IPs) methods \cite{quinonero2005unifying, sneldon_Gharamani_IP}. The first helps generalization in the low data regime and, unlike common inducing point methods, does not reduce the computational costs. The second does focus on reducing the computational cost to make our approach scalable and work in low-resource clients. We also adjust previous PAC-Bayes generalization bounds for GPs \cite{reeb2018learning, seeger2002pac} to include the \pg augmentation scheme. These bounds are suitable for cases where the kernel is not learned, such as when new clients arrive after the shared NN was already learned.

Therefore, this paper makes the following contributions: (i) introduce pFedGP as a natural solution to PFL; (ii) develop two IP methods to enhance GP classifiers that use the \pg augmentation scheme and integrate them with pFedGP; (iii) derive a PAC-Bayes generalization bound on novel clients and show empirically that it gives meaningful guarantees; (iv) achieve state-of-the-art results in a wide array of experiments, improving accuracy by up to $21\%$ \footnote{Our code is publicly available at \textcolor{magenta}{\url{https://github.com/IdanAchituve/pFedGP}}}.


\begin{figure*}[!t]
\centering
    \includegraphics[width=0.75\linewidth]{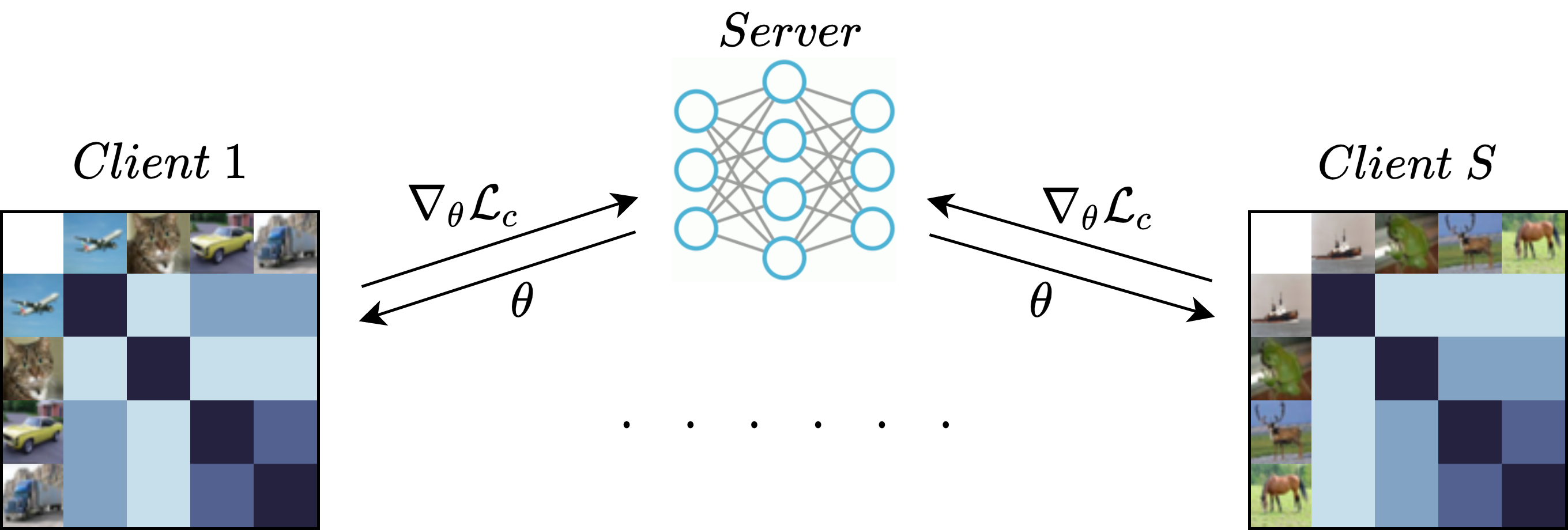}
    \caption{pFedGP - learning a shared deep kernel function with client-specific GP models. Each client stores private data, possibly from a different distribution. The data is first mapped to an embedding space with a shared neural network across all clients. Then, using common kernels a GP is applied to the data of the client for model learning and inference. We illustrate the per-client kernel matrix $k_\theta(\rvx_i,\rvx_j)$. Bold cells indicate a stronger covariance.}
    \label{fig:gp_sys}
\end{figure*}

\section{Related work} \label{sec:related_work}
\textbf{Federated learning.}
In FL, clients collaboratively solve a learning task while preserving data privacy and maintaining communication efficiency~\cite{9220780,kairouz2019advances,li2020federated,mcmahan2017communication,mothukuri2021survey,yang2019federated,ZHANG2021106775}. FedAvg~\cite{mcmahan2017communication} is an early but effective FL approach that updates models locally and averages them into a global model. 
Several optimization methods have been proposed for improving convergence in FL~\cite{li2018convergence,lin2018don,stich2018local,wang2018cooperative}. Other approaches focus on preserving client privacy~\cite{agarwal2018cpsgd,duchi2014privacy, mcmahan2017learning, zhu2019federated}, improving robustness to statistical diversity~\cite{haddadpour2019convergence,hanzely2020federated,Hsu2019MeasuringTE,karimireddy2019scaffold,zhao2018federated,zhou2017convergence}, and reducing communication cost~\cite{dai2019hyper,reisizadeh2020fedpaq}. 
These methods aim to learn a global model across clients, limiting their ability to deal with heterogeneous datasets.

\textbf{Personalized federated learning.}
To overcome client heterogeneity, PFL aims to introduce some personalization for each client in the federation~\cite{Kulkarni2020SurveyOP,tan2021towards}. 
Recent methods include adapting multitask learning~\cite{dinh2021fedu,smith2017federated}, meta-learning approaches~\cite{behl2019alpha,Fallah2020PersonalizedFL,fallah2020convergence, jiang2019improving,li2017meta, zhou2019efficient}, and model mixing, where clients learn a mixture of the global and local models \cite{arivazhagan2019federated,deng2020adaptive,hanzely2020federated,liang2020think}.
Other approaches utilize different regularization schemes to enforce soft parameter sharing~\cite{Huang2020PersonalizedCF, t2020personalized}. 
Personalization in FL has also been explored through clustering approaches in which similar clients within the federation have a greater effect on one another~\cite{Mansour2020ThreeAF,zhang2020personalized}. 
Recently, \cite{shamsian2021personalized_icml} proposed learning a central hypernetworks that acts on client representation vectors for generating personalized models.

\textbf{Bayesian FL.}
Some studies put forward a Bayesian treatment to the FL setup. \cite{bui2018partitioned, corinzia2019variational} used variational inference with Bayesian NNs. \cite{wang2019federated, yurochkin2019bayesian} proposed a matching algorithm between local models based on the Beta-Bernoulli process to construct a global model. \cite{dai2020federated} extended Bayesian optimization to FL setting via Thompson sampling. To scale the GP model they used random Fourier features. We use inducing points instead.
\cite{yin2020fedloc} proposed a federated learning framework that uses a global GP model for regression tasks and without DKL. Unlike this study, we focus on classification tasks with a personal GP classifier per client and advocate sharing information between clients through the kernel. \cite{tang2021fedgp} used GPs in a client selection strategy. In \cite{kassab2020federated} an approach based on stein variational gradient descent was suggested. This method does not scale beyond small-sized networks. \cite{liu2021bayesian} proposed a multivariate Gaussian product mechanism to aggregate local models. As we will show, this method is less suited when the data heterogeneity between clients is large.

\textbf{Gaussian process classification.} 
Unlike regression, in classification approximations must be used since the likelihood is not a Gaussian \cite{gp_book}. Classic approaches include the Laplace approximation \cite{williams1998bayesian}, expectation-propagation \cite{minka2001family}, and least squares \cite{rifkin2004defense}. Recently, several methods were proposed based on the \pg augmentation \cite{polya_gamma} for modeling multinomial distributions \cite{linderman2015dependent}, GP classification \cite{ galy2020multi, galy2020automated, WenzelGDKO19}, few-shot learning \cite{snell2020bayesian}, and incremental learning \cite{achituve2021gp_icml}. Here we build on the last approach. Classification with GPs is commonly done with variational inference techniques \cite{hensman2015scalable}, here we wish to exploit the conjugacy of the model to take Gibbs samples from the posterior. This approach yields well calibrated \cite{snell2020bayesian} and more accurate models \cite{achituve2021gp_icml}.

\section{Gaussian processes background} \label{sec:background}

We first provide a brief introduction to the main components of our model. Detailed explanations are deferred to the Appendix. 
Scalars are denoted with lower-case letters (e.g., $x$), vectors with bold lower-case letters (e.g., $\rvx$), and matrices with bold capital letters (e.g., $\rmX$). In general, $\rvy=[y_1,...,y_N]^T$ is the vector of labels, and $\rmX \in \sR^{N\times d}$ is the design matrix with $N$ data points whose $i^{th}$ row is $\rvx_i$.



\textbf{Gaussian processes.}\label{sec:GPC}
GPs map input points to target output values  via a random latent function $f$. $f$ is assumed to follow a Gaussian process prior  $f\sim\mathcal{GP}(m(\rvx),~k(\rvx,\rvx'))$, where the evaluation vector of $f$ on $\rmX$,  $\rvf=[f(\rvx_1),...,f(\rvx_N)]^T$, has a Gaussian distribution $\rvf\sim\mathcal{N}(\boldsymbol{\mu},~\bK)$ with means $\boldsymbol{\mu}_i=m(\rvx_i)$ and covariance $\bK_{ij}=k(\rvx_i,\rvx_j)$. The mean $m(\rvx)$ is often set to be the constant zero function, and the kernel $k(\rvx,\rvx')$ is a positive semi-definite function. The target values are assumed to be independent when conditioned on $\rvf$. 
For Gaussian process regression the likelihood is Gaussian, $p(y|f)=\mathcal{N}(f,~\sigma^2)$. Therefore, the posterior $p(f|y,\rmX)$ is also Gaussian, and both the marginal and the predictive distributions 
have known analytic expressions. This is one of the main motivations behind using GPs, as most other Bayesian models have intractable inference.

Unfortunately, for Gaussian process classification (GPC) the likelihood, $p(y|f)$, is not a Gaussian and the posterior does not admit a closed-form expression. One approach for applying GPs to binary classification tasks is the \pg augmentation \cite{polya_gamma}. Using this approach, we can augment the GP model with random variables $\rvomega$ from a \pg distribution, one for each example. As a result, $p(\rvf|\rvy,\rmX,\rvomega)$ is a Gaussian density and $p(\rvomega|\rvy,\rmX,\rvf)$ is a \pg density. This allows to use Gibbs sampling to efficiently sample from the posterior $p(\rvf,\rvomega|\rvy,\rmX)$ for inference and prediction. A key advantage of the \pg augmentation is that it benefits from fast mixing and has the ability of even a single value of $\rvomega$ to capture much of the volume of the marginal distribution over function values \cite{linderman2015dependent}. Full equations and further details on the \pg augmentation scheme are given in Appendix \ref{sec_app:pg_back}.

\textbf{Deep kernel learning (DKL).} 
The quality of the GP model heavily depends on the kernel function $k(\rvx_i,\rvx_j)$.
For many data modalities, such as images, common kernels are not a good measure of semantic similarity. Therefore, in \cite{calandra2016manifold, gordon16_DKL} standard kernels are used over features outputted by a neural network $g_\theta$. For example, the RBF kernel $k_\theta(\rvx_i,\rvx_j)=\exp\left(-\frac{||g_\theta(\rvx_i)-g_\theta(\rvx_j)||^2}{2\ell^2}\right)$. 
In regression, it is possible to directly backpropagate through the GP inference as it is given in closed-form. In our case, we use Fisher's identity \cite{douc2014nonlinear} to obtain stochastic gradients \cite{snell2020bayesian}.

\textbf{Inducing points.} GPs require storing and inverting a kernel matrix on the entire training set which often limits its usage. A common solution to this problem is to use inducing point methods \cite{quinonero2005unifying, sneldon_Gharamani_IP}. The key idea is to replace the exact kernel with an approximation for fast computation. Usually, $M \ll N$ pseudo-inputs are learned such that the main computational bottleneck is in inverting $M \times M$ matrices. 
 
\textbf{GP-Tree.} 
We build on GP-Tree \cite{achituve2021gp_icml}, a recent GP classifier that was shown to scale well with dataset size and the number of classes. GP-Tree turns the multi-class classification problem into a sequence of binary decisions along the tree nodes. Each node in the tree fits a binary GP classifier based on the \pg augmentation scheme and the data associated with that node. The leaf nodes correspond to the classes in the dataset. The tree is constructed by first computing a prototype for each class and then recursively performing divisive hierarchical clustering on these prototypes to two clusters at each node. 
Further details are given in Appendix \ref{sec_app:gp_tree}.



\section{pFedGP: federated learning with Gaussian processes} \label{sec:pFedGP}
Now we describe our approach for applying personalized federated learning (PFL) with Gaussian processes.
First, we extend GP-Tree to the FL setup and show how to use Gibbs sampling to learn the NN parameters. Then, we present two alternatives for this method that use inducing points. The first is for extremely limited-size datasets, while the second allows controlling the computational resources. We name our method \textit{pFedGP}. An illustration of our method is given in \Figref{fig:gp_sys}. 

\subsection{A full GP model} \label{sub_sec:full_gp}
The training procedure follows the standard protocol in this field \cite{arivazhagan2019federated, liang2020think, mcmahan2017communication}. We assume the existence of a server that holds the shared parameters $\theta$ (a NN). 
Let $C$ denote the set of clients. For each client $c\in C$ we denote by $D_c$  its local dataset of size $N_c$. At each training iteration (round) the model is sent to $S$ clients to perform kernel learning $(|S| \leq |C|)$. 
Each client $c \in S$ updates its copy of the global model and then sends the updated model to the server. The server then averages over the updates to obtain a new global model. 

At each client $c$, we perform kernel learning in the following manner. We first compute the feature representation of the data samples associated with the client using the shared network. Then, we build the hierarchical classification tree as discussed in \Secref{sec:GPC} \& Appendix \ref{sec_app:gp_tree}. In \cite{achituve2021gp_icml} the tree was built only once after a pre-training stage and the model parameters were learned using a variational inference approach. Here, we re-build the tree at each round using the most recent features and we use a Gibbs sampling procedure, as it allows this flexibility in building the tree and performs better when not prohibitive by computational limitations.
Learning the network parameters $\NNP$ with the Gibbs sampling approach can be done with two common objectives, the marginal likelihood, and the predictive distribution. 

We denote by $\rmX_v$ the data associated with the tree node $v$, i.e., the data points which have $v$ on the path from the root node to their class leaf node. We denote by $\rvy_v$ the binary label of these points, i.e., does their path go left or right at this node. And we denote by $\rvomega_v$ the \pg random variables associated with node $v$. The marginal likelihood term for the full hierarchical classification tree is the  sum of the separate marginal likelihood terms of all the nodes $v$ in the tree:
\begin{equation}\label{eq:tree_objective_ml}
        \mathcal{L}^{ML}_c (\NNP ; D_c) = \sum_{v} \log~p_{\NNP}(\rvy_{v} | \rmX_{v}) = \sum_{v} \log~\int p_{\NNP}(\rvy_{v} | \rvomega_{v}, \rmX_{v}) p(\rvomega_{v}) d\rvomega_{v}.
\end{equation}
Similar to \cite{snell2020bayesian} we use a gradient estimator based on Fisher's identity \cite{douc2014nonlinear}:
\begin{equation}\label{eq:tree_objective_ml_gradients}
    \small
    \begin{aligned}
        \nabla_{\NNP}\mathcal{L}^{ML}_c (\NNP ; D_c) &= \sum_{v } \int p_{\NNP}(\rvomega_{v} | \rvy_{v}, \rmX_{v})\nabla_{\NNP}log~p_{\NNP}(\rvy_{v} | \rvomega_{v}, \rmX_{v})d\rvomega_{v}
        \approx \sum_{v } \frac{1}{L} \sum_{l=1}^{L} \nabla_{\NNP}log~p_{\NNP}(\rvy_{v} | \rvomega_{v}^{(l)}, \rmX_{v}). 
    \end{aligned}
\end{equation}
Here, $\rvomega_{v}^{(1)}, ..., \rvomega_{v}^{(L)}$ are samples from the posterior at node ${v}$. Due to the \pg augmentation $p_{\NNP}(\rvy_{v} | \rvomega_{v}^{(l)}, \rmX_{v})$ is proportional to a Gaussian density. The exact expression is give in Appendix \ref{sec_app:gp_tree}.

To use the predictive distribution as an objective, in each training iteration, after building the tree model, at each node we randomly draw a portion from the (node) training data and use it to predict the class label for the remaining part. We denote with $\rmX_v$ and $\rvy_v$ the training portion, $\rvx^{*}_v$ and $y^{*}_v$ the input and the label of the point we are predicting, and $P^{y^*}$ the path from the root node to the $y^*$ leaf node (i.e., the original class). Here we also take advantage of the independence between nodes to maximize the predictive distribution per node individually. The predictive distribution for a single data point: 
\begin{equation}\label{eq:tree_objective_pl}
    \small
    \begin{aligned}
        \mathcal{L}^{PD}_c (\NNP ;\rvx^*,y^*) &= \sum_{v \in P^{y^*}} \log~p_{\NNP}( y_{v}^{*} | \rvx_{v}^{*}, \rvy_{v},  \rmX_{v})
        = \sum_{v \in P^{y^*}} \log~\int p_{\NNP}( y_{v}^{*} | \rvomega_{v}, \rvx_{v}^{*}, \rvy_{v}, \rmX_{v}) p(\rvomega_{v} | \rvy_{v}, \rmX_{v}) d\rvomega_{v}.
    \end{aligned}
\end{equation}
We use an approximate-gradient estimator based on posterior samples of $\rvomega$:
\begin{equation}\label{eq:tree_objective_pl_gradients}
        \nabla_{\NNP}\mathcal{L}^{PD}_c (\NNP ;\rvx^*,y^*) \approx \sum_{v \in P^{y^*}} \frac{1}{L} \sum_{l=1}^{L} \nabla_{\NNP}log~p_{\NNP}(y_{v}^{*} | \rvomega_{v}^{(l)}, \rvx_{v}^{*}, \rvy_{v}, \rmX_{v}).
\end{equation}
Where $p_{\NNP}(y_{v}^{*} | \rvomega_{v}^{(l)}, \rvx_{v}^{*}, \rvy_{v}, \rmX_{v})=\int p(y_{v}^{*} | f^*) p_{\NNP}(f^* | \rvomega_{v}^{(l)}, \rvx_{v}^{*}, \rvy_{v}, \rmX_{v})df^*$ does not have an analytical expression, 
but $p_{\NNP}(f^* | \rvomega_{v}^{(l)}, \rvx_{v}^{*}, \rvy_{v}, \rmX_{v})=\int p_{\NNP}(f^* |\rvf,\rvx_{v}^{*}, \rmX_{v})p_{\NNP}(\rvf | \rvomega_{v}^{(l)}, \rvy_{v}, \rmX_{v})df^*$ is Gaussian with known parameters. We then compute the predictive distribution by performing Gauss-Hermite integration over $f^*$. See exact expression in Appendix \ref{sec_app:gp_tree}.

\subsection{Augmenting the model with inducing points: sample efficiency} \label{subsec_method:gp_IP-data}
The GP model described in \Secref{sub_sec:full_gp} works well in most situations. However, when the number of data points per client is small, performance naturally degrades. To increase information sharing between clients and improve the per-client performance, we suggest augmenting the model with global inducing points shared across clients. When sending the model from the server to a client, we also send the inducing inputs and their labels. To streamline optimization and reduce the communication burden, we define the inducing inputs in the feature space of the last embedding layer of the shared NN. Therefore, usually, their size will be negligible compared to the network size. 

We denote by $\rmXbar$ the learned inducing inputs and by $\rvybar$ their fixed class labels. They are set evenly across classes. During training, we  regard \textit{only} the set of inducing inputs-labels ($\rmXbar, \rvybar$) as the available (training) data and use them for posterior inference. More formally, we first compute  $p_{\NNP}(\rvf | \bar{\rvomega}, \bar{\rvy}, \bar{\rmX}, \rmX)=\int p_{\NNP}(\rvf | \bar{\rvf}, \bar{\rmX}, \rmX)p_{\NNP}(\bar{\rvf} | \bar{\rvomega}, \bar{\rvy}, \bar{\rmX})d\bar{\rvf}$ using its analytical expression for the actual training data and then compute the probability of $\rvy$ using Gauss-Hermite integration. Then we use Eq. \ref{eq:tree_objective_pl} \& \ref{eq:tree_objective_pl_gradients} for learning the network parameters and the inducing locations.
At test time, to make full use of the training data, we combine the inducing inputs with the training data and use both to obtain the GP formulas and to make predictions. We note that with just using the inducing inputs at test time the model performs remarkably well, despite having almost no personalization component.
See Appendix \ref{sec_app:pFedGP_data_wo_pers} for a further discussion.

One potential issue with using IPs in this manner is that it distorts the true class distribution. As a result, the classifier may be more likely to predict a low-probability class during test time. 
We address this issue by adjusting the output distribution. 
In general, let $p( y, \rvx)$ and $q( y, \rvx)$ be two distributions that differ only in the class probabilities, i.e. $p(\rvx|y) = q(\rvx|y)$, the predictive distribution follows:
\begin{equation}\label{eq:fix_class_dist}
    \begin{aligned}
        \frac{q( y^{*} | \rvx^{*})}{p( y^{*} | \rvx^{*})} &\propto \frac{q( \rvx^{*} | y^{*}) q(y^{*})}{p( \rvx^{*} | y^{*}) p(y^{*})} \Longrightarrow
        q( y^{*} | \rvx^{*}) \propto \frac{q(y^{*})}{p(y^{*})}p( y^{*} | \rvx^{*}).
    \end{aligned}
\end{equation}
We use this to correct the GP predictions to the original class ratios at each tree node. We found in our experiments that this correction generally improves the classifier performance for class imbalanced data. As an example for this phenomena, consider a binary classification problem having $90$ examples from the first class and $10$ examples from the second class (therefore, $q(y=0) = 0.9$, and $q(y=1) = 0.1$). Assume we defined $50$ inducing inputs per class, so now during test time the model sees $140$ samples from the first class and $60$ samples from the second class which corresponds to probabilities $p(y=0)=0.7$ and $p(y=1)=0.3$.



\subsection{Augmenting the model with inducing points: computational efficiency} \label{subsec_method:gp_IP-compute}
Learning the full GP model described in \Secref{sub_sec:full_gp} requires inverting a matrix of size $N_c$ in the worst case (at the root node), which has $\mathcal{O}(N_c^3)$ run-time complexity and $\mathcal{O}(N_c^2)$ memory complexity. Therefore, we propose an additional procedure based on inducing points to allow reduced complexity in low resource environments and scalability to larger dataset sizes.

\begin{algorithm}[!t] 
    \label{algo:pfedgp}
    \caption{\textit{pFedGP}. $C$ clients indexed by c; $E$ - number of local epochs; $|S|$ - number of sampled clients; $M$ - number of inducing inputs per class}
	\vspace{0.1cm}
	{\bf Server executes:}\\
	\hspace*{5mm} Initialize shared network $\theta \leftarrow \theta_0$\\
	\hspace*{5mm} Initialize $M$ inducing inputs per class for all classes in the system  \textcolor{OliveGreen}{\# in pFedGP-IP variants only}\\
	\hspace*{5mm} \textbf{for} each round $t \leftarrow 1, 2, ...$ \textbf{do}:\\
	\hspace*{10mm} Sample $S$ clients uniformly at random\\
	\hspace*{10mm} \textbf{for} each client $c \in S$ \textbf{in parallel}:\\
	\hspace*{15mm} $\theta_{t + 1}^c, M_{t+1}^c \leftarrow $ \textit{ClientUpdate}($\theta_t, M_t$) \textcolor{OliveGreen}{\# obtain updates from client c}\\
	\hspace*{10mm} Update  $\theta_{t + 1}, M_{t+1}$ using FedAvg \cite{mcmahan2017communication} update rule.
	\\~\\
	\textbf{ClientUpdate($\theta, M$):}\\
	\hspace*{5mm} \textbf{for} each local epoch $e \leftarrow 1, ..., E$ \textbf{do}:\\
	\hspace*{10mm} \textbf{if} $e$ = 1:\\
	\hspace*{15mm} Build GP tree classifier using the personal dataset $D_c$\\
	\hspace*{10mm} Update $\theta, M$ using gradient-based optimization methods on $D_c$ with $\mathcal{L}^{ML}$ or $\mathcal{L}^{PD}$\\
	\hspace*{5mm} \textbf{return} $\theta, M$
	
\end{algorithm}

This variant is based on the fully independent training conditional (FITC) method \cite{sneldon_Gharamani_IP}. The key idea is to cast all the dependence on the inducing points and assume independence between the latent function values given the inducing points. Here for brevity, we omit the subscripts denoting the client and the tree node. However, all quantities and data points are those that belong to a specific client and tree node. Let $\rmXbar \in \sR^{M\times d}$ denote the pseudo-inputs (defined in the embedding space of the last layer of the NN), and $\rvfbar \in \sR^{M}$ the corresponding latent function values. Here as well, the inducing inputs are defined globally at the server level and they are set evenly across classes.
We assume the following GP prior $p(\rvf, \rvfbar) = \normal\Big(\bld{0},\Big[\begin{matrix} \Knn & \Knm \\ \Kmn & \Kmm \end{matrix}\Big]\Big)$, where $\Kmm$ is the kernel between the inducing inputs, $\Knn$ is a diagonal matrix between the actual training data, $\Knm$ is the kernel between the data and the inducing inputs, and we placed a zero mean prior. The likelihood of the dataset when factoring the inducing variables and the \pg variables (one per training sample), and the posterior over $\rvf$, both have known analytical expressions. 
We can then obtain the posterior and marginal distributions by marginalizing over $\rvfbar$. Here we will present the posterior and marginal distributions:
\begin{align} \label{eq:IP-compute_posterior_f}
    p(\rvf | \rmX, \rvy, \rvomega, \rmXbar) &= \normal(\rvf | \Knm\rmQInv\Kmn\rmLambdaInv\rmOmegaInv\rvkappa,~\Knn - \Knm(\KmmInv - \rmQInv)\Kmn),\\
    p(\rvy | \rmX, \rvomega, \rmXbar) &\propto \normal(\rmOmegaInv\rvkappa | \bld{0},~\rmLambda + \Knm\KmmInv\Kmn).
\end{align}
Where $\rmOmega = diag(\rvomega)$, $\kappa_j = y_j - 1/2$, $\rmLambda = \rmOmegaInv + diag(\Knn - \Knm\KmmInv\Kmn)$, and $\rmB = \Kmm + \Kmn\rmLambdaInv\Knm$. Importantly, we only need to invert $M\times M$ or diagonal matrices. See full derivation in Appendix \ref{sec_app:pFedGP_comp_derivation}. During test time, we use $\rvfbar$ to get the posterior of $f^*$ to compute the predictive distribution. 

Now we can use either the marginal or the predictive distribution to learn the shared NN parameters and the inducing locations. The complexity of applying this procedure is reduced to $\mathcal{O}(M^2N_c + M^3)$ in run-time, and $\mathcal{O}(MN_c + M^2)$ in memory.
While the (conditional) independence assumption between the latent function values may be restrictive, we found this method to be comparable with the full GP alternative in our experiments. Potentially, this can be attributed to the effect of sharing the inducing inputs among clients and the information that $\rvomega$ stores on $\rvf$. 


\section{Generalization bound} \label{sec:analysis}
It is reasonable to expect that after we learned the system new clients will arrive. In such cases, we would like to use pFedGP without re-training the kernel function. Under this scenario, we can derive generalization bounds concerning only the GP classifier without taking into account the fixed neural network using PAC-Bayes bound \cite{mcallester2003pac}. Having meaningful guarantees can be very important in safety-critical applications. The PAC-Bayes bound for GPC \cite{seeger2002pac}  (with the Gibbs risk):

\begin{theorem}
Given i.i.d. samples $D_c = \{(\rvx_i, y_i)\}_{i=1}^{N_c}$ of size $N_c$ drawn from any data distribution over $\mathcal{X} \times \{-1, 1\}$, a GP posterior $Q$, and a GP prior $P$, the following bound holds, where the probability is over random data samples:
\begin{equation}\label{eq:theorm_gp_pac_bayes}
    Pr_{D_c}\{R(Q) > R_{D_c}(Q) +\KL_{ber}^{-1}(R_{D_c}(Q), \epsilon(\delta, n, P, Q))\} \leq \delta.
\end{equation}
Here, we have,
\begin{equation}
    \small
    \begin{aligned}
    &R(Q) = \E_{(\rvx^*, y^*)}[Pr_{f^* \sim Q(f^* | \rvx^*, D_c)} \{sign~f^* \neq y^*\}],\quad 
    R_{D_c}(Q) = \frac{1}{{N_c}}\sum_{i=1}^{N_c} Pr_{f_i \sim Q(f_i | D_c)} \{sign~f_i \neq y_i\} \\
    &\epsilon(\delta, {N_c}, P, Q) = \frac{1}{{N_c}} \Big(\KL[Q~||~P] + \log~\frac{{N_c}+1}{\delta}\Big),\quad
     \KL^{-1}_{ber}(q, \epsilon) = max_{p \in [0, 1]} \KL_{ber}[q~||~p] \leq \epsilon
    \end{aligned}
\end{equation}
\end{theorem}

An important observation in \cite{seeger2002pac} is that the KL-divergence between the posterior and prior Gaussian processes is equivalent to the KL-divergence between the posterior and prior distribution of their values on the $N_c$ training samples. While \cite{seeger2002pac} assumed $Q$ to be Gaussian, this observation still holds even without this assumption. However, when $Q$ is no longer Gaussian, as is the case here, $\KL[Q(\rvf)~||~P(\rvf)]$ no longer has a closed-form expression. We can show that for the \pg augmentation:

\begin{equation}
\label{eq:kl_GP-PG}
\footnotesize
\begin{aligned}
    \KL[Q(\rvf)~||~P(\rvf)] &= \E_{Q(\rvomega)}\{\KL[Q(\rvf | \rvomega) || P(\rvf)]\} - MI[\rvf;\rvomega] \\
    &=\E_{Q(\rvomega)}\{\KL[Q(\rvf | \rvomega) || P(\rvf)]\}+ \E_{Q(\rvf, \rvomega)}\left[\log~\frac{Q(\rvomega)}{Q( \rvomega|\rvf)}\right]
\end{aligned}
\end{equation}


\begin{wrapfigure}[18]{r}{0.4\textwidth}
  \centering
  \scalebox{.8}{
  \includegraphics[width=0.8\linewidth]{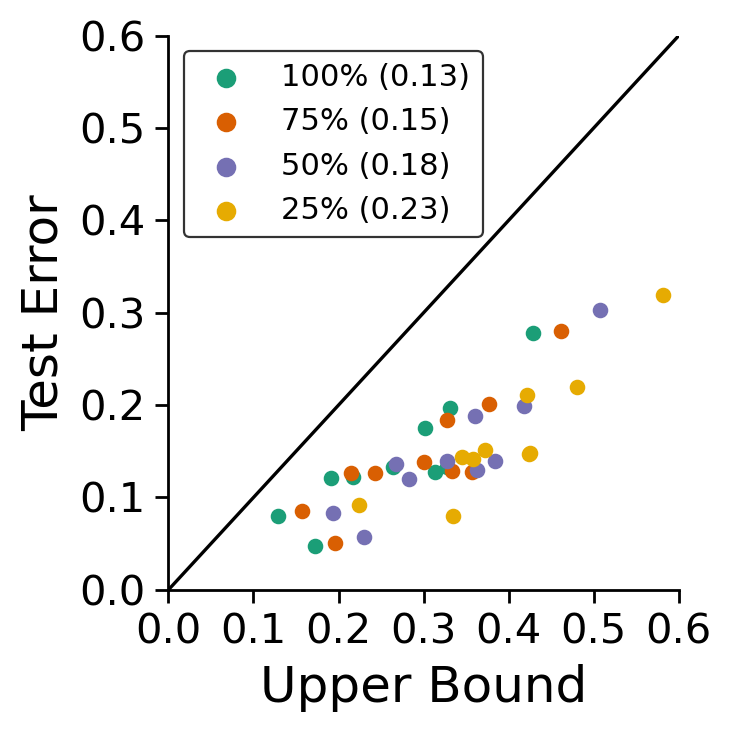}
  }
  \caption{Test error vs an estimated upper bound over 10 clients with varying degrees of a training set data size. Each dot represents a combination of client and data size. 
  In parenthesis - the average difference between the empirical and the test error. }
  \label{fig:gen_vs_test_error}
\end{wrapfigure}

where MI denotes the mutual information. Since $Q(\rvf|\rvomega)$ and $P(\rvf)$ are Gaussian, the  $\KL[Q(\rvf | \rvomega) || P(\rvf)]$ term has a close form expression so we only need to perform Monte-Carlo approximation on the expectation on $\rvomega$ on the first element. In the second expectation, $Q(\rvomega)$ does not have a known expression. To estimate it, given $\{(\rvomega_i,\rvf_i)\}_{i=1}^N$ samples, we use $Q(\rvomega_i)\approx \frac{1}{N-1}\sum_{j\neq i}Q(\rvomega_i|\rvf_j)$. 
Note that if the summation for $j$ includes $\rvf_i$, it might result in a biased estimator. Further details on estimating $\KL[Q(\rvf)~||~P(\rvf)]$ are in Appendix \ref{sec_app:generalization_bound}.

To assess the quality of the bound, we partitioned the CIFAR-10 dataset to 100 clients. We trained a shared network using our full-GP variant on $90$ clients and then recorded the generalization and test error on the remaining $10$ clients four times, each with a different training set size. \Figref{fig:gen_vs_test_error} shows the estimation of the generalization error bound ($\delta = 0.01$) vs the actual error on the novel clients with the Gibbs classifier. First, we observe that indeed the bound is greater than the actual test error for all points and that it is not vacuous. There is a strong correlation between the actual error and the bound. Secondly, unlike worst-case bounds (e.g., VC-dimension), this bound depends on the actual data and not only the number of data points. 
\section{Experiments} \label{sec:experiments}

We evaluated pFedGP against baseline methods in various learning setups. We present the result for the following model variants: \textit{(i) pFedGP}, the full GP model (\Secref{sub_sec:full_gp}); \textit{(ii) pFedGP-IP-data}, the model with IPs described in \Secref{subsec_method:gp_IP-data}; and \textit{(iii) pFedGP-IP-compute}, the model with IPs described in \Secref{subsec_method:gp_IP-compute}. For pFedGP and pFedGP-IP-compute, the results obtained by maximizing the predictive and marginal likelihood were similar, with a slight advantage to the former. Therefore, we present here the results only for the predictive alternative and defer the results of the marginal alternative to the Appendix. Additional experiments, ablation study, and further analyses are provided in Appendix \ref{sec:additional_exp}. Unless stated otherwise, we report the average and the standard error of the mean (SEM) over three random seeds of the federated accuracy, defined as the average accuracy across all clients and samples.

\textbf{Datasets.} All methods were evaluated on CIFAR-10, CIFAR-100 \cite{krizhevsky2009learning}, and CINIC-10 \cite{darlow2018cinic} datasets. CINIC-10 is more diverse since it combines images from CIFAR-10 and ImageNet \cite{deng2009imagenet}. 

\textbf{Compared methods.} We compared our method against the following baselines: \textit{(1) Local}, pFedGP full model on each client with a private network and no collaboration with other clients; \textit{(2) FedAvg} \cite{mcmahan2017communication}, a standard FL model with no personalization component; \textit{(3) FOLA} \cite{liu2021bayesian}, a Bayesian method that used a multivariate Gaussian product mechanism to aggregate local models; \textit{(4) FedPer} \cite{arivazhagan2019federated}, a PFL approach that learns a personal classifier for each client on top of a shared feature extractor; \textit{(5) LG-FedAvg} \cite{liang2020think}, a PFL method that uses local feature extractor per client and global output layers; \textit{(6) pFedMe} \cite{t2020personalized}, a PFL method which adds a
Moreau-envelopes loss term; \textit{(7) FedU} \cite{dinh2021fedu}, a recent multi-task learning approach for PFL that learns a model per client; \textit{(8) pFedHN} \cite{shamsian2021personalized_icml}, a recent PFL approach that uses a hypernetwork to generate client-specific networks.

\textbf{Training protocol.} We follow the training strategy proposed in \cite{shamsian2021personalized_icml}. 
We limit the training process to $1000$ communication rounds, in each we sample five clients uniformly at random for model updates. The training procedure is different in the FOLA and pFedHN baselines, so we used an equivalent communication cost. In LG-FedAvg, we made an extra $200$ communication rounds after a pre-training stage with the FedAvg model for $1000$ communication rounds. In the local model, we performed $100$ epochs of training for each client. In all experiments, we used a LeNet-based network \cite{lecun1998gradient} having two convolution layers followed by two fully connected layers and an additional linear layer. We tuned the hyperparameters of all methods using a pre-allocated held-out validation set. Full experimental details are given in Appendix \ref{sec:exp_details}.


\subsection{Standard PFL setting} \label{sec:hetro}
\setlength{\tabcolsep}{2pt}
\begin{table*}[!t]
\small
\centering
\caption{ Test accuracy ($\pm$ SEM) over 50, 100, 500 clients on CIFAR-10, CIFAR-100, and CINIC-10. The \textit{\# samples/client} indicates the average number of training samples per client.}
\scalebox{.78}{
    \begin{tabular}{l c cc cc cc c cc cc cc c cc cc}
    \toprule
    \multirow{2}{*}{}
    && \multicolumn{5}{c}{CIFAR-10} && \multicolumn{5}{c}{CIFAR-100} && \multicolumn{5}{c}{CINIC-10}\\
    \cmidrule(l){3-7}  \cmidrule(l){9-13} \cmidrule(l){15-19}
    ~\quad\quad \# clients &&50 &&100 &&500 &&50 &&100 &&500 &&50 &&100 &&500 \\
    ~\quad\quad \# samples/client &&800 &&400 &&80 &&800 &&400 &&80 &&1800 &&900 &&180\\
    \midrule
    Local && 86.2 $\pm$ 0.2 && 82.9 $\pm$ 0.4 && 74.8 $\pm$ 0.5 && 52.1 $\pm$ 0.2 && 45.6 $\pm$ 0.3 && 30.9 $\pm$ 0.2 && 61.1 $\pm$ 0.3 && 56.9 $\pm$ 0.7 && 46.4 $\pm$ 0.1\\
    FedAvg \cite{mcmahan2017communication} && 56.4 $\pm$ 0.5 && 59.7 $\pm$ 0.5 && 54.0 $\pm$ 0.5 && 23.6 $\pm$ 0.2 && 24.0 $\pm$ 0.2 && 20.4 $\pm$ 0.0 && 45.6 $\pm$ 0.4 && 44.7 $\pm$ 0.5 && 45.7 $\pm$ 0.5\\
    FOLA \cite{liu2021bayesian} && 55.9 $\pm$ 3.3 && 52.1 $\pm$ 3.1 && 45.9 $\pm$ 0.3 && 25.5 $\pm$ 1.5 && 22.4 $\pm$ 1.3 && 18.7 $\pm$ 0.1 && 45.2 $\pm$ 0.3 && 43.4 $\pm$ 0.3 && 38.3 $\pm$ 0.2 \\
    \midrule
    FedPer \cite{arivazhagan2019federated} && 83.8 $\pm$ 0.8 && 81.5 $\pm$ 0.5 && 76.8 $\pm$ 1.2 && 48.3 $\pm$ 0.6 && 43.6 $\pm$ 0.2 && 25.6 $\pm$ 0.3 && 70.6 $\pm$ 0.2 && 68.4 $\pm$ 0.5 && 62.2 $\pm$ .05\\
    LG-FedAvg \cite{liang2020think} && 87.9 $\pm$ 0.3 && 83.6 $\pm$ 0.7 && 64.7 $\pm$ 0.7 && 43.6 $\pm$ 0.2 && 37.5 $\pm$ 0.9 && 20.3 $\pm$ 0.5 && 59.5 $\pm$ 1.1 && 59.9 $\pm$ 2.1 && 52.5 $\pm$ 0.8\\
    pFedMe \cite{t2020personalized} && 86.4 $\pm$ 0.8 && 85.0 $\pm$ 0.3 && 80.3 $\pm$ 0.5 && 49.8 $\pm$ 0.5 && 47.7 $\pm$ 0.4 && 32.5 $\pm$ 0.8 && 69.9 $\pm$ 0.5 && 68.9 $\pm$ 0.7 && 58.8 $\pm$ 0.1 \\
    FedU \cite{dinh2021fedu} && 80.6 $\pm$ 0.3 && 78.1 $\pm$ 0.5 && 65.6 $\pm$ 0.4 && 41.1 $\pm$ 0.2 && 36.0 $\pm$ 0.2 && 15.9 $\pm$ 0.4 && 59.3 $\pm$ 0.2 && 55.4 $\pm$ 0.6 && 41.6 $\pm$ 0.5 \\
    pFedHN \cite{shamsian2021personalized_icml} && \textbf{90.2 $\pm$ 0.6} && 87.4 $\pm$ 0.2 && 83.2 $\pm$ 0.8 && 60.0 $\pm$ 1.0 && 52.3 $\pm$ 0.5 && 34.1 $\pm$ 0.1 && 70.4 $\pm$ 0.4 && 69.4 $\pm$ 0.5 && 64.2 $\pm$ .05 \\
    \midrule
    \textbf{Ours} && && && && && && && && &&\\
    pFedGP-IP-data && 88.6 $\pm$ 0.2 && 87.4 $\pm$ 0.2 && 86.9 $\pm$ 0.7 && 60.2 $\pm$ 0.3 && 58.5 $\pm$ 0.3 && \textbf{55.7 $\pm$ 0.4} && 69.8 $\pm$ 0.2 && 68.3 $\pm$ 0.6 && 67.6 $\pm$ 0.3 \\
    pFedGP-IP-compute && 89.9 $\pm$ 0.6 && \textbf{88.8 $\pm$ 0.1} && 86.8 $\pm$ 0.4 && 61.2 $\pm$ 0.4 && 59.8 $\pm$ 0.3 && 49.2 $\pm$ 0.3 && \textbf{72.0 $\pm$ 0.3} && \textbf{71.5 $\pm$ 0.5} && \textbf{68.2 $\pm$ 0.2} \\
    pFedGP &&  89.2 $\pm$ 0.3 && \textbf{88.8 $\pm$ 0.2} && \textbf{87.6 $\pm$ 0.4} && \textbf{63.3 $\pm$ 0.1} && \textbf{61.3 $\pm$ 0.2} && 50.6 $\pm$ 0.2 && \textbf{71.8 $\pm$ 0.3} && \textbf{71.3 $\pm$ 0.4} && \textbf{68.1 $\pm$ 0.3} \\
    \bottomrule
    \end{tabular}
}
\label{tab:1}
\end{table*}
We first evaluated all methods in a standard PFL setting~\cite{shamsian2021personalized_icml, t2020personalized}.
We varied the total number of clients in the system from 50 to 500 and we set the number of classes per client to two/ten/four for CIFAR-10/CIFAR-100/CINIC-10 respectively. Since the total number of samples in the system is fixed, the number of samples per client changed accordingly.
For each client, the same classes appeared in the training and test set.  

The results are presented in \tblref{tab:1}. They show that: (1) The performance of the \textit{local} baseline is significantly impaired when the number of samples per client decreases, emphasizing the importance of federated learning in the presence of limited local data. (2) FedAvg and FOLA, which do not use personalized FL, perform poorly in this heterogeneous setup. (3) pFedGP outperforms or is on par with previous state-of-the-art approaches when local data is sufficient (e.g., 50 clients on all datasets). When the data per client becomes limited, pFedGP achieves significant improvements over competing methods; note the $9\%$ and $21\%$ difference in CIFAR-100 over $100$ and $500$ clients, respectively. (4) pFedGP-IP-compute often achieves comparable results to pFedGP and is often superior to pFedGP-IP-data. We believe that it can be attributed to the fact that in pFedGP-IP-compute the training data take an active part in the GP inference formulas (\Eqref{eq:IP-compute_posterior_f}), while in pFedGP-IP-data the data impact in a weak manner only through the loss function. (5) pFedGP-IP-data is especially helpful when few samples per class are available, e.g., CIFAR-100 with 500 clients. That last point is further illustrated by decoupling the effect of the \textit{number of clients} from that of the \textit{training set size}. To illustrate that, in Appendix \ref{sec:varying_training_size} we fixed the number of clients and varied the number of training samples per class. From this experiment, we deduced that both factors (individually) contribute to pFedGP success.

\begin{wraptable}[16]{r}{0.5\linewidth}
\vspace{-13pt}
\setlength{\tabcolsep}{3pt}
\small
\caption{Test accuracy ($\pm$ SEM) over 100 clients on noisy CIFAR-100. We also provide the relative accuracy decrease (\%) w.r.t. the performance on the original CIFAR-100 data (see Table~\ref{tab:1}).
}
\centering
\begin{tabular}{l c c c} 
    \toprule
    \multicolumn{1}{l}{Method} && \multicolumn{1}{c}{Accuracy} & Decrease (\%) \\
    \midrule
    FedPer \cite{arivazhagan2019federated} && 28.1 $\pm$ 0.9 & -35.6 \\
    LG-FedAvg \cite{liang2020think} && 26.9 $\pm$ 0.9 & -28.3 \\
    pFedme \cite{t2020personalized} && 33.2 $\pm$ 0.6 & -30.4 \\
    FedU  \cite{dinh2021fedu} && 35.0 $\pm$ 0.2 & \textbf{-2.8} \\
    pFedHN \cite{shamsian2021personalized_icml} && 38.9 $\pm$ 0.5 & -25.7 \\
    \midrule
    \textbf{Ours} &&\\
    pFedGP-IP-data && 45.0 $\pm$ 0.3 & -23.1 \\
    pFedGP-IP-compute && 47.1 $\pm$ .05 & -21.2 \\
    pFedGP && \textbf{49.5 $\pm$ 0.1} & -19.2 \\
    \bottomrule
\end{tabular}
\label{tab:noisy_cifar}
\end{wraptable}

\begin{figure}[!t]
\centering
\tiny
\begin{tabular}{c c c c c}
    ~~~~FedAvg & ~~~~~~FOLA & ~~~~FedPer & ~~~~LG-FedAvg & ~~~~pFedMe \\
    \includegraphics[width=25mm]{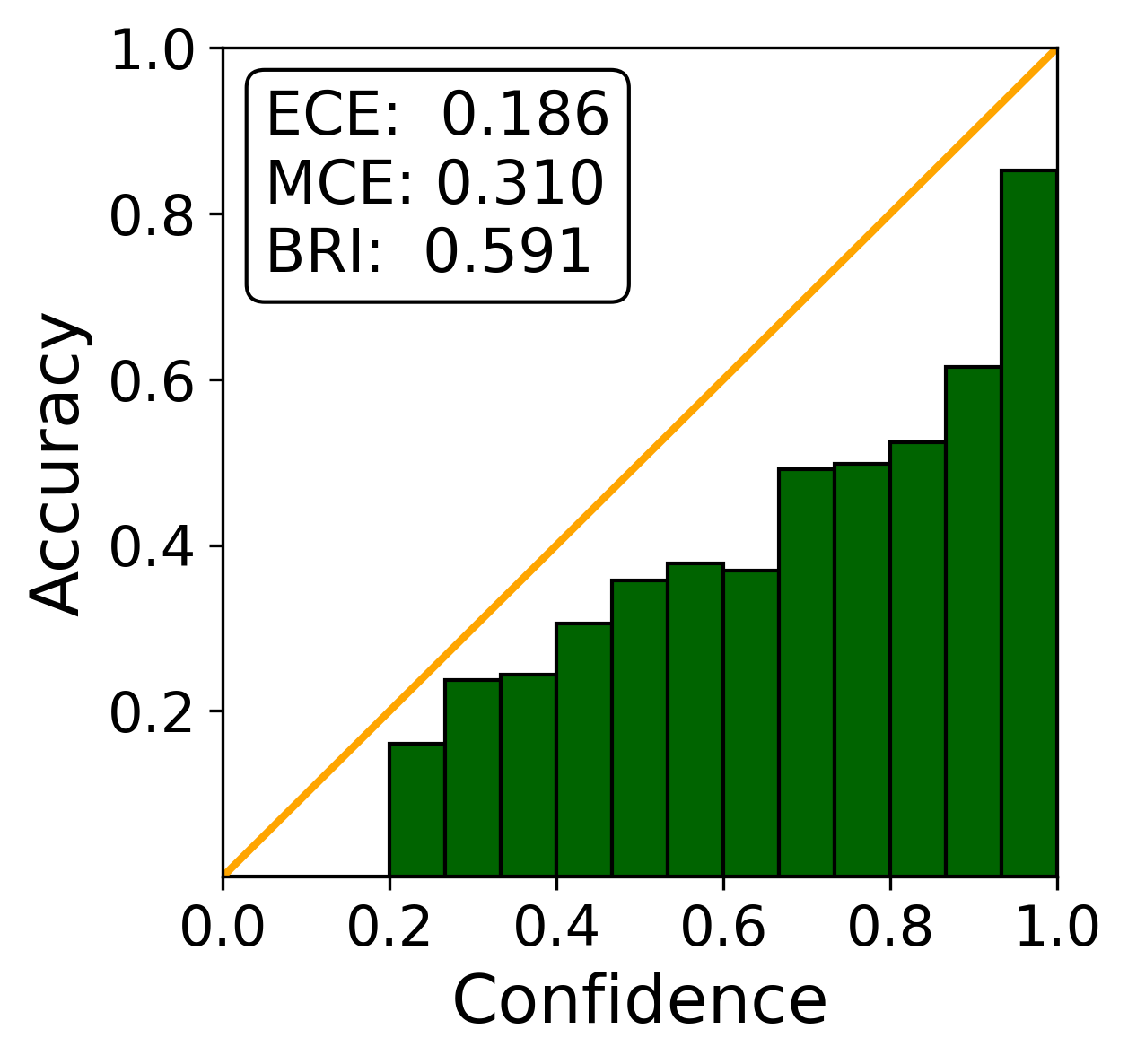} &   \includegraphics[width=25mm]{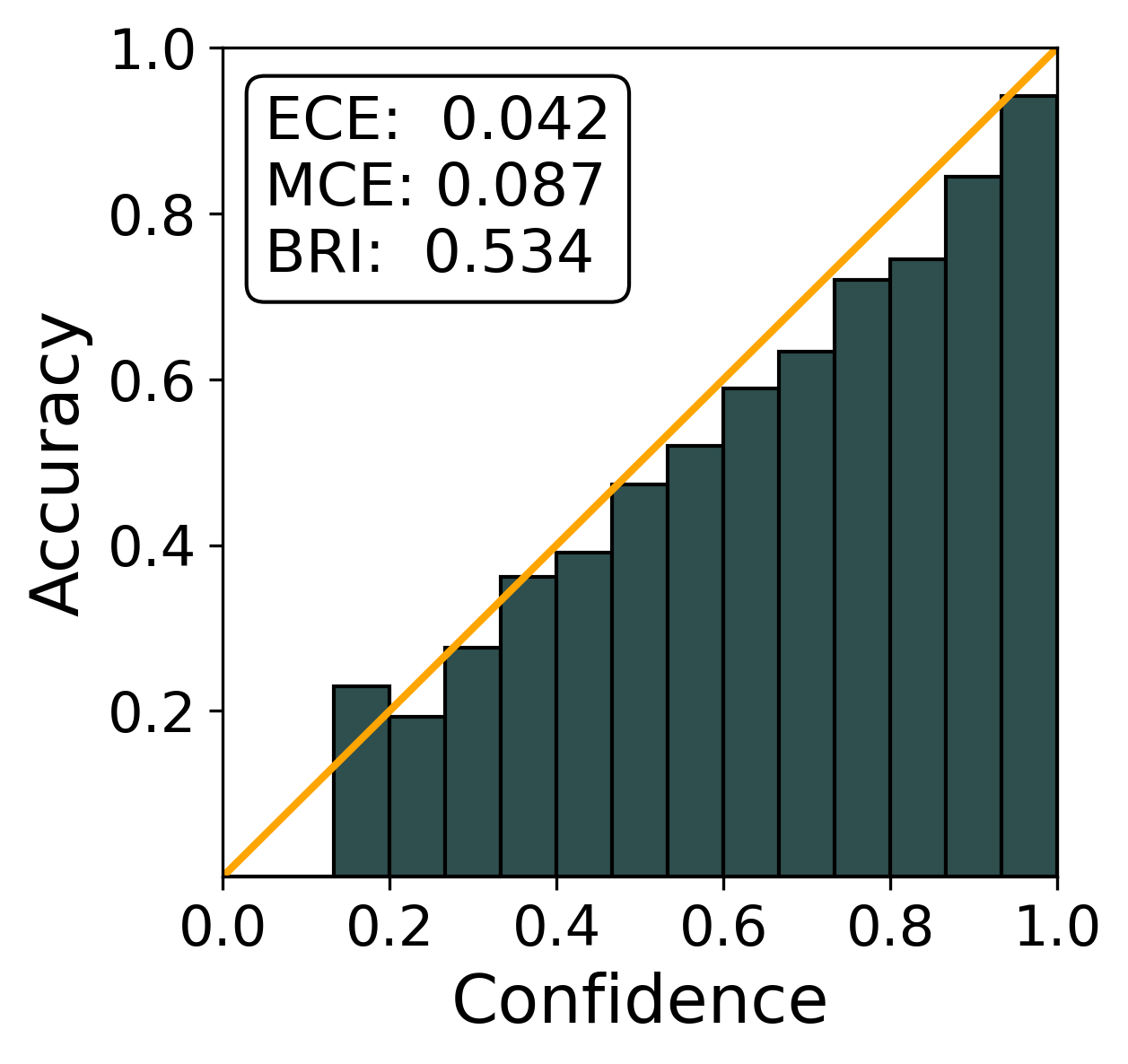} & \includegraphics[width=25mm]{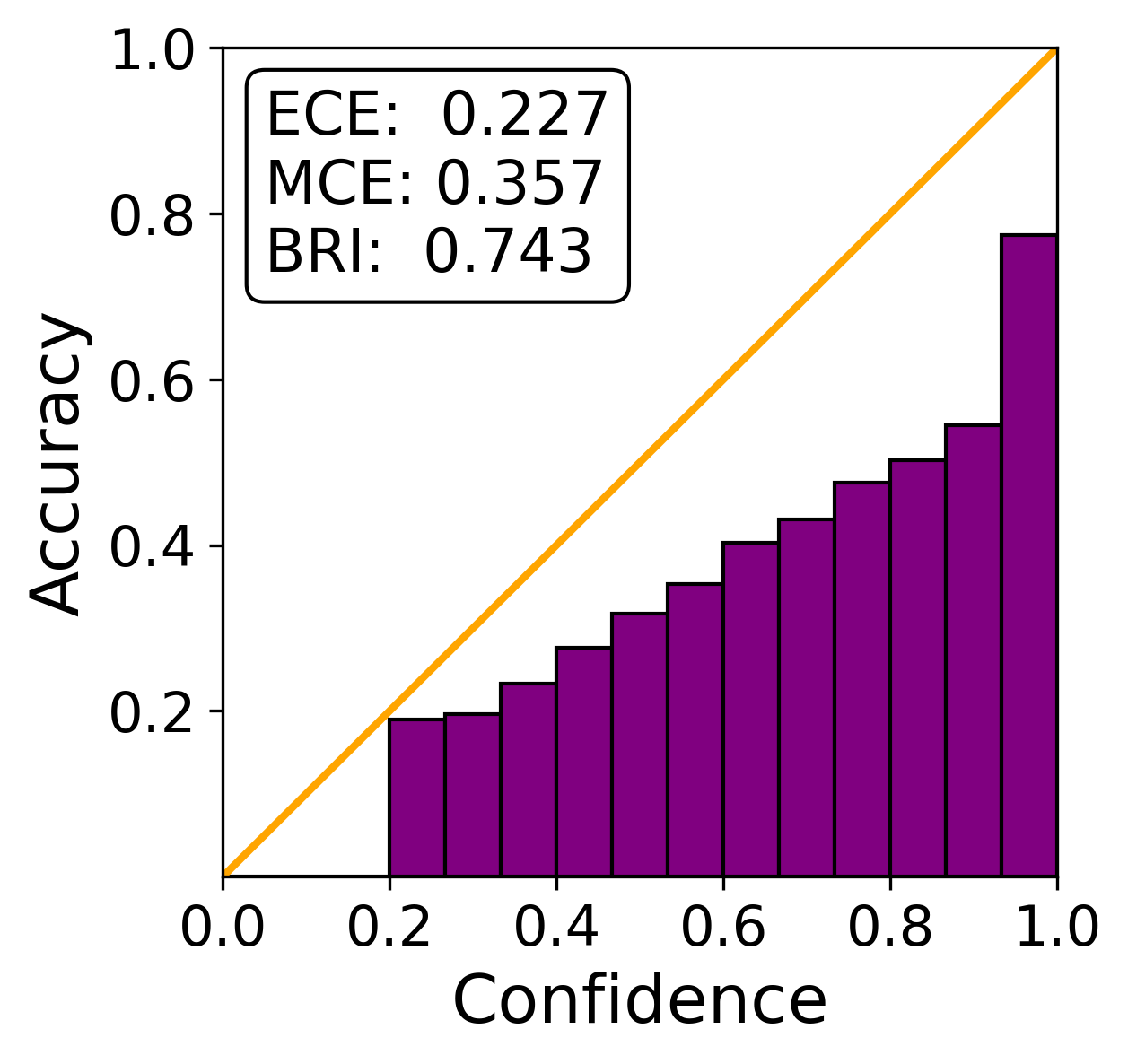} &
    \includegraphics[width=25mm]{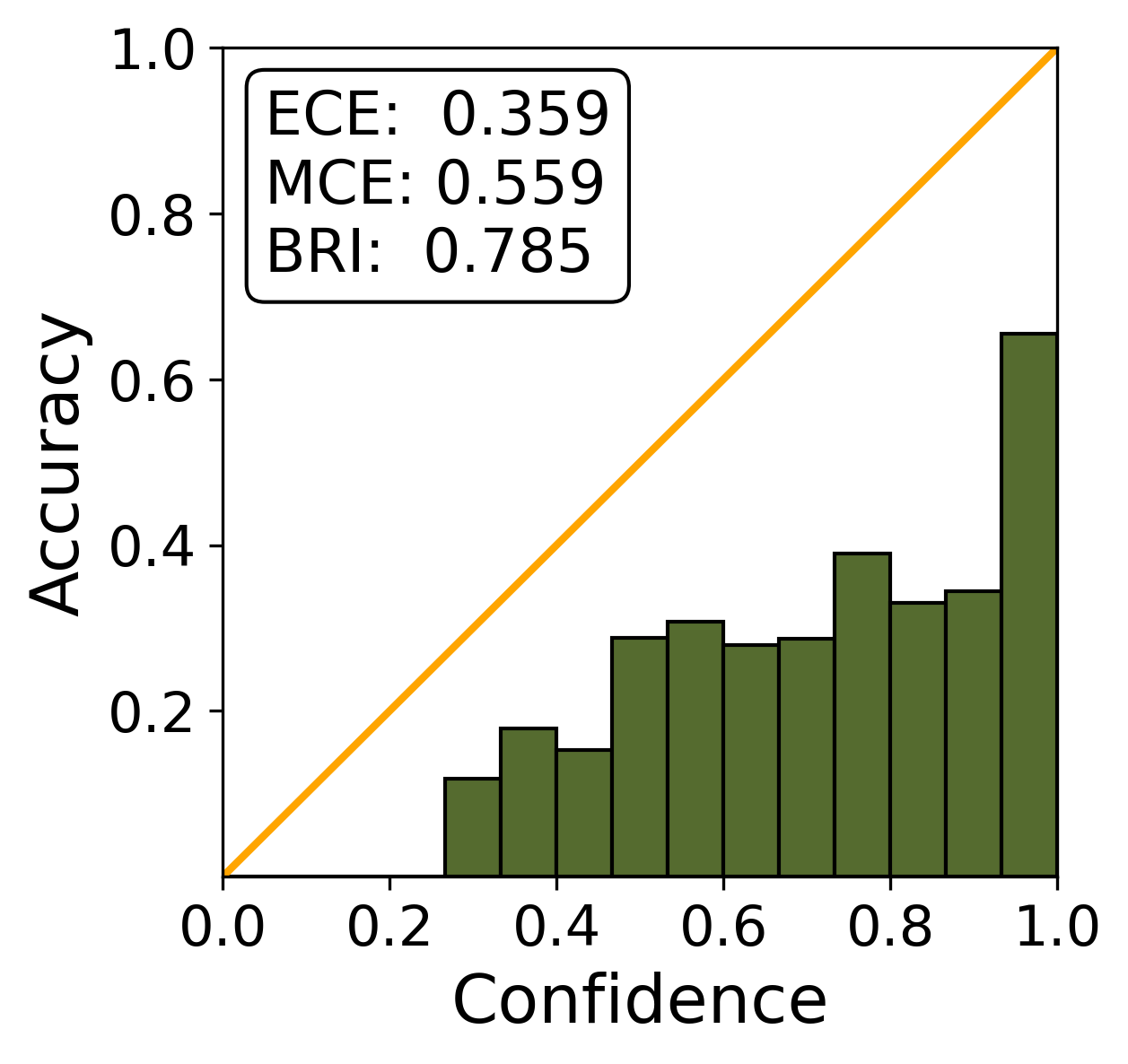} &
    \includegraphics[width=25mm]{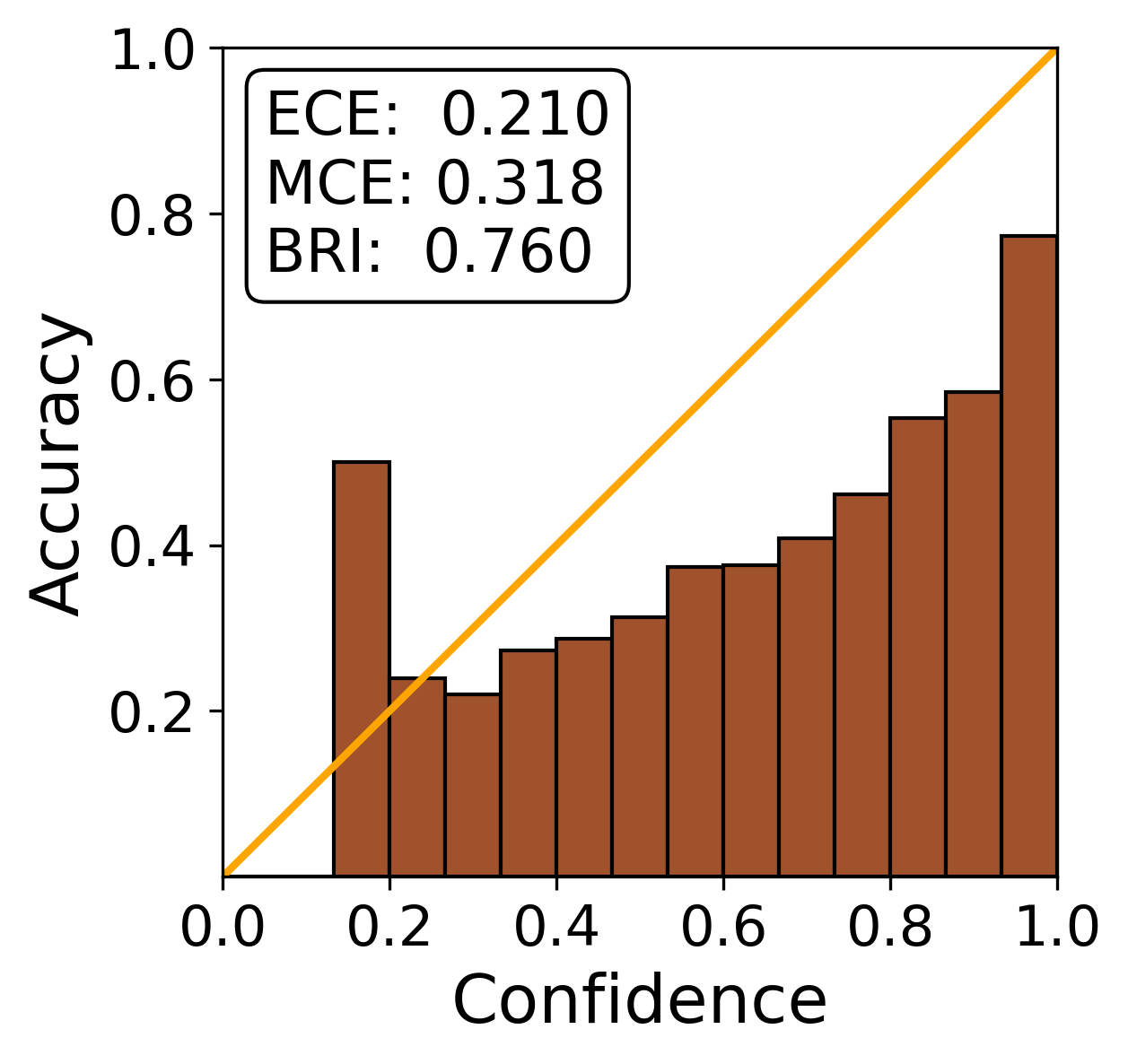}
    \\
    ~~~~~~FedU & ~~~~pFedHN & ~~~~pFedGP-IP-data (ours) & ~~~~~pFedGP-IP-compute (ours) & ~~~~pFedGP (ours)\\
    \includegraphics[width=25mm]{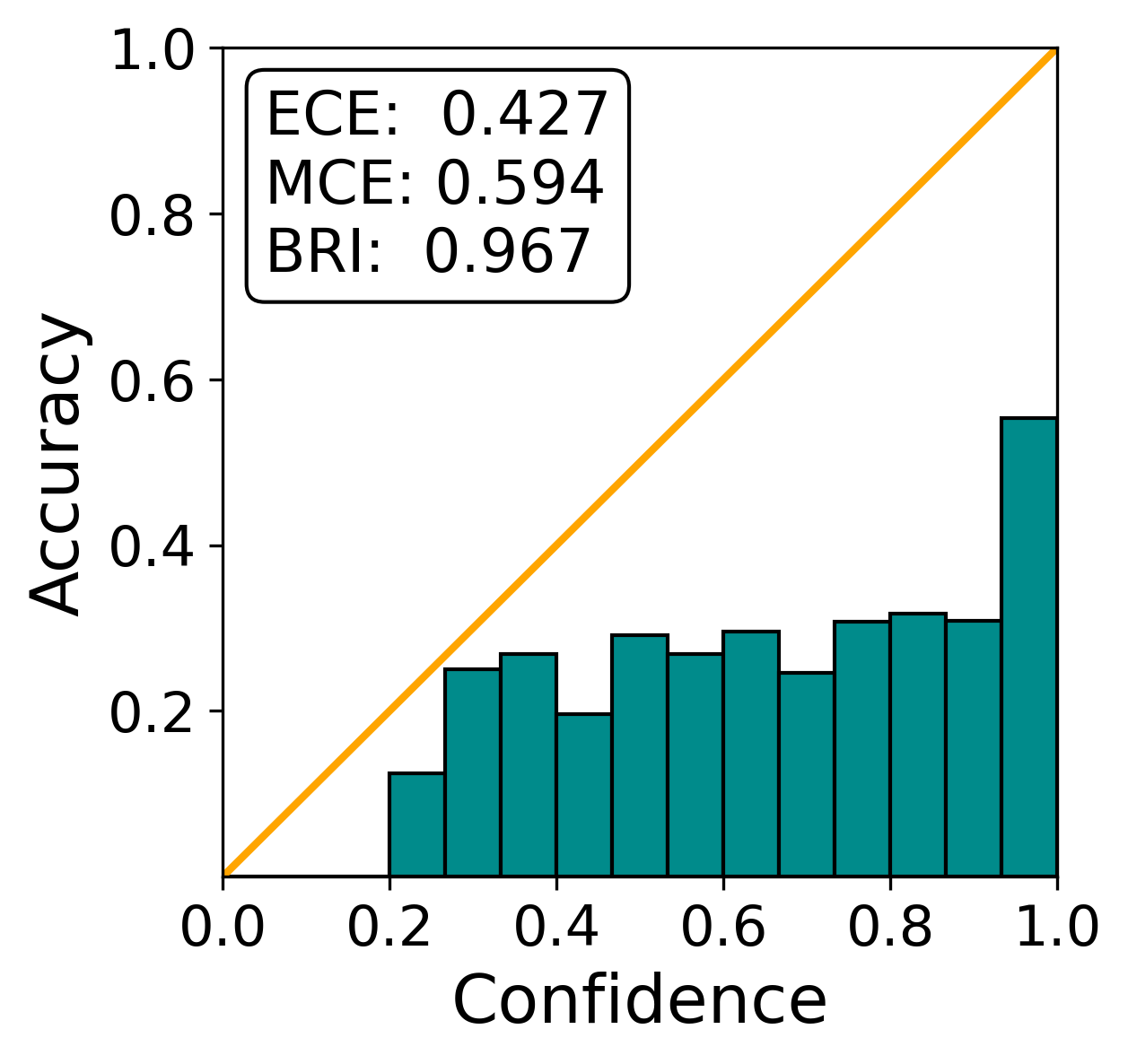} &
    \includegraphics[width=25mm]{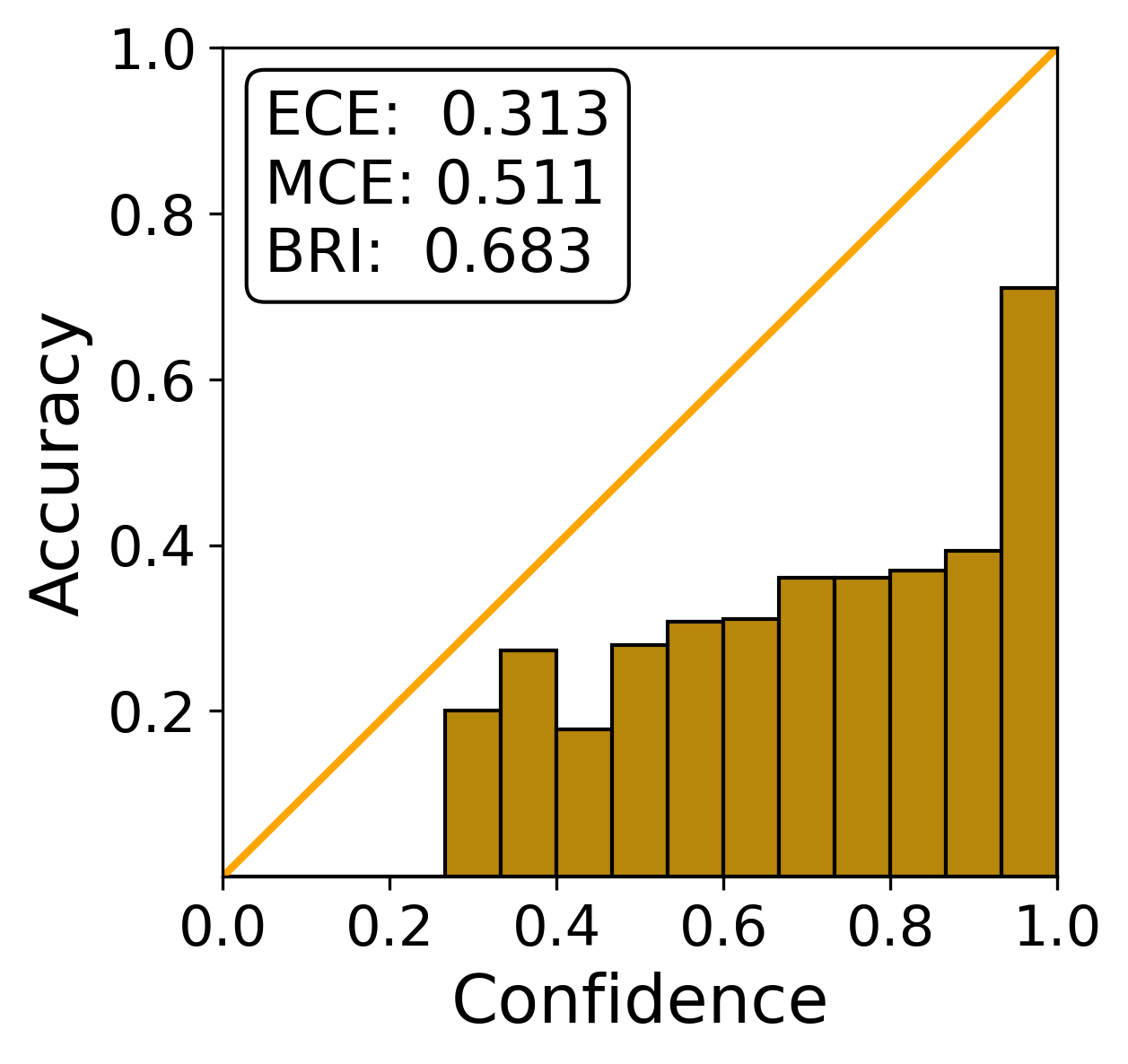} &
    \includegraphics[width=25mm]{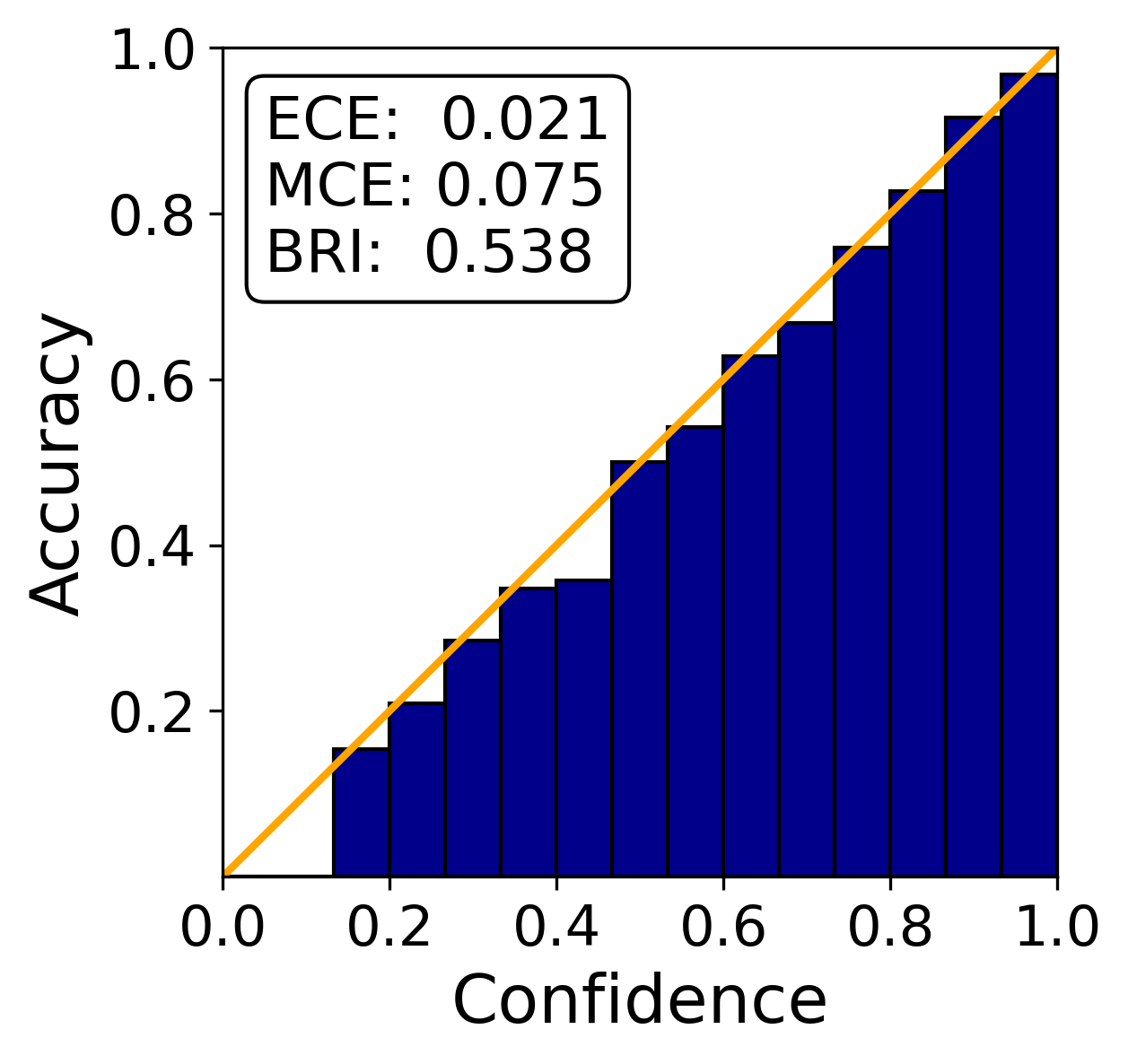} &
    \includegraphics[width=25mm]{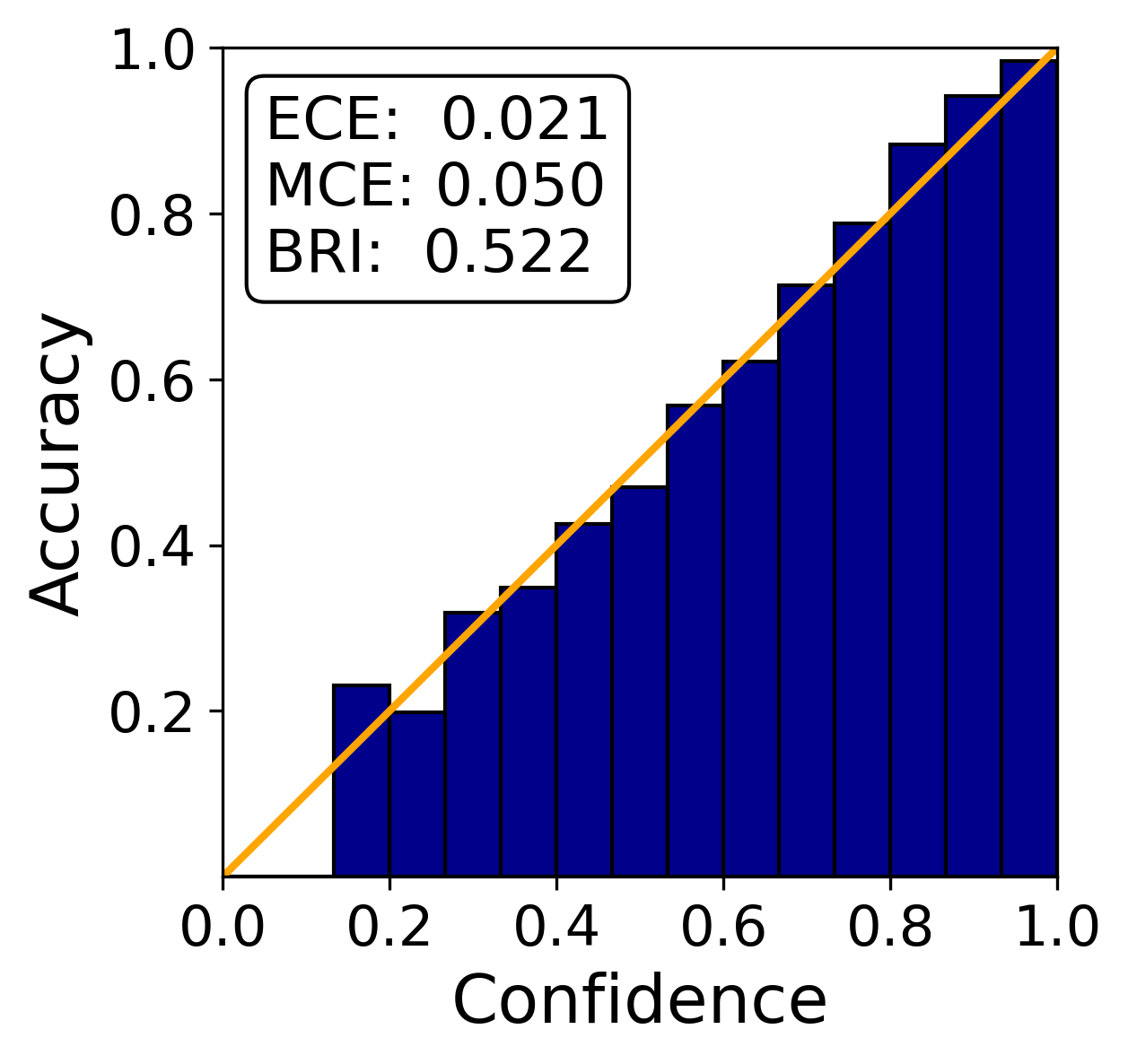} &
    \includegraphics[width=25mm]{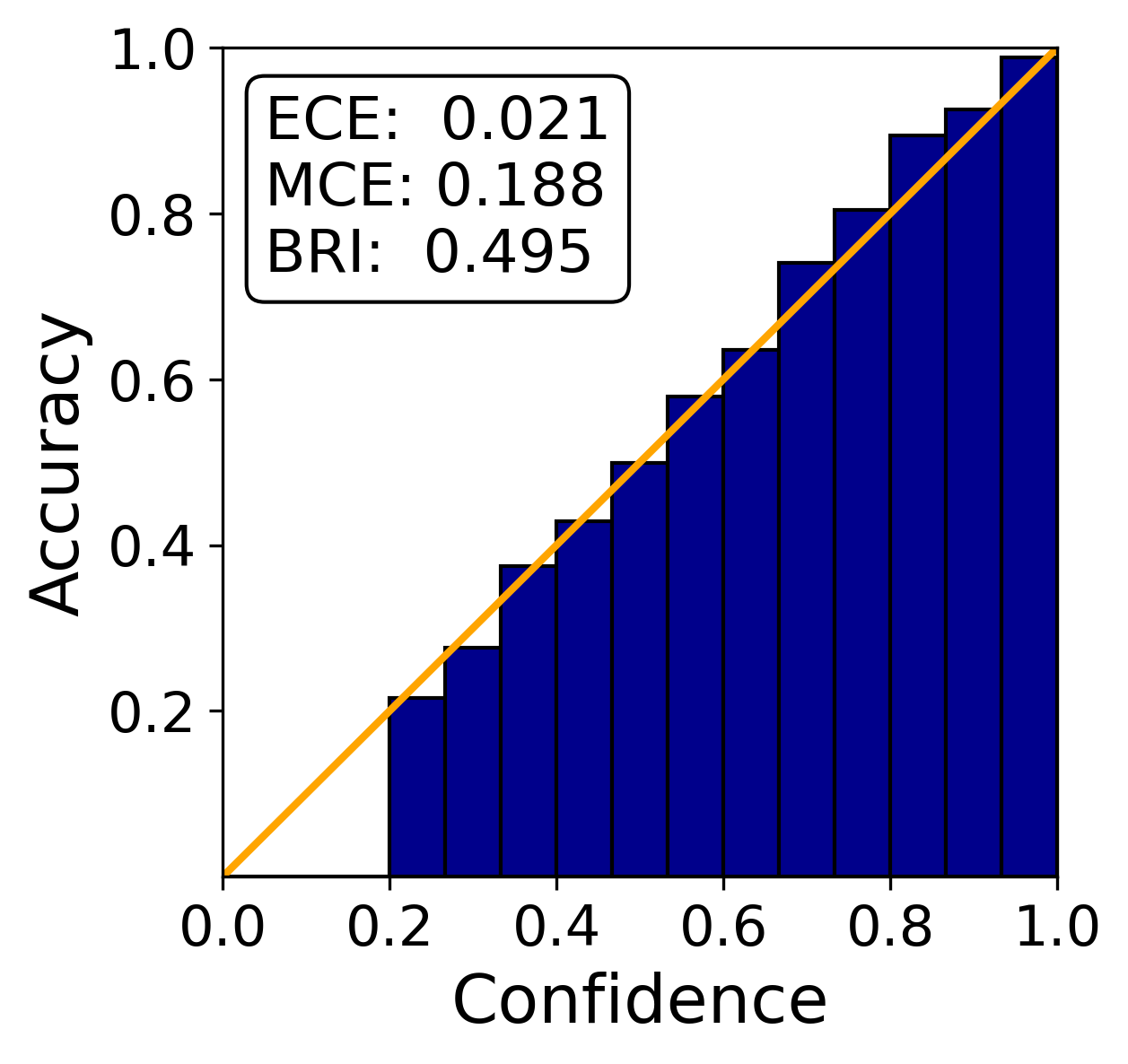}
    \end{tabular}
    \caption{Reliability diagrams on CIFAR-100 with 50 clients. Diagonal indicates perfect calibration. Each plot also shows the expected \& maximum calibration error (ECE \& MCE) and the Brier Score (BRI). Lower is better.}
    \label{fig:calibration_50_temp1}
\end{figure}

A desired property from PFL classifiers is the ability to provide uncertainty estimation. For example, in decision support systems, such as in healthcare applications, the decision-maker should have an accurate estimation of the classifier confidence in the prediction.
Here, we quantify the uncertainty through calibration. \Figref{fig:calibration_50_temp1} compares all methods both visually and using common metrics \cite{brier1950verification,guo2017calibration, naeini2015obtaining} on the CIFAR-100 dataset with 50 clients. Expected calibration error (ECE) measures the weighted average between the classifier confidence and accuracy. Maximum calibration error (MCE) takes the maximum instead of the average. And, Brier score (BRI) \cite{brier1950verification} measures the average squared error between the labels and the prediction probabilities. The figure shows that pFedGP classifiers are best calibrated across all metrics in almost all cases. We note that with temperature scaling, the calibration of the baseline methods can be improved \cite{guo2017calibration}; however, choosing the right temperature requires optimization over a separate validation set, which our model does not need. Additional calibration results, including temperature scaling, are presented in Appendix \ref{sec:reliability_diad_appendix}.

\subsection{PFL with input noise} \label{sec:pfl_in_noise}

In real-world federated systems, the clients may employ different measurement devices for data collection (cameras, sensors, etc.), resulting in different input noise characteristics per client.
Here, we investigate pFedGP performance in this type of personalization. To simulate that, we partitioned CIFAR-10/100 to 100 clients similar to the protocol described in Section~\ref{sec:hetro}, we defined 57 unique distributions of image corruption noise \cite{hendrycks2018benchmarking}, and we assigned a noise model to each client. Then for each example in each client, we sampled a corruption noise according to the noise model allocated to that client. Here we show the results for the noisy CIFAR-100 dataset in \tblref{tab:noisy_cifar}.
Further details on the perturbations performed and result for the noisy CIFAR-10 are given in the Appendix\footnote{The noisy CIFAR-10/100 datasets are available at: \textcolor{magenta}{\href{https://idanachituve.github.io/projects/pFedGP/}{https://idanachituve.github.io/projects/pFedGP}}}. We observe a significant gap in favor of the pFedGP variants compared to baseline methods. 
Note that using global inducing points is slightly less beneficial in this case since they are defined globally and therefore are not tied to a specific noise type as the real client data is. 

\subsection{Generalization to out-of-distribution (OOD) novel clients} \label{sec:exp_gen_ood_clients}

\begin{wrapfigure}[14]{r}{0.5\textwidth}
    \vspace{-10pt}
    \centering
    \includegraphics[width=.9\linewidth]{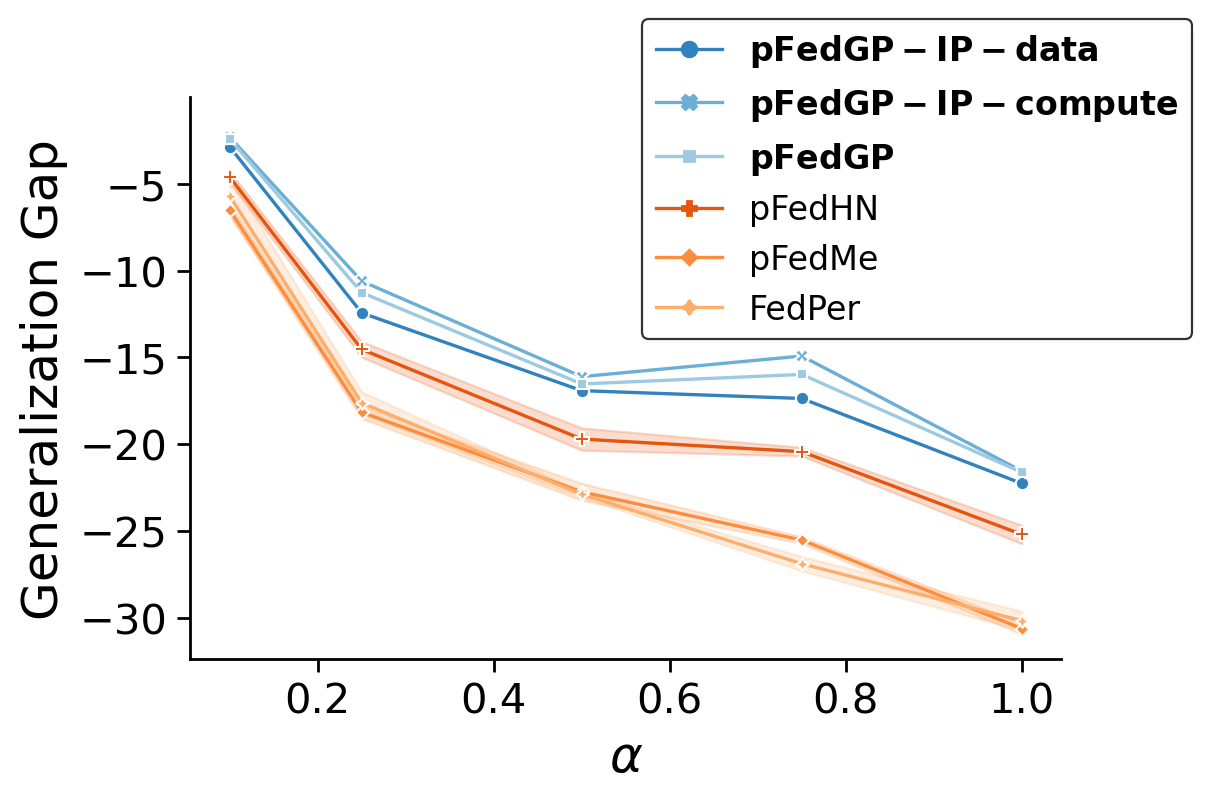}
    \caption{Generalization to novel clients on CIFAR-10.}
    \label{fig:gen}
\end{wrapfigure}

FL are dynamic systems. For example, novel clients may enter the system after the model was trained, possibly with a data distribution shift. Adapting to a new OOD client is both challenging and important for real-world FL systems. To evaluate pFedGP in this scenario, we followed the learning protocol proposed in~\cite{shamsian2021personalized_icml}. We partitioned the CIFAR-10 dataset into two groups. The data in the first group was distributed between 90 clients for model training. The remaining data from the second group was distributed between an additional 10 clients that were excluded during training.
Within each group, we set the class probabilities in each client by sampling from a Dirichlet distribution with the same $\alpha$ parameter. 
For the training group, we set $\alpha=0.1$, trained the shared model using these clients, and froze it. Then, we evaluated the models on the second group by varying $\alpha\in\{.1, .25, .5, .75, 1\}$, 
on the remaining 10 clients. As $\alpha$ moves away from $0.1$ the distribution shift between the two groups increases, resulting in more challenging OOD clients. 
Figure~\ref{fig:gen} reports the generalization gap as a function of the Dirichlet parameter $\alpha$. The generalization gap is computed by taking the difference between the average test accuracy of the (ten) novel clients and the average test accuracy of the (ninety) clients used for training. From the figure, here as well, pFedGP achieves the best generalization performance for all values of $\alpha$. Moreover, unlike baseline methods, pFedGP does not require \textit{any} parameter tuning. Several baselines were excluded from the figure since they had a large generalization gap. 

\vspace{-5pt}
\section{Conclusion} \label{sec:discussion}
\vspace{-10pt}
In this study, we proposed pFedGP, a novel method for PFL. pFedGP learns a kernel function, parameterized by a NN, that is shared between all clients using a personal GP classifier on each client. We proposed three variants for pFedGP, a full model approach that generally shows the best performance and two extensions to it. The first is most beneficial when the number of examples per class are small while the second allows controlling the computational requirements of the model. We also derived PAC-Bayes generalization bound on novel clients and empirically showed that it gives non-vacuous guarantees. pFedGP provides well-calibrated predictions, generalizes well to OOD novel clients, and consistently outperforms competing methods.

\textbf{Broader impact:} Our method shares the standard communication procedure of FL approaches, where no private data is directly communicated across the different nodes in the system. This protocol does not explicitly guarantee that no private information can be inferred at this time. As we show, pFedGP is particularly useful for clients with little data, and for clients that have strongly different distribution. This has great potential to improve client personalization in real-world systems, and do better at handling less common data. The latter is of great interest for decision support systems in sensitive domains such as health care or legal.

\section*{Acknowledgements} 
This study was funded by a grant to GC from the Israel Science Foundation (ISF 737/2018), and by an equipment grant to GC and Bar-Ilan University from the Israel Science Foundation (ISF 2332/18). IA  was funded by a grant from the Israeli innovation authority, through the AVATAR consortium.


\bibliography{Ref}
\bibliographystyle{plain}

\clearpage

\title{Supplementary Material for Personalized Federated Learning With Gaussian Processes}

\maketitle
\appendix

\section{Extended background}
\subsection{The \pg augmentation} \label{sec_app:pg_back}
A random variable $\omega$ has a \pg distribution if it can be written as an infinite sum of independent gamma random variables:
\begin{equation} \label{eq:pg_rv}
    \begin{aligned}
    \omega~{\buildrel D \over =}~\frac{1}{2\pi^2}\sum_{k=1}^{\infty} \frac{g_k}{(k - 1/2)^2 + c^2/(4\pi^2)},\\
    \end{aligned}
\end{equation}
where $b >0$, $c \in \sR$, and $g_k \sim Gamma(b,~1)$. 

This random variable was proposed in \cite{polya_gamma} as it has the following desired property - For $b > 0$ the following identity holds:
\begin{equation}\label{eq:pg_identity}
    \frac{(e^{\rf})^a}{(1+e^{\rf})^b}=2^{-b}e^{\kappa\rf}\mathbb{E}_{\omega}[e^{-\omega\rf^2/2}],
\end{equation}
where $\kappa=a-b/2$, and $\omega$ has the \pg distribution, $\omega\sim PG(b,~0)$. In \cite{polya_gamma} the authors also devised an efficient sampling algorithms for \pg random variables.

Suppose we are given with latent function values $\rvf \in \sR^N$ having a binary classification assignment $\rvy \in \{0, 1\}^N$. Let the prior over $\rvf \sim \normal(\rvmu,~\rmK)$. The likelihood can be written as, 
\begin{equation}\label{eq:node_likelihood}
    \begin{aligned}
    p(\rvy | \rvf) &= \prod_{j = 1}^{n} \sigma(f_j)^{y_j} (1 - \sigma(f_j))^{1 - y_j} = \prod_{j = 1}^{n} \frac{(e^{f_j})^{y_j}}{1 + e^{f_j}}=\E_{\rvomega}\left[2^{-n}\exp\left(\sum_{j=1}^{n} \kappa_j f_j- \frac{\omega_j f_j^2}{2} \right)\right],
    \end{aligned}
\end{equation}
where $\sigma(\cdot)$ is the sigmoid function. Namely, we used \Eqref{eq:pg_identity} to augment the model with \pg variables (one per sample) such that the original likelihood is recovered when marginalizing them out. 

Now, the augmented likelihood, $p(\rvy | \rvf, \rvomega)$, is proportional to a diagonal Gaussian and the posteriors in the augmented space have known expressions:
\begin{equation}\label{eq:posterior_dist}
    \begin{aligned}
        p(\rvf | \rvy, \rvomega) &= \normal(\rvf | \rmSigma(\K^{-1}\rvmu + \rvkappa),~\rmSigma),\\
        p(\rvomega | \rvy, \rvf) &= PG(\bld{1},~\rvf).
    \end{aligned}
\end{equation}
Where $\kappa_j = y_j - 1/2$, $\bld{\Sigma} = (\K^{-1} + \rmOmega)^{-1}$, and $\rmOmega = diag(\rvomega)$. We can now sample from $p(\rvf,\rvomega|\rvy, \rmX)$ using block Gibbs sampling and get Monte-Carlo estimations of the marginal and predictive distributions.

\subsection{GP-Tree} \label{sec_app:gp_tree}
Our method builds upon the method presented in \cite{achituve2021gp_icml}. It was shown to scale well both with dataset size and the number of classes, outperforming other GPC methods on standard benchmarks. We provide here a summary of this method, termed GP-Tree. GP-Tree uses the \pg augmentation, which is designed for binary classification tasks, in a (binary) tree-structure hierarchical model for multi-class classification tasks. Given a training dataset $D = (\rmX, \rvy)$ of features and corresponding labels from \{1, ..., T\} classes, $D$ is partitioned recursively to two subsets, according to classes, at each tree level until reaching leaf nodes with data from only one class. More concretely, initially, feature vectors for all samples are obtained (using a NN), then a class prototype is generated by averaging the feature vectors belonging to the same class for all classes. Finally, a tree is formed using the divisive hierarchical clustering algorithm k-means++ \cite{arthur2007k} with $k=2$ on the class prototypes. Partitioning the data in this manner is sensible since NNs tend to generate points that cluster around a single prototype for each class \cite{snell2017prototypical}. After building the tree, a GP model is assigned to each node to make a binary decision based on the data associated with that node. Let $f_{v} \sim\mathcal{GP}(m_{v}, k_{v})$ be the GP associated with node $v$. We denote all the GPs in the tree with $\bld{\mathcal{F}}$. The induced likelihood of a data point having the class $t$ is given by the unique path $P^t$ from the root to the leaf node corresponding to that class:
\begin{align} \label{eq:tree_likelihood}
    & p(y=t | \bld{\mathcal{F}}) = \prod_{v \in P^t} \sigma(f_{v})^{y_{v}} (1 - \sigma(f_{v}))^{1 - y_{v}},
\end{align}
where $y_{v}=1$ if the path goes left at $v$ and zero otherwise. $y_{v}$ can be viewed as the (local) node label assignment of the example. Since this likelihood factorizes over the nodes, we can look at the nodes separately. Therefore, in the following, we will omit the subscript $v$ for brevity; however, all datum and quantities are those that belong to a specific node $v$.

In \cite{achituve2021gp_icml} two methods for applying GP inference were suggested: a variational inference (VI) approach and a Gibbs sampling procedure. The former is used when datasets are large by constructing a variational lower bound to learn the model parameters (e.g., the NN parameters), while the latter is used for Bayesian inference only when using a fixed model. Here we will focus on learning and inference with the Gibbs sampling procedure only (see main text for further details). To obtain the augmented marginal distribution and augmented predictive distribution for a novel point $\rvx^*$ at each node, we can sample $\rvomega$ (a vector for each node) and use the following rules:
\begin{equation} \label{eq:node_marginal_likelihood}
    \begin{aligned}
    p(\rvy | \rvomega, \rmX) &= \int p(\rvy | \rvomega, \rmX, \rvf)p(\rvf)d\rvf \\
    &\propto \normal(\rmOmega^{-1} \rvkappa | \bld{0}, \K + \rmOmega^{-1}),
    \end{aligned}
\end{equation}
\begin{equation} \label{eq:node_predictive_posterior}
    \begin{aligned}
    p(f^* | \rvx^*, \rmX, \rvy, \rvomega) &= \normal({f^* |\mu^*,  \Sigma^*}), \\
    \mu^* &= (\rvks)^T(\rmOmega^{-1} + \K)^{-1} \rmOmega^{-1}\rvkappa,\\
    \Sigma^* &= k^{**} - (\rvks)^T(\rmOmega^{-1} + \K)^{-1}\rvks,
    \end{aligned}
\end{equation}
\begin{equation} \label{eq:node_predictive_likelihood}
    \begin{aligned}
    p(y^* | \rvx^*, \rmX, \rvy, \rvomega) &= \int p(y^* | f^*)p(f^* | \rvx^*, \rmX, \rvy, \rvomega)df^*.
    \end{aligned}
\end{equation}

Where we assumed a zero mean prior, $k^{**} = k(\rvx^*, \rvx^*)$, $\rvk^*[i]=k(\rvx_i, \rvx^*)$, and $\rmK[i,j]=k(\rvx_i,\rvx_j)$. The integral in \Eqref{eq:node_predictive_likelihood} is intractable, but can be computed numerically with 1D Gaussian-Hermite quadrature.

\section{pFedGP-IP-compute detailed derivation} \label{sec_app:pFedGP_comp_derivation}
Here we describe in more detail our pFedGP-IP-compute variant. 
The key idea behind this method is to cast all the dependence on the inducing points and assume independence between the latent function values given the inducing points. Since the inference problem factorizes over the clients and tree nodes, we may compute all quantities separately for each client and tree node and only afterward aggregate the results. Therefore, in the below, we omit the subscripts denoting the client and node; however, all quantities belong to a specific client and node. You may recall that $\rmXbar \in \sR^{M\times d}$ denote the pseudo-inputs, defined in the embedding space of the last layer of the NN and are shared by all clients, and $\rvfbar \in \sR^{M}$ are the corresponding latent function values. We assume the following GP prior $p(\rvf, \rvfbar) = \normal(\bld{0},[\begin{matrix} \Knn & \Knm \\ \Kmn & \Kmm \end{matrix}])$. The data likelihood of the dataset when factoring the inducing variables and the \pg variables (one per training sample) is proportional to a Gaussian:
\begin{equation} \label{eq:ip2_likelihood}
    \begin{aligned}
    p(\rvy | \rmX, \rvomega, \rvfbar, \rmXbar) &= \displaystyle \prod_{n=1}^{N} p(y_n | \rvx_n, \rvomega, \rvfbar, \rmXbar) \propto \normal(\rmOmega^{-1} \rvkappa | \Knm\KmmInv\rvfbar,~\rmLambda),
    \end{aligned}
\end{equation}
where, $\rmOmega = diag(\rvomega)$, $\kappa_j = y_j - 1/2$, and $\rmLambda = \rmOmegaInv + diag(\Knn - \Knm\KmmInv\Kmn)$. 

The posterior over $\rvfbar$ is obtained using Bayes rule:
\begin{equation} \label{eq:posterior_fbar}
    \begin{aligned}
    p(\rvfbar | \rvomega, \rvy, \rmX, \rmXbar) = \normal(\rvfbar | \Kmm\rmQInv\Kmn\rmLambdaInv \rmOmega^{-1} \rvkappa,~\Kmm\rmQInv\Kmm),
    \end{aligned}
\end{equation}
where $\rmB = \Kmm + \Kmn\rmLambdaInv\Knm$.

The posterior distribution over $\rvf$ can be obtained by marginalization over $\rvfbar$:
\begin{equation} \label{eq:ip2_posterior_f_appendix}
    \begin{aligned}
    p(\rvf | \rvomega, \rvy, \rmX, \rmXbar) &= \int p(\rvf | \rvfbar, \rmX, \rmXbar)p(\rvfbar | \rvomega, \rvy, \rmX, \rmXbar) d\rvfbar  \\
    &= \normal(\rvf | \Knm\rmQInv\Kmn\rmLambdaInv\rmOmegaInv\rvkappa,~\Knn - \Knm(\KmmInv - \rmQInv)\Kmn).
    \end{aligned}
\end{equation}
We note that while we use $\rvfbar$ for predictions, we still need $\rvf$ in order to sample $\rvomega$.

Given $\rvomega$ and the expression for the posterior over $\rvfbar$, we can compute the predictive distribution for a novel input $\rvx^*$:
\begin{equation} \label{eq:ip2_posterior_pred_f_appendix}
    \begin{aligned}
    p(f^* | \rvx^*, \rvomega, \rvy, \rmX, \rmXbar) &= \int p(f^* | \rvx^*, \rvfbar)p(\rvfbar | \rvomega, \rvy, \rmX, \rmXbar) d\rvfbar  \\
    &= \normal(f^* | (\rvk^*)^T\rmQInv\Kmn\rmLambdaInv\rmOmegaInv\rvkappa,~k^{**} - (\rvk^{*})^T(\KmmInv - \rmQInv)\rvk^{*}),
    \end{aligned}
\end{equation}
where $k^{**} = k(\rvx^*, \rvx^*)$, and $\rvk^*[i]=k(\rvxbar_i, \rvx^*)$.

The marginal distribution is given by:
\begin{equation} \label{eq:ip2_marginal_appendix}
    \begin{aligned}
    p(\rvy | \rvomega, \rmX, \rmXbar) &= \int p(\rvy | \rvomega, \rvfbar, \rmX, \rmXbar)p(\rvfbar | \rmXbar) d\rvfbar \propto \normal(\rmOmegaInv\rvkappa | \bld{0},~\rmLambda + \Knm\KmmInv\Kmn).
    \end{aligned}
\end{equation}

The full model marginal likelihood is given by:
\begin{equation}\label{eq:tree_objective_ml_appendix}
    \begin{aligned}
        \mathcal{L}^{ML}_c (\NNP ; D_c) &= \sum_{v} \log~p_{\NNP}(\rvy_{v} | \rmX_{v}, \rmXbar_{v}) = \sum_{v} \log~\int p_{\NNP}(\rvy_{v} | \rvomega_{v}, \rmX_{v}, \rmXbar_{v}) p(\rvomega_{v}) d\rvomega_{v},
    \end{aligned}
\end{equation}

and the predictive distribution for a single data point $\rvx^*$ having the class $y^*$:
\begin{equation}\label{eq:tree_objective_pl_appendix}
    \begin{aligned}
        \mathcal{L}^{PD}_c (\NNP ;\rvx^*,y^*) &= \sum_{v \in P^{y^*}} \log~p_{\NNP}( y_{v}^{*} | \rvx_{v}^{*}, \rvy_{v},  \rmX_{v}, \rmXbar_{v})\\
        &= \sum_{v \in P^{y^*}} \log~\int p_{\NNP}( y_{v}^{*} | \rvomega_{v}, \rvx_{v}^{*}, \rvy_{v}, \rmX_{v}, \rmXbar_{v}) p(\rvomega_{v} | \rvy_{v}, \rmX_{v}) d\rvomega_{v}.\\
        &= \sum_{v \in P^{y^*}} \log~\int p(\rvomega_{v} | \rvy_{v}, \rmX_{v})~\int p_{\NNP}( y_{v}^{*} | f_{v}^{*}) p(f_{v}^{*} | \rvomega_{v}, \rvx_{v}^{*}, \rvy_{v}, \rmX_{v}, \rmXbar_{v}) df_v^{*} d\rvomega_{v}.
    \end{aligned}
\end{equation}

To learn the model parameters, we first use block gibbs sampling with the posterior distributions $p(\rvf | \rvy, \rvomega, \rmX, \rmXbar)$, and $p(\rvomega | \rvy, \rvf) = PG(\bld{1}, \rvf)$. Then, we use Fisher's identity \cite{douc2014nonlinear} to obtain gradients w.r.t the model parameters with the marginal or predictive distributions.

Note that to speed up inference at test time, some computations that do not depend on $\rvx^*$ can be done offline. Importantly, we can sample and cache $\rvomega$ and use it to compute $\rmLambda$ and the Cholesky decomposition of $\rmB$.

\section{Generalization bound - derivation} \label{sec_app:generalization_bound}
In \Secref{sec:analysis} we presented an expression for the KL-divergence between the posterior and the prior (\Eqref{eq:kl_GP-PG}). Now we present the derivation:
\begin{equation}\label{eq:kl_GP-PG_dev}
\begin{aligned}
    \KL[Q(\rvf)~||~P(\rvf)] &= \int Q(\rvf)\log~\frac{Q(\rvf)}{P(\rvf)}d\rvf\\
    &= \int Q(\rvf, \rvomega) \log~\frac{Q(\rvf)}{P(\rvf)}d\rvf d\rvomega\\
    &= \int Q(\rvf, \rvomega) \log~\frac{Q(\rvf)Q(\rvomega|\rvf)}{P(\rvf)Q(\rvomega|\rvf)}d\rvf d\rvomega\\
    &= \int Q(\rvf, \rvomega) \log~\frac{Q(\rvf | \rvomega)}{P(\rvf)}d\rvf d\rvomega + \int Q(\rvf, \rvomega)\log~\frac{Q(\rvomega)}{Q(\rvomega | \rvf)}d\rvf d\rvomega\\
     &= \int Q(\rvf, \rvomega) \log~\frac{Q(\rvf | \rvomega)}{P(\rvf)}d\rvf d\rvomega + \int Q(\rvf, \rvomega)\log~\frac{Q(\rvf)Q(\rvomega)}{Q(\rvf, \rvomega)}d\rvf d\rvomega\\
    &= \E_{Q(\rvomega)}\{\KL[Q(\rvf | \rvomega) || P(\rvf)]\} - MI[\rvf;\rvomega],
\end{aligned}
\end{equation}

Where, the KL-divergence in the expectation term is between the prior, $P(\rvf) = \normal(\rvmu, \rmK)$, and the posterior, $Q(\rvf | \rvomega) = \normal(\rmSigma(\rmKInv\rvmu + \rvkappa),~\rmSigma)$. Therefore, it has the following closed-form:
\begin{equation} \label{eq:kl_gaussians}
    \small
    \begin{aligned}
    \KL[Q(\rvf | \rvomega) || P(\rvf)] = \frac{1}{2}\{log~\frac{|\rmK|}{|\rmSigma|} - N_c + tr(\rmKInv\rmSigma) + (\rmSigma(\rmKInv\rvmu + \rvkappa) - \rvmu)^T \rmKInv (\rmSigma(\rmKInv\rvmu + \rvkappa) - \rvmu)\}
    \end{aligned}
\end{equation}

	

\section{Experimental details} \label{sec:exp_details}

\paragraph{Datasets.} We evaluated pFedGP on the CIFAR-10, CIFAR-100 \cite{krizhevsky2009learning}, and CINIC-10 \cite{darlow2018cinic} datasets. CIFAR-10/100 contain 60K images from 10/100 distinct classes, respectively, split to 50K for training and 10K for testing. CINIC-10 is a more diverse dataset. It is an extension of CIFAR-10 via the addition of down-sampled ImageNet \cite{deng2009imagenet} images. It contains 270K images split equally to train, validation, and test sets from 10 distinct classes.
\begin{figure}[!t]
    \centering
    \begin{minipage}{.45\textwidth}
        \centering
        \includegraphics[width=0.7\linewidth]{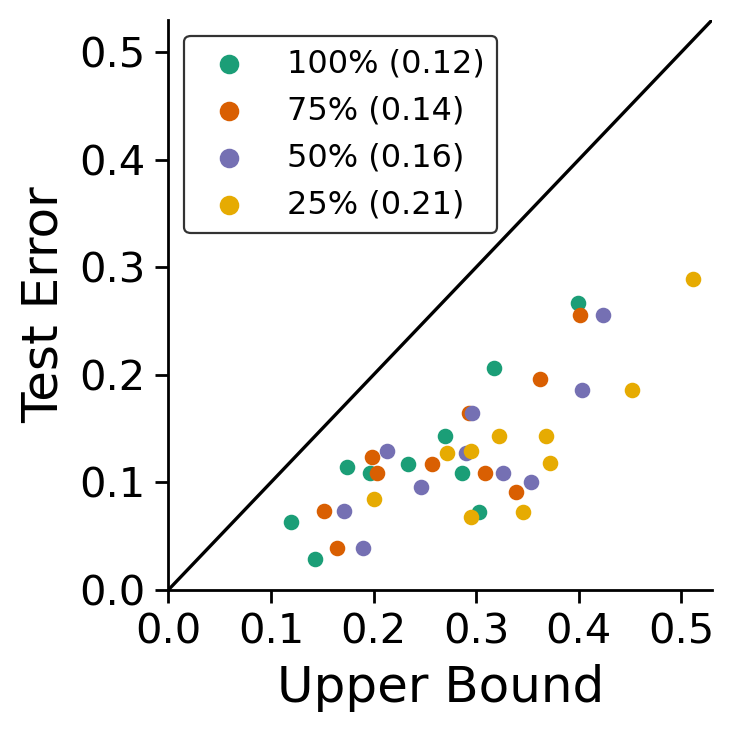}
        \caption{Test error vs. an estimated upper bound over 10 clients on CIFAR-10 with varying degrees of a training set data size using the Bayes classifier. Each dot represents a combination of client and data size. In parenthesis - the average difference between the empirical and the test error.}
        \label{fig:gen_vs_test_error_bayes}
    \end{minipage}%
    \hskip 1.0cm
    \begin{minipage}{0.45\textwidth}
        \vspace{-25pt}
        \centering
        \includegraphics[width=1.\linewidth]{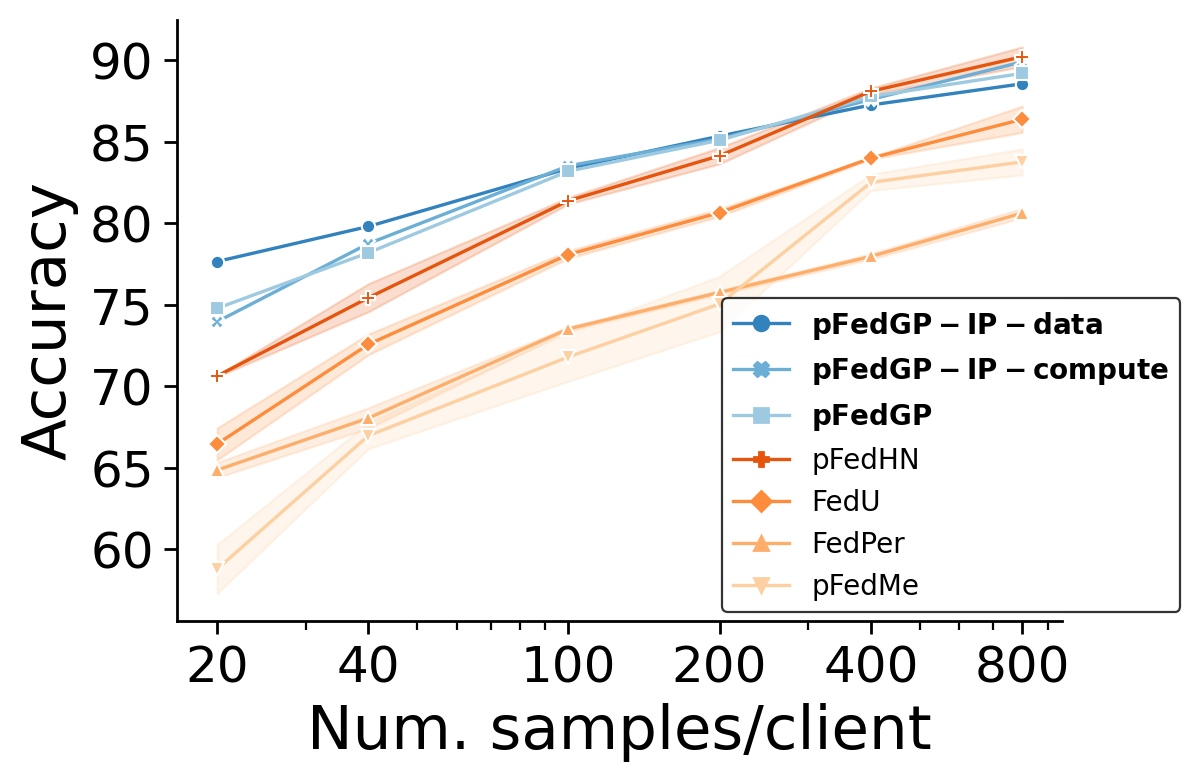}
        \caption{Model performance with varying degrees of an average number of training samples per client (x-axis in log scale). Results are over 50 clients on CIFAR-10.}
        \label{fig:varying}
    \end{minipage}
\end{figure}

\textbf{Data assignment.} For partitioning the data samples across clients we followed the procedure suggested in \cite{shamsian2021personalized_icml, t2020personalized}. This procedure produces clients with a varying number of samples and a unique set of $k$ classes for each client. First, we sampled $k$ classes for each client. In general, we used $k=2$ in CIFAR-10 experiments (e.g., Sections \ref{sec:analysis} \& \ref{sec:hetro}), $k=10$ in CIFAR-100 experiments (e.g., Sections \ref{sec:hetro} \& \ref{sec:pfl_in_noise}), and $k=4$ in CINIC-10 experiments (e.g., Section \ref{sec:hetro}). Next, to assign unique images for each client, we iterated over the classes and clients; for each client $i$ having the class $c$, we sampled an unnormalized class fraction $\alpha_{i,c}\sim U(.4,.6)$. We then assigned to the $i^{th}$ client $\frac{\alpha_{i,c}}{\sum_j \alpha_{j,c}}$ images from the overall samples of class $c$.

\textbf{Hyperparameter tuning.} We used a validation set for hyperparameter selection and early stopping in all methods. For the CIFAR datasets, we pre-allocated a validation set of size 10K from the training set. For the CINIC-10 dataset, we used the original split having a validation set of size 90K. The hyperparameters for all methods and all datasets were selected based on the learning setups with $50$ clients. We searched over the learning-rates $\{1e-1, 5e-2, 1e-2, 5e-3\}$ for all methods, and personal learning-rates $\{5e-2, 1e-2, 5e-3, 1e-3\}$ for baseline methods only (pFedGP is a non-parametric approach and therefore does not optimize any private parameters). For pFedGP we searched over the number of epochs on sampled clients during training in $\{1, 3\}$. For baseline methods, since the training procedure varies significantly, we ran the baselines FOLA, LG-FedAvg, pFedMe, Per-FedAvg, FedU, and pFedHN according to the recommended configuration in their papers or code (which is usually a few epochs). Regarding FedAvg, FedPer, and pFedMe, we followed the protocol suggested by pFedME of using 20 iterations per client. For pFedGP We also searched over a scaling factor for the loss function in $\{1, 2\}$. We used the RBF kernel function with a fixed length scale of $1$ and an output scale of $8$. We used $20$ parallel Gibbs chains for training and $30$ parallel Gibbs chains for testing with $5$ MCMC steps between samples in both. To compute the predictive distribution, at each tree node, we averaged over the log probabilities since it didn't impact the results but yielded a more calibrated model. In the reliability diagrams, we used $50$ steps since as the number of steps increases usually the model becomes better calibrated (without a discernible accuracy change). In pFedGP-IP-data and pFedGP-IP-compute experiments, we used $100$ inducing points per class. In all baselines that use FedAvg update rule, we used a variant of FedAvg in which a uniform weight was given to all clients during the global network update.
All experiments were done on NVIDIA GeForce RTX 2080 Ti having 11GB of memory.

\textbf{Noisy datasets (Sections \ref{sec:pfl_in_noise} \& \ref{sec:in_noise_cifar10}).} To generate a noisy version of the CIFAR-10 and CIFAR-100 datasets, we used the image-corruptions package \cite{hendrycks2018benchmarking}. We simulated the following $19$ corruptions (Gaussian noise, shot noise, impulse noise, defocus blur, glass blur, motion blur, zoom blur, snow, frost, fog,
brightness, contrast, elastic transform, pixelate, jpeg compression,
speckle noise, Gaussian blur, spatter, saturate) with a corruption severity of (3, 4, 5), resulting in $57$ unique noise models.

\section{Additional experiments} \label{sec:additional_exp}

\subsection{Generalization bound - additional experiments}

\begin{figure}[!t]
\centering
    \begin{subfigure}[ResNet-18]{
    \includegraphics[width=0.35\linewidth]{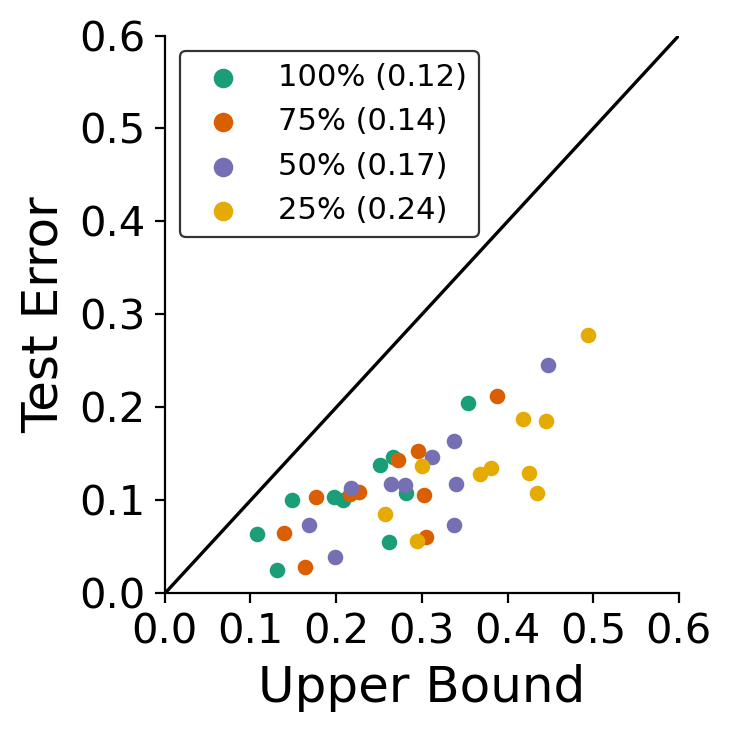}
    }
    \end{subfigure}
    \begin{subfigure}[MobileNetV2]{
    \includegraphics[width=0.35\linewidth]{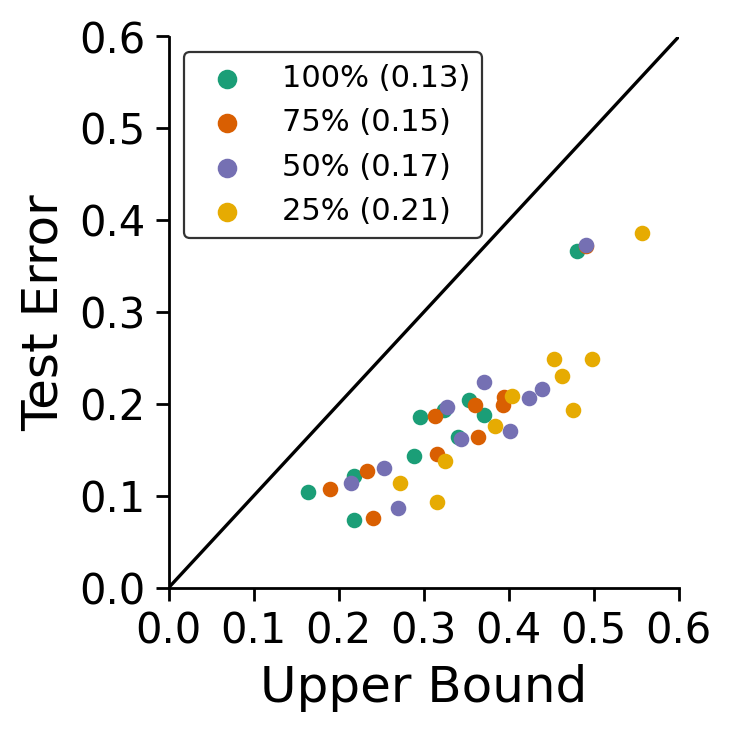}
    }
     \end{subfigure}
    \caption{Test error vs. an estimated upper bound over 10 clients on CIFAR-10 with varying degrees of a training set data size on ResNet-18 (left) and MobileNetV2 (right). Each dot represents a combination of client and data size. In parenthesis - the average difference between the empirical and the test error.}
    \label{fig:bound_larger_networks}
\end{figure}
\textbf{Generalization bound with the Bayes risk.} In \Secref{sec:analysis} we assessed the quality of the lower bound with the Gibbs risk. However, often we are interested in a deterministic predictor. Also, we would like to get an estimate of the error with a classifier that is closer to our estimate of $y^*$ with the Gauss-Hermite quadrature. This can be achieved with the Bayes risk \cite{reeb2018learning} defined by $R_{Bayes}(Q) := \E_{(\rvx^*, y^*)}[sign\left( \E_{f^* \sim Q(f^* | \rvx^*, D_c)}[f^*] \right)\neq y^*]$. \Figref{fig:gen_vs_test_error_bayes} shows an estimation of the generalization error bound vs. the actual error on the novel clients with the Bayes classifier. From this figure we observe similar patterns to those seen in \Figref{fig:gen_vs_test_error}. In general, the Bayes classifier performs better than the Gibbs classifier.

\textbf{Generalization with bigger networks.} To test how the bound behaves with larger networks, we also evaluated the Gibbs risk on ResNet18 \cite{he2016deep} and MobileNetV2 \cite{sandler2018mobilenetv2}, having $\sim$ 11.4M and $\sim$ 2.8M parameters correspondingly. In both networks, we replaced the final fully-connected layer with a linear layer of size $512$ and we removed batch normalization layers. Results are presented in \Figref{fig:bound_larger_networks}. We observe a similar behavior with these networks to the one seen with the LeNet backbone. Namely, the bound is data-dependent and gives non-vacuous guarantees.

\subsection{Varying the training set size} \label{sec:varying_training_size}


To decouple the effect of the local dataset size from the number of clients in the system, we altered the setting in \Secref{sec:hetro}. Here, we fixed the number of clients to $50$ and used stratified sampling to sample $\{1000, 2000, 5000, 10000, 20000, 40000\}$ examples from the training dataset of CIFAR-10, where $40000$ is the total number of training examples. Then, similar to \Secref{sec:hetro} we randomly assigned two classes per client and partitioned the (sampled) data across clients. \Figref{fig:varying} shows that all methods suffer from accuracy degradation when the training data size decrease; however, in pFedGP methods, the reduction is less severe compared to baseline methods. We especially note pFedGP-IP-data which shows remarkable accuracy in the extremely low data regime ($77.7\%$ accuracy with only $1000$ training examples). These results can be attributed to the sharing of the inducing inputs which effectively increase the size of the training data per client. The LG-FedAvg baseline was excluded from this figure since it showed low accuracy.

\begin{table}[!t]
\parbox{.45\linewidth}{
\setlength{\tabcolsep}{3pt}
\small
\caption{Test accuracy ($\pm$ SEM) over 100 clients on noisy CIFAR-10 under a homogeneous class distribution.}
\vskip 0.15in
\centering
\begin{tabular}{l c c} 
    \toprule
    \multicolumn{1}{l}{Method} && \multicolumn{1}{c}{Accuracy}\\
    \midrule
    FedAvg \cite{mcmahan2017communication} && 42.0 $\pm$ 0.2 \\
    \midrule
    FedPer \cite{arivazhagan2019federated} && 38.2 $\pm$ 2.0 \\
    LG-FedAvg \cite{liang2020think} && 41.5 $\pm$ 0.2 \\
    pFedme \cite{t2020personalized} && 36.3 $\pm$ 0.0 \\
    FedU  \cite{dinh2021fedu} && 24.8 $\pm$ 0.0 \\
    pFedHN \cite{shamsian2021personalized_icml} && 35.7 $\pm$ 0.3 \\
    \midrule
    \textbf{Ours} &&\\
    pFedGP-IP-data && 46.1 $\pm$ .05 \\
    pFedGP-IP-compute && 46.5 $\pm$ 0.5 \\
    pFedGP && \textbf{46.7 $\pm$ 0.2} \\
    \bottomrule
\end{tabular}
\label{tab:noisy_cifar10}
}
\hfill
\parbox{.45\linewidth}{
\setlength{\tabcolsep}{3pt}
\caption{Test accuracy ($\pm$ SEM) over 50, 100 clients on CIFAR-10 under a homogeneous class distribution.}
\vskip 0.15in
\centering
\scalebox{.85}{
\begin{tabular}{l c cc c} 
    \toprule
    \multirow{2}{*}{Method} && \multicolumn{3}{c}{\# clients}\\
    \cmidrule(l){2-5}
     && 50 && 100\\
    \midrule
    FedAvg \cite{mcmahan2017communication} && \textbf{66.2 $\pm$ 0.3} && 64.6 $\pm$ 0.2\\
    \midrule
    FedPer \cite{arivazhagan2019federated} && 56.8 $\pm$ 0.1 && 50.9 $\pm$ 0.4\\
    pFedMe \cite{t2020personalized} && 46.9 $\pm$ 0.5 && 44.4 $\pm$ 0.3 \\
    FedU \cite{dinh2021fedu} && 29.6 $\pm$ 0.6 && 26.3 $\pm$ 0.5\\
    pFedHN \cite{shamsian2021personalized_icml} && 62.8 $\pm$ 0.5 && 56.5 $\pm$ 0.1\\
    \midrule
    \textbf{Ours} &&&&\\
    pFedGP-IP-data && 62.7 $\pm$ 0.7 && 62.4 $\pm$ 0.5\\
    pFedGP-IP-compute && 65.8 $\pm$ 0.3 && 65.2 $\pm$ 0.4\\
    pFedGP &&  \textbf{66.2 $\pm$ 0.3} && \textbf{65.7$\pm$ 0.2}\\
    \bottomrule
\end{tabular}}
\label{tab:cifar10_10classes}
}
\end{table}
\subsection{PFL with input noise under a homogeneous class distribution} \label{sec:in_noise_cifar10}
In \Secref{sec:pfl_in_noise} we evaluated pFedGP under two types of variations between the clients: (i) a unique input noise, and (ii) in the class distribution. It was done on the CIFAR-100 dataset with $100$ clients. Here, we consider only a shift in the input noise between clients while having a balanced class distribution in all clients. We do so on the CIFAR-10 dataset (i.e., each client has $10$ classes distributed approximately equal). Similarly to \Secref{sec:pfl_in_noise} we configured $100$ clients, distributed the data among them, and assigned a noise model to each client from a closed-set of $57$ noise distributions. \tblref{tab:noisy_cifar10} shows that in this setting as well, pFedGP and its variants achieve high accuracy and surpass all baseline methods by a large margin. From the table, we also notice that FedAvg, which performs well in balanced class distribution setups, outperforms all competing methods except ours.
\begin{table}[!t]
\setlength{\tabcolsep}{3pt}
\caption{pFedGP-IP-compute vs. pFedGP full model test accuracy and average predictive posterior inference run-time ($\pm$ STD) as a function of the number of inducing points (IPs) over 50 clients on CINIC-10.
}
\vskip 0.15in
\centering
\scalebox{.78}{
\begin{tabular}{l c c c c c c c c} 
    \toprule
    Num. IPs && $80$ & $160$ & $240$ & $320$ & $400$ & | & Full GP\\
    \midrule
    Accuracy (\%) && 70.2 & 71.0 & 71.4 & 71.5 & 71.5 & | & 71.3\\
    Run time (sec.) &&  0.15 \scriptsize{$\pm$ .02} & 0.21 \scriptsize{$\pm$ .03} & 0.27 \scriptsize{$\pm$ .04} &  0.34 \scriptsize{$\pm$ .05} & 0.42 \scriptsize{$\pm$ .06} & | & 1.08 \scriptsize{$\pm$ .16}\\
    \bottomrule
\end{tabular}}
\label{tab:num_inducing}
\end{table}

\subsection{Homogeneous federated learning with CIFAR-10} \label{sec:cifar10_iid}
pFedGP is a non-parametric classifier offered for personalized federated learning. Therefore, it only needs to model classes that are present in the training set. Here, for completeness, we evaluate pFedGP against top-performing baselines on the CIFAR-10 dataset, where all classes are represented in all clients. We do so on a homogeneous federated learning setup, i.e., all classes are distributed equally across all clients. Data assignment was done similarly to the procedure described in \Secref{sec:hetro} (see \Secref{sec:exp_details} for more details). \tblref{tab:cifar10_10classes} shows that pFedGP outperforms all PFL baseline methods by a large margin in this setting as well. An interesting, yet expected, observation from the table is that FedAvg performs well under this (IID) setup. This result connects to a recent study that suggested that under a smooth, strongly convex loss when the data heterogeneity is below some threshold, FedAvg is minimax optimal \cite{chen2021theorem}. We note here that modeling classes that are not present in the training set with pFedGP can be accomplished easily with one of the inducing points variants of pFedGP.

\subsection{Computing demands}
\setlength{\tabcolsep}{2pt}
\begin{table*}[!t]
\small
\centering
\caption{pFedGP model variants test accuracy ($\pm$ SEM) over 50, 100, 500 clients on CIFAR-10, CIFAR-100, and CINIC-10. The \textit{\# samples/client} indicates the average number of training samples per client.}
\scalebox{.71}{
    \begin{tabular}{l c cc cc cc c cc cc cc c cc cc}
    \toprule
    \multirow{2}{*}{}
    && \multicolumn{5}{c}{CIFAR-10} && \multicolumn{5}{c}{CIFAR-100} && \multicolumn{5}{c}{CINIC-10}\\
    \cmidrule(l){3-7}  \cmidrule(l){9-13} \cmidrule(l){15-19}
    ~\quad\quad\quad \# clients &&50 &&100 &&500 &&50 &&100 &&500 &&50 &&100 &&500 \\
    ~\quad\quad\quad \# samples/client &&800 &&400 &&80 &&800 &&400 &&80 &&1800 &&900 &&180\\
    \midrule
    pFedGP-IP-data w/o personalization && 88.6 $\pm$ 1.0 && 87.0 $\pm$ 0.2 && 86.4 $\pm$ 0.7 && 58.1 $\pm$ 0.3 && 57.4 $\pm$ 0.6 && 55.4 $\pm$ 0.2 && 69.2 $\pm$ 0.3 && 68.2 $\pm$ 0.9 && 67.9 $\pm$ 0.1 \\
    pFedGP-IP-data && 88.6 $\pm$ 0.2 && 87.4 $\pm$ 0.2 && 86.9 $\pm$ 0.7 && 60.2 $\pm$ 0.3 && 58.5 $\pm$ 0.3 && \textbf{55.7 $\pm$ 0.4} && 69.8 $\pm$ 0.2 && 68.3 $\pm$ 0.6 && 67.6 $\pm$ 0.3 \\
    \midrule
    pFedGP-IP-compute-marginal && \textbf{89.8 $\pm$ 0.6} && \textbf{88.8 $\pm$ 0.3} && \textbf{87.8 $\pm$ 0.3} && 60.9 $\pm$ 0.4 && 58.8 $\pm$ 0.3 && 46.7 $\pm$ 0.3 && \textbf{72.1 $\pm$ 0.2} && 71.1 $\pm$ 0.6 && 67.5 $\pm$ 0.2 \\
    pFedGP-IP-compute-predictive && \textbf{89.9 $\pm$ 0.6} && \textbf{88.8 $\pm$ 0.1} && 86.8 $\pm$ 0.4 && 61.2 $\pm$ 0.4 && 59.8 $\pm$ 0.3 && 49.2 $\pm$ 0.3 && \textbf{72.0 $\pm$ 0.3} && \textbf{71.5 $\pm$ 0.5} && 68.2 $\pm$ 0.2 \\
    \midrule
    pFedGP-marginal &&  89.0 $\pm$ 0.1 && 88.0 $\pm$ 0.2 && 86.8 $\pm$ 0.2 && \textbf{63.7 $\pm$ 0.1} && \textbf{61.4 $\pm$ 0.3} && 50.3 $\pm$ .05 && 71.6 $\pm$ 0.3 && 71.0 $\pm$ 0.6 && \textbf{68.5 $\pm$ 0.2} \\
    pFedGP-predictive &&  89.2 $\pm$ 0.3 && \textbf{88.8 $\pm$ 0.2} && 87.6 $\pm$ 0.4 && 63.3 $\pm$ 0.1 && \textbf{61.3 $\pm$ 0.2} && 50.6 $\pm$ 0.2 && 71.8 $\pm$ 0.3 && 71.3 $\pm$ 0.4 && 68.1 $\pm$ 0.3 \\
    \bottomrule
    \end{tabular}
}
\label{tab:pred_vs_marginal}
\end{table*}
In this section we evaluate pFedGP computational requirements.
First we compared between pFedGP full model (\Secref{sub_sec:full_gp}) and pFedGP-IP-compute  (\Secref{subsec_method:gp_IP-compute}) in terms of accuracy and run-time during test time. The key component controlling the computational demand of pFedGP during test time is the predictive distribution (Equations \ref{eq:node_predictive_posterior} \& \ref{eq:ip2_posterior_pred_f_appendix}). After the training phase, when a new test sample arrives, computing the predictive distributions can be done efficiently by using cached components that depend only on the training data (e.g., the Cholesky decomposition of $\rmB$) and can be calculated offline. Therefore, to quantify the impact of using pFedGP-IP-compute compared to the full GP model, we recorded in \tblref{tab:num_inducing} the federated accuracy and average time per client for calculating the predictive distribution for all test examples as a function of the number of inducing points. The comparison was done on the pre-allocated test set from the CINIC-10 dataset over $50$ clients (i.e., $\sim 1800$ test examples per client divided to $4$ classes) using a model that was trained with $100$ inducing points. The table shows a significant improvement in the run time compared to the full model without any (or only minor) accuracy degradation. We note here that including the network processing time will add a constant factor of $0.03$ seconds.

In addition to the above test, we also tracked pFedGP full model and baseline methods memory usage and runtime on CIFAR-10 and CIFAR-100 with 50 clients during training. For comparability, we fixed the number of epochs that each sampled client makes to one for all methods. We found that pFedGP computational requirements are reasonable for running it in current FL systems \cite{cai2021towards}. 
pFedGP needed $\sim$ 1.4/1.6 GB memory, and a run-time of $\sim$ 1/2 hours for CIFAR-10/100 correspondingly. Baseline methods needed 1.1-1.3 GB memory and took $\sim$ 25 minutes to run on both datasets. According to this naive testing, pFedGP is computationally more intensive compared to standard methods such as FedAvg. However, in return for that additional complexity, it obtains substantial performance gains compared to the baseline methods. Furthermore, when implementing pFedGP there are some trade-offs, for example, iterating over the tree can be either sequential to obtain lower memory cost or parallelized to obtain shorter running times. Or, when applying the Gibbs sampling, the complexity can be affected by the number of parallel Gibbs chains and the number of MCMC steps which constitute a trade-off between accuracy and runtime. Finally, note that using only one epoch of training damaged severely the performance of several baseline methods. Therefore, to obtain the same accuracy as reported in this paper the gaps in run-time are actually smaller.

\subsection{Meta-Learning approaches for FL}
In this study we advocate the use of GPs in general, and pFedGP specifically, in FL systems. A key motivation for using GPs is that often the data on clients is limited. Alternatively, we could have used other methods that were found to work well with limited data. Specifically, methods that are based on the model-agnostic meta-learning (MAML) \cite{li2017meta} learning procedure. Here, we compare pFedGP against the MAML-based federated learning approach \textit{Per-FedAvg} \cite{Fallah2020PersonalizedFL} under the setting presented in \Secref{sec:hetro} in the main text. We observed the following results: $83.5 \pm .05 / 81.7 \pm 0.4 / 76.4 \pm 1.0$ on CIFAR-10, and $45.6 \pm 0.2 / 41.0 \pm 1.4 / 30.2 \pm .02$ on CIFAR-100 with 50/100/500 clients respectively. Comparing this method to pFedGP reveals that our approach outperforms this baseline as well.

\subsection{Predictive vs marginal likelihood}
In the main text, we presented two alternatives for learning the model parameters with pFedGP and pFedGP-IP-compute, the predictive distribution, and the marginal likelihood (see \Secref{sec:pFedGP}). We now compare between these two alternatives in \tblref{tab:pred_vs_marginal} under the standard setup presented in \Secref{sec:hetro}. The table shows that for both pFedGP and pFedGP-IP-compute the two variants are comparable with a slight advantage to the predictive distribution objective. Nevertheless, using the marginal likelihood usually results in a better-calibrated model (\Secref{sec:reliability_diad_appendix}).

\subsection{pFedGP-IP-data Ablation} \label{sec_app:pFedGP_data_wo_pers}
Recall that for the pFedGP-IP-data variant during training we build the kernel with the (shared) inducing inputs only. Yet, during test time, to account for the personal data, we use both the inducing inputs and the training data of the client for building the kernel. This method is especially effective in cases where the data per client is limited. Here, we evaluate this method without using the actual training data during test time. This means that the only personalization derives from the personal tree structure that is formed based on the actual training data of the client. Remarkably, \tblref{tab:pred_vs_marginal} shows that using this strategy yields high accuracy as well and it is often comparable to pFedGP-IP-data.

In pFedGP-IP-data we also introduced a correction term based on the class probabilities. Here we investigated the impact of this functionality as well.
When the data is distributed uniformly among the client's classes, this correction term does not have any effect. Therefore, we tested its effect under a similar setting to the one presented in \Secref{sec:exp_gen_ood_clients} on CIFAR-10. Namely, we sampled examples from classes according to a Dirichlet distribution with parameter $\alpha = 0.1$ for each client. With the class balancing, we noticed an accuracy of $84.4 \pm 0.5$ without it the accuracy dropped to $83.7 \pm 0.5$.

\subsection{Reliability diagrams} \label{sec:reliability_diad_appendix}
\begin{figure}[!t]
\centering
\tiny
\scalebox{1.0}{
\begin{tabular}[!t] {c c c c}
    ~~~~50 clients. $t = 1$ & ~~~~~~~50 clients. Best temp. & ~~~~~~~100 clients. $t = 1$ & ~~~~~~~~100 clients. Best temp.\\
    \includegraphics[width=33mm]{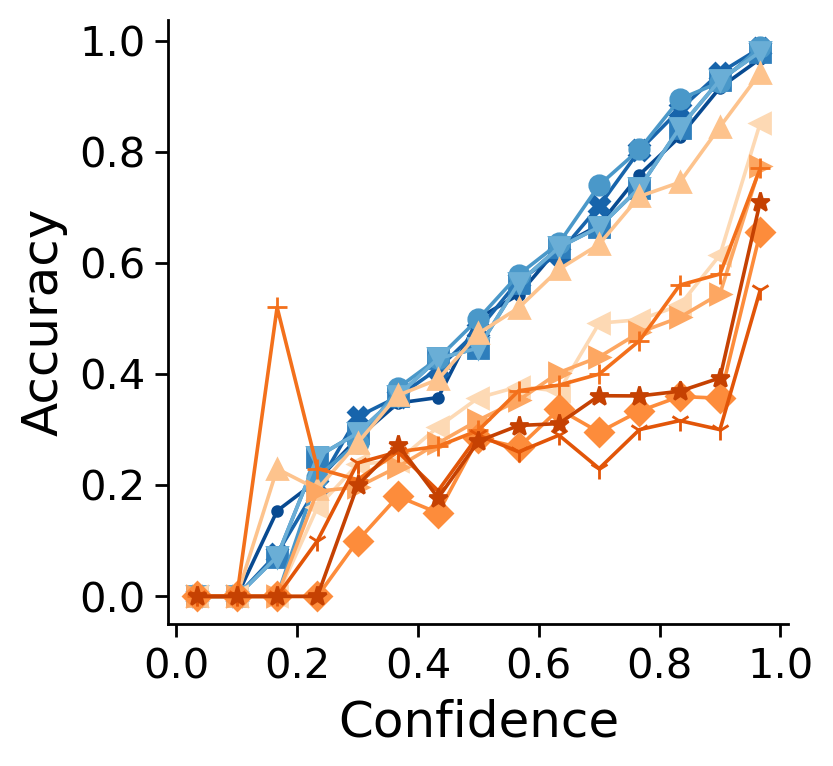} &   \includegraphics[width=33mm]{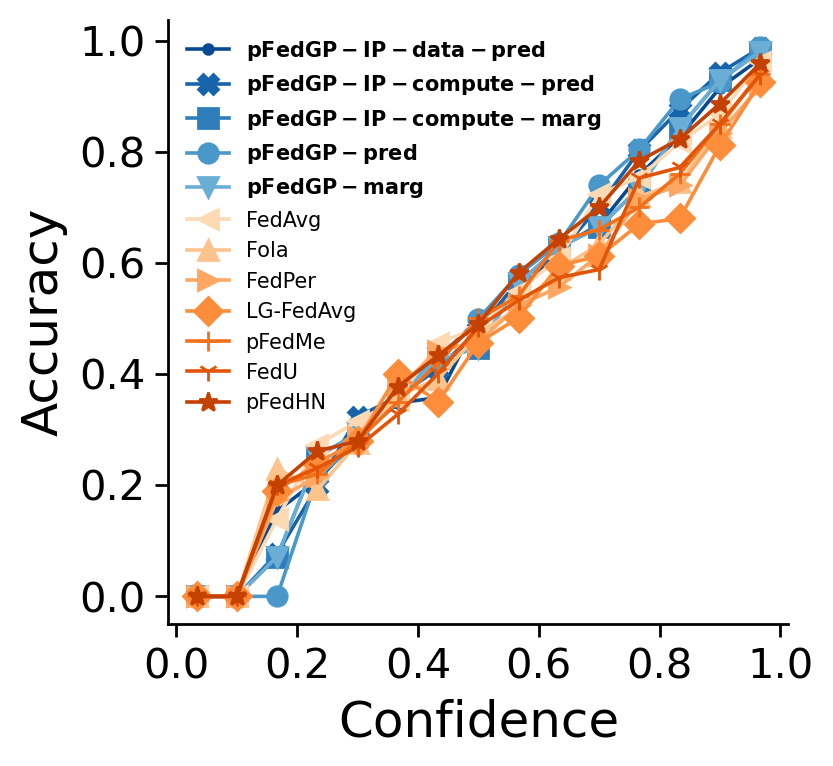} & \includegraphics[width=33mm]{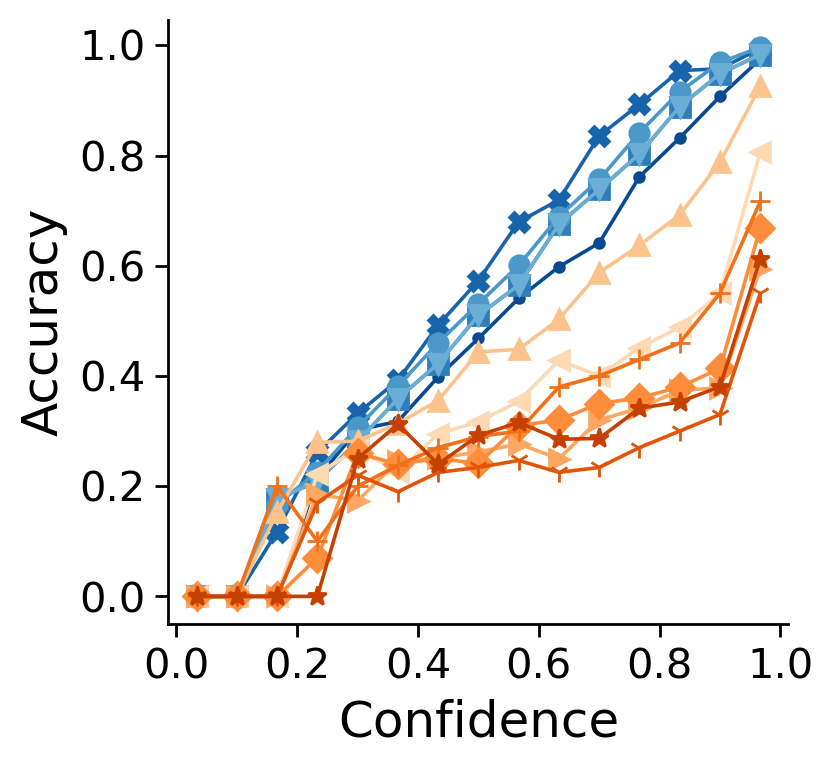} &
    \includegraphics[width=33mm]{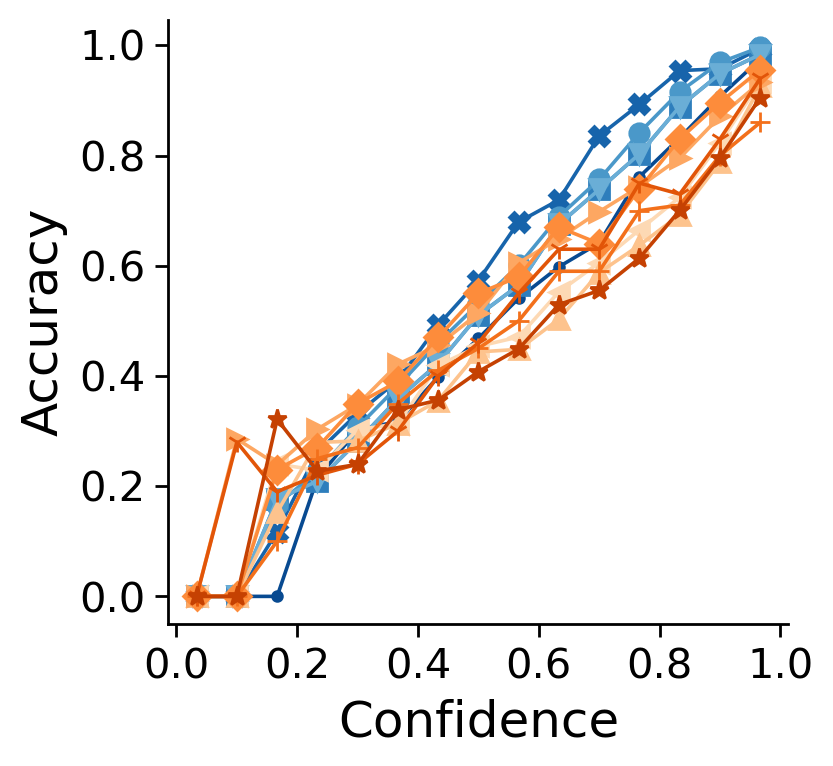}
\end{tabular}}
\caption{Reliability diagrams over 50, 100 clients on CIFAR-100.}
\label{fig:calibration_all}
\end{figure}

\begin{figure}[!t]
\centering
\tiny
\scalebox{1.0}{
\begin{tabular}{c c c c c c}
    ~~~~FedAvg & ~~~~~~FOLA & ~~~~FedPer & ~~~~LG-FedAvg & ~~~~pFedMe & ~~~~~~FedU\\
    \includegraphics[width=20mm]{figures/calibration/calibration50temp1/fedavg_fix_50_calibrate_temp1_stats.png} &   \includegraphics[width=20mm]{figures/calibration/calibration50temp1/Fola50clients_temp1.png} & \includegraphics[width=20mm]{figures/calibration/calibration50temp1/fedper_50_calibrate_temp1_stats.png} &
    \includegraphics[width=20mm]{figures/calibration/calibration50temp1/lg_fedavg_50users_temp1.png} &
    \includegraphics[width=20mm]{figures/calibration/calibration50temp1/pfedme_50_calibrate_temp1_stats.png} &
    \includegraphics[width=20mm]{figures/calibration/calibration50temp1/fedu_50_calibrate_temp1_stats.png}
    \\
     ~~~~pFedHN & ~~~~pFedGP-IP-data & ~~~~pFedGP-IP-compute-pred. & ~~~~~pFedGP-IP-compute-marg. & ~~~~pFedGP-pred. & ~~~~pFedGP-marg.
     \\
    \includegraphics[width=20mm]{figures/calibration/calibration50temp1/pfedhn_pc_50_calibrate_temp1_stats.png} &
    \includegraphics[width=20mm]{figures/calibration/calibration50bestTemp/pFed-GP-IP-data_50clients.png} &
    \includegraphics[width=20mm]{figures/calibration/calibration50bestTemp/pFed-GP-IP-compute_50clients_predictive.png} &
    \includegraphics[width=20mm]{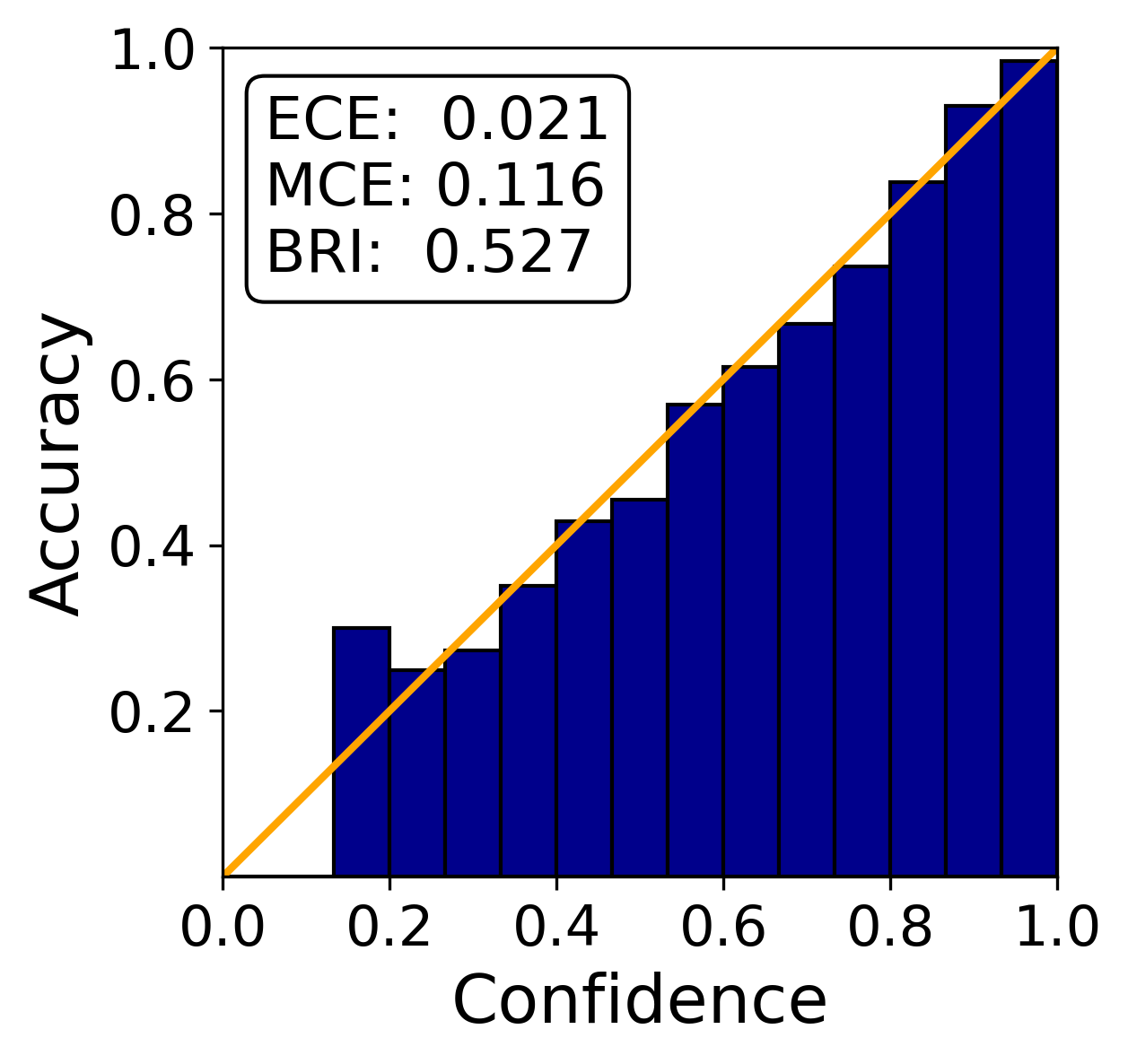} &
    \includegraphics[width=20mm]{figures/calibration/calibration50bestTemp/pFed-GP_50clients_predictive.png} &
    \includegraphics[width=20mm]{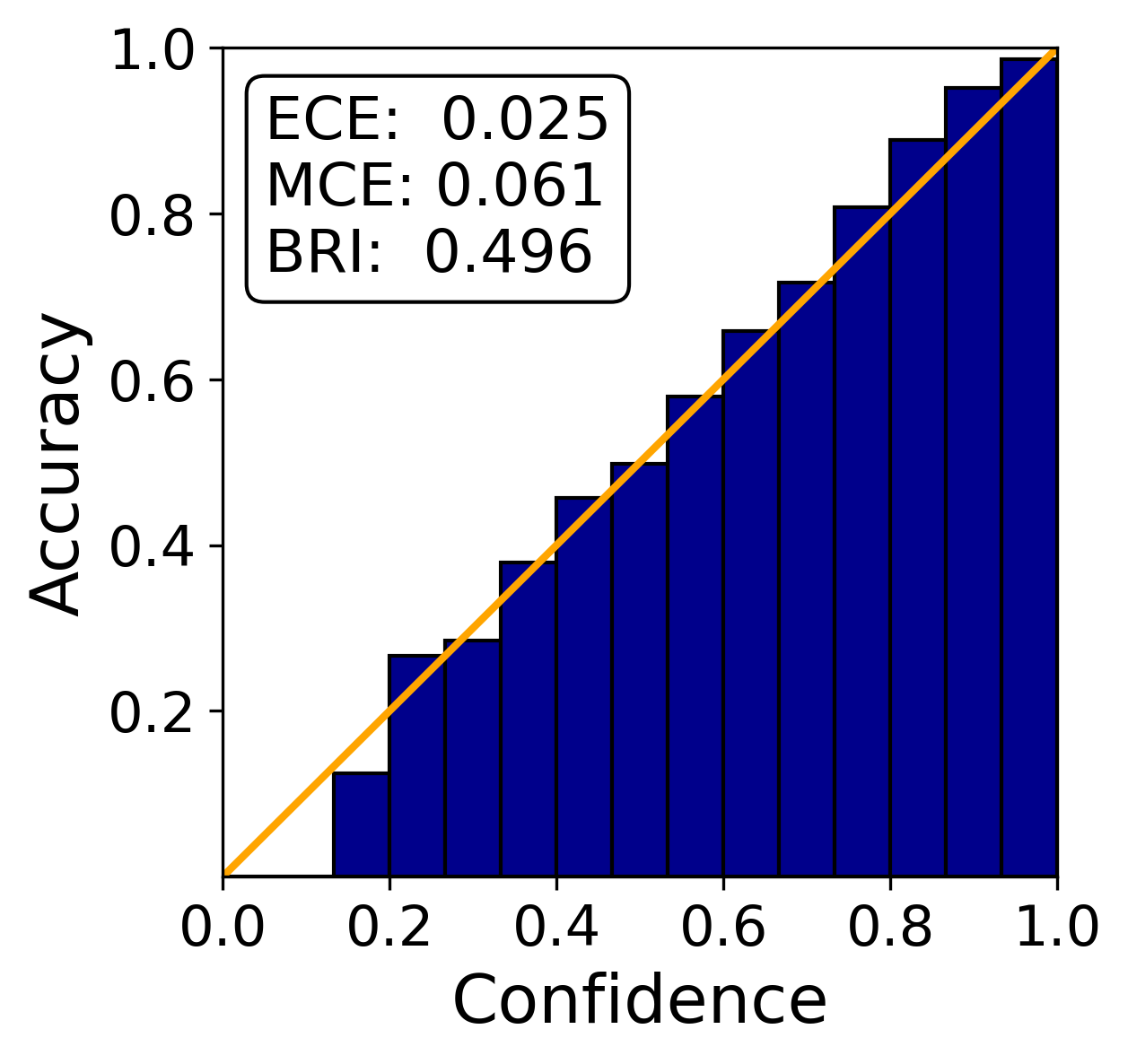}
    \end{tabular}}
    \caption{Reliability diagrams on CIFAR-100 with 50 clients. Default temperature ($t=1$). The last 5 figures are ours.}
    \label{fig:calibration_50_temp1_appendix}
\end{figure}

\begin{figure}[!t]
\centering
\tiny
\scalebox{1.0}{
\begin{tabular}{c c c c c c}
    ~~~~FedAvg & ~~~~~~FOLA & ~~~~FedPer & ~~~~LG-FedAvg & ~~~~pFedMe & ~~~~~~FedU\\
    \includegraphics[width=20mm]{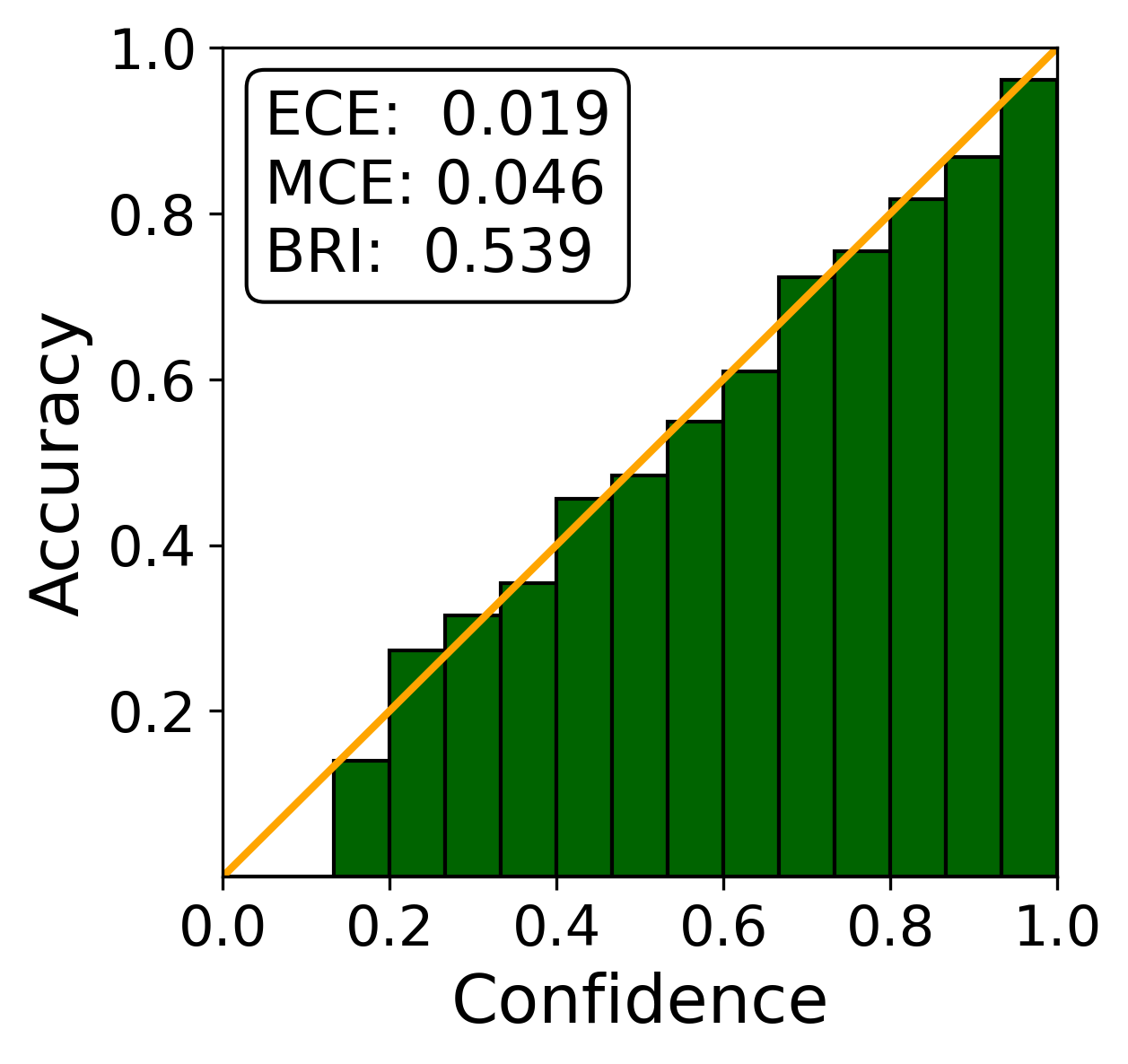} &   \includegraphics[width=20mm]{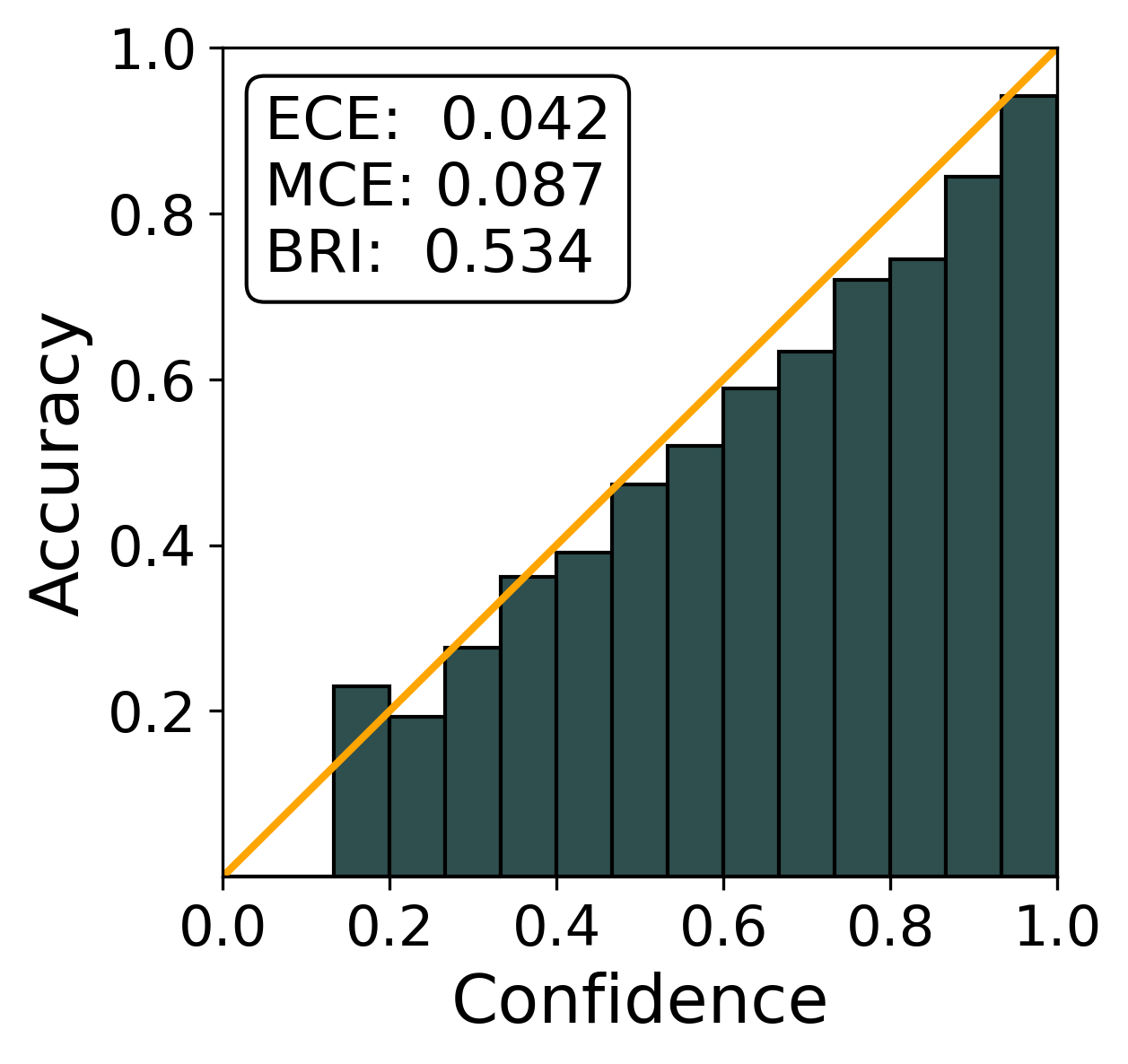} & \includegraphics[width=20mm]{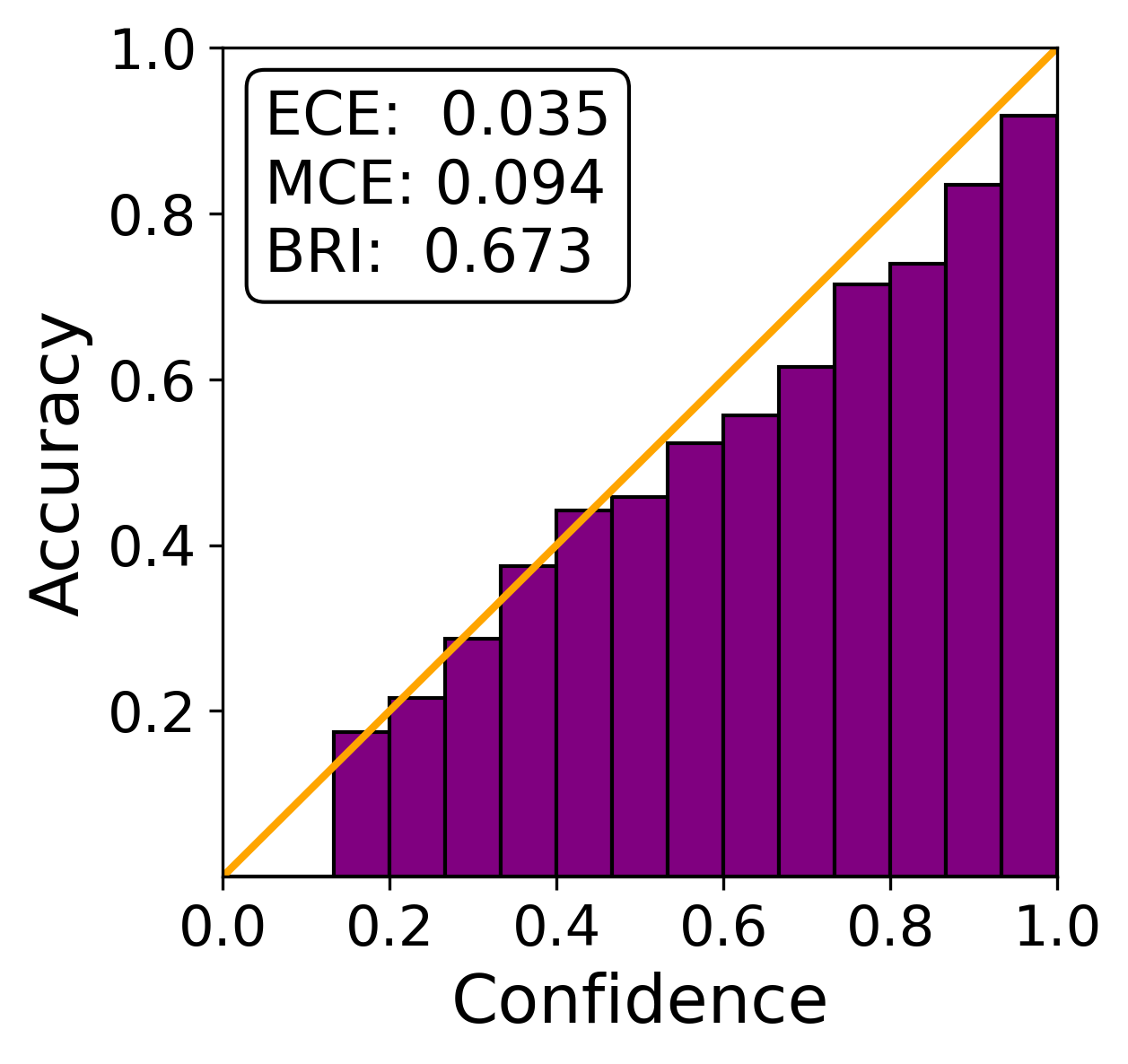} &
    \includegraphics[width=20mm]{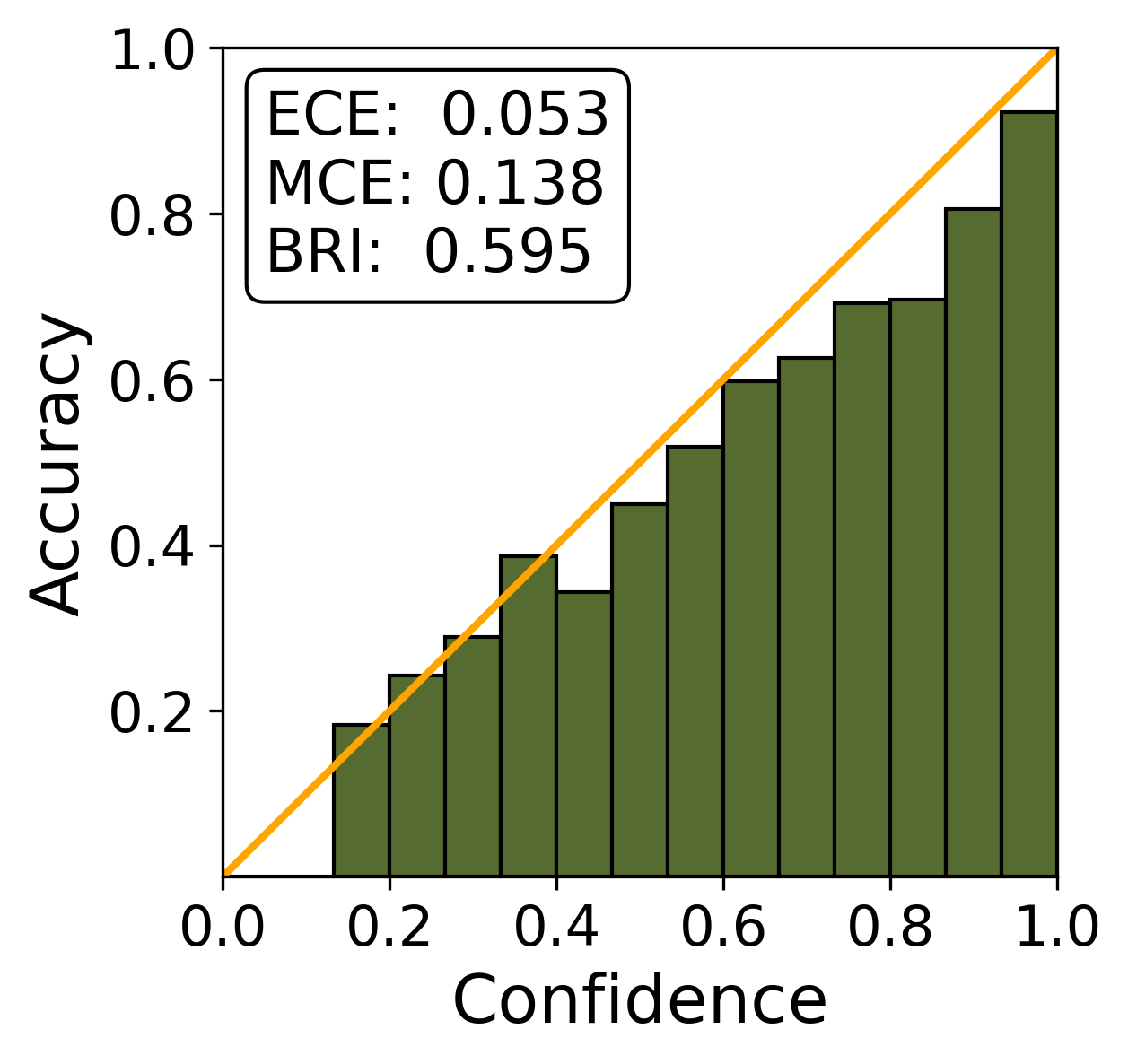} &
    \includegraphics[width=20mm]{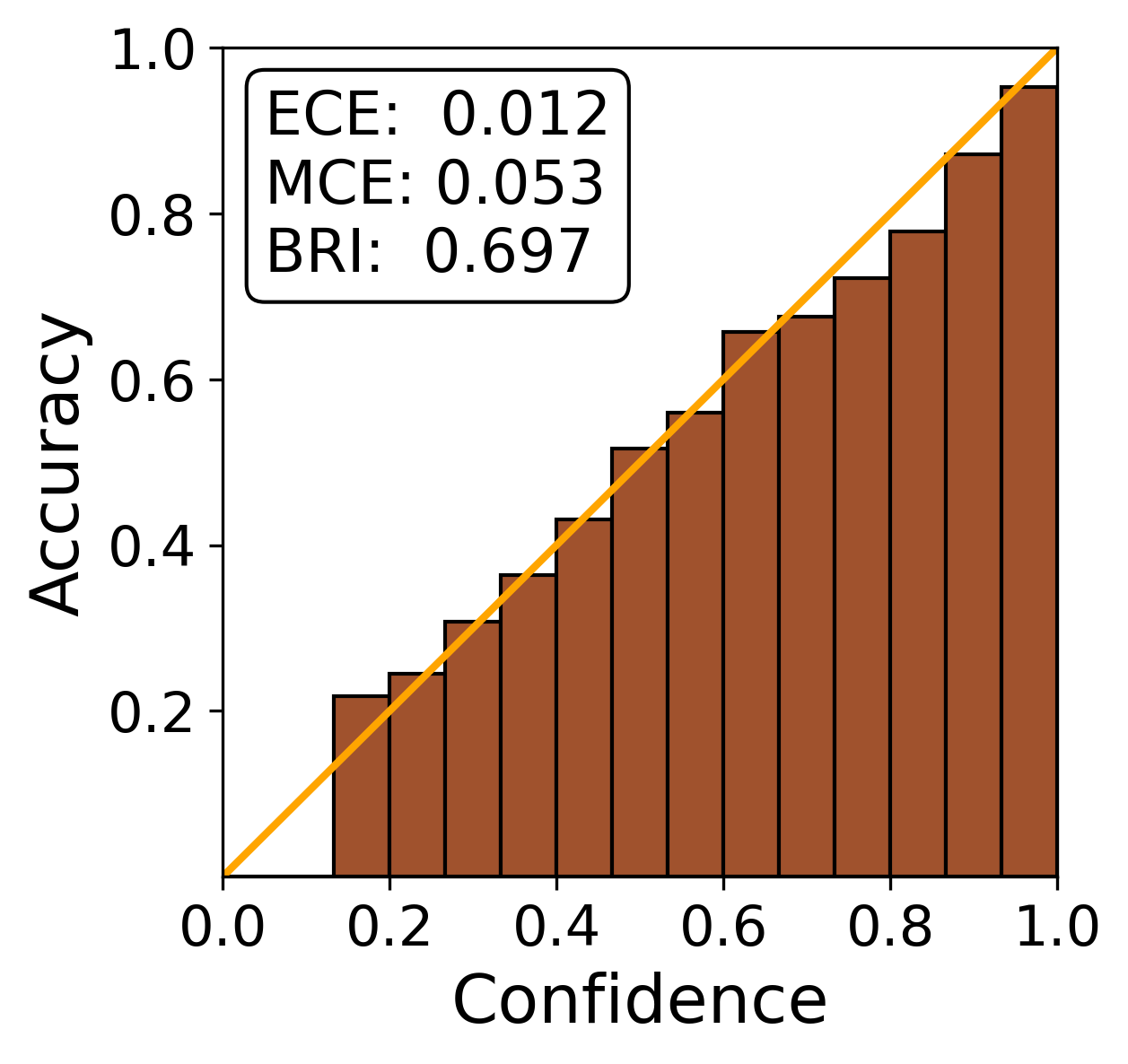} &
    \includegraphics[width=20mm]{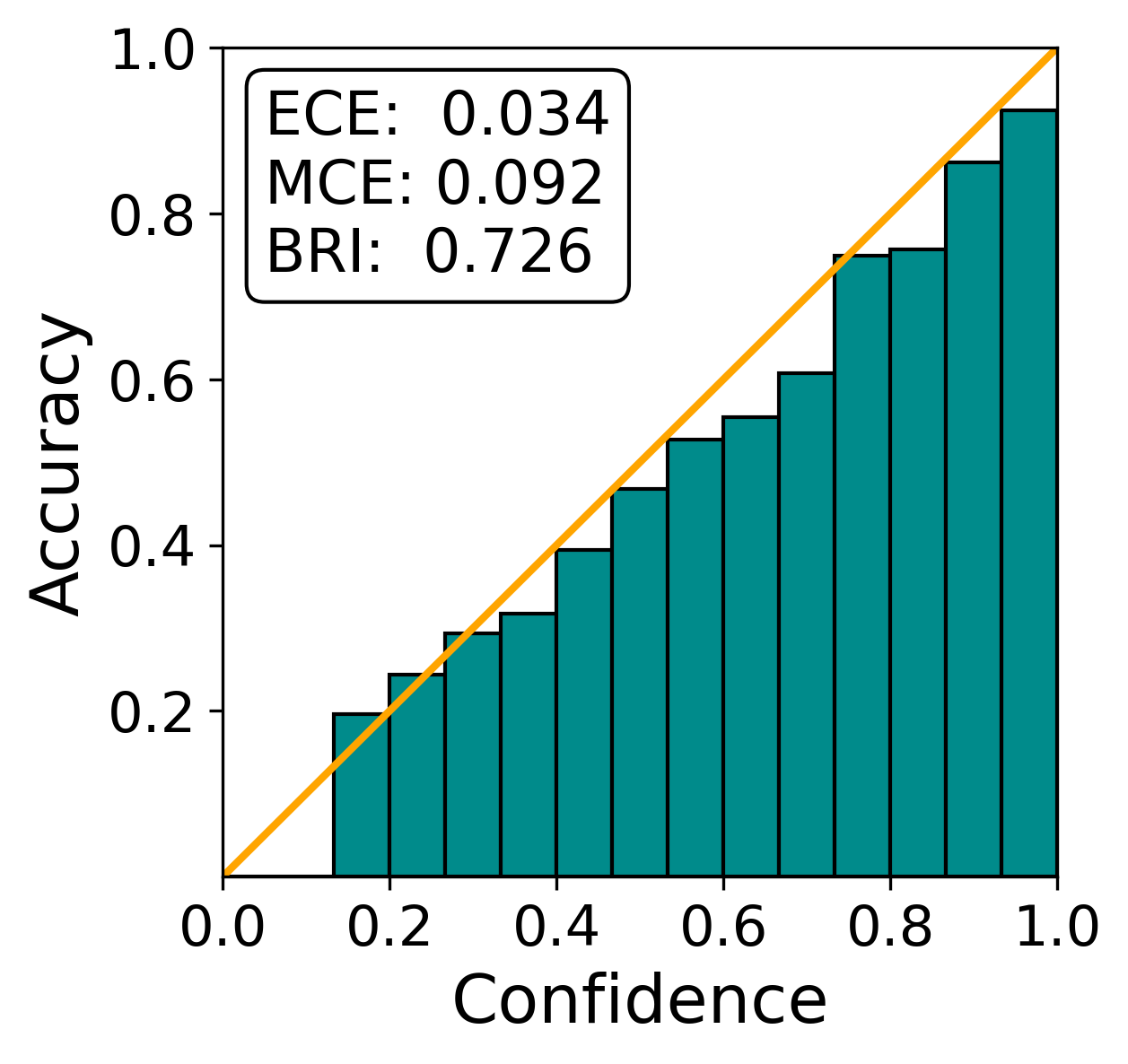}
    \\
     ~~~~pFedHN & ~~~~pFedGP-IP-data & ~~~~pFedGP-IP-compute-pred. & ~~~~~pFedGP-IP-compute-marg. & ~~~~pFedGP-pred. & ~~~~pFedGP-marg.
     \\
    \includegraphics[width=20mm]{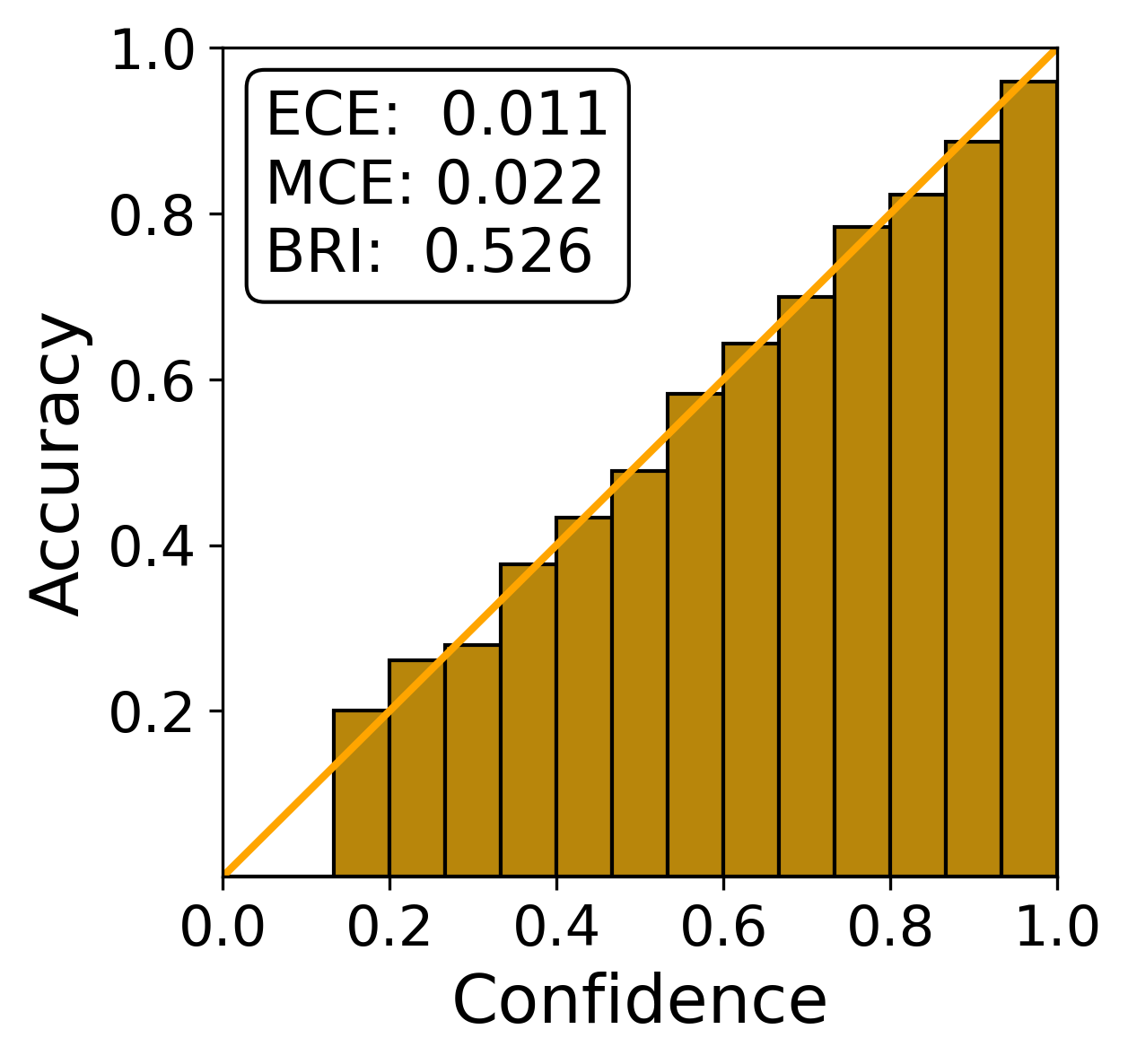} &
    \includegraphics[width=20mm]{figures/calibration/calibration50bestTemp/pFed-GP-IP-data_50clients.png} &
    \includegraphics[width=20mm]{figures/calibration/calibration50bestTemp/pFed-GP-IP-compute_50clients_predictive.png} &
    \includegraphics[width=20mm]{figures/calibration/calibration50bestTemp/pFed-GP-IP-compute_50clients_marginal.png} &
    \includegraphics[width=20mm]{figures/calibration/calibration50bestTemp/pFed-GP_50clients_predictive.png} &
    \includegraphics[width=20mm]{figures/calibration/calibration50bestTemp/pFed-GP_50clients_marginal.png}
    \end{tabular}}
    \caption{Reliability diagrams on CIFAR-100 with 50 clients. Best temperature. The last 5 figures are ours.}
    \label{fig:calibration_50_best_temp}
\end{figure}

\begin{figure}[!t]
\centering
\tiny
\scalebox{1.0}{
\begin{tabular}{c c c c c c}
    ~~~~FedAvg & ~~~~~~FOLA & ~~~~FedPer & ~~~~LG-FedAvg & ~~~~pFedMe & ~~~~~~FedU\\
    \includegraphics[width=20mm]{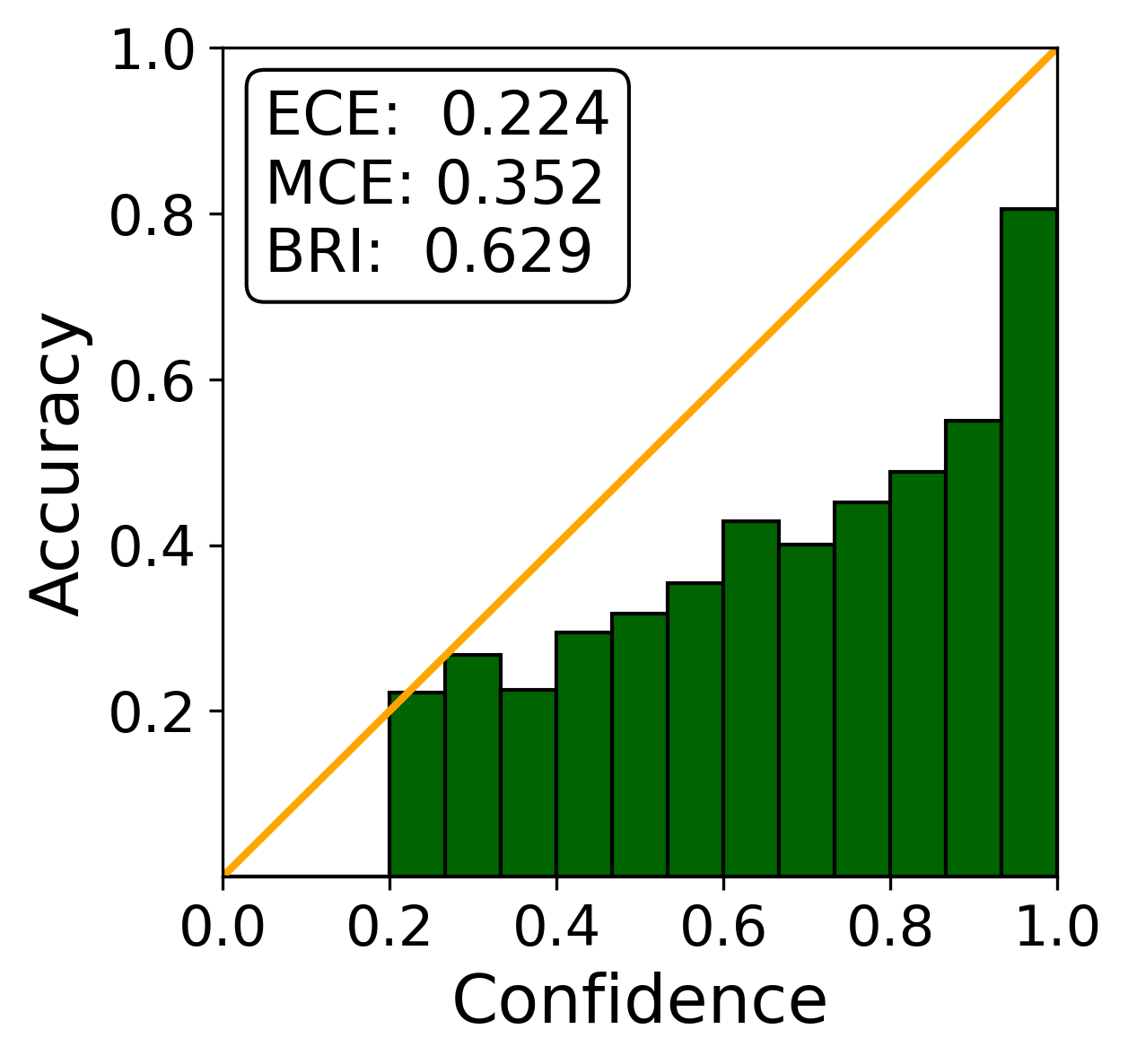} &   \includegraphics[width=20mm]{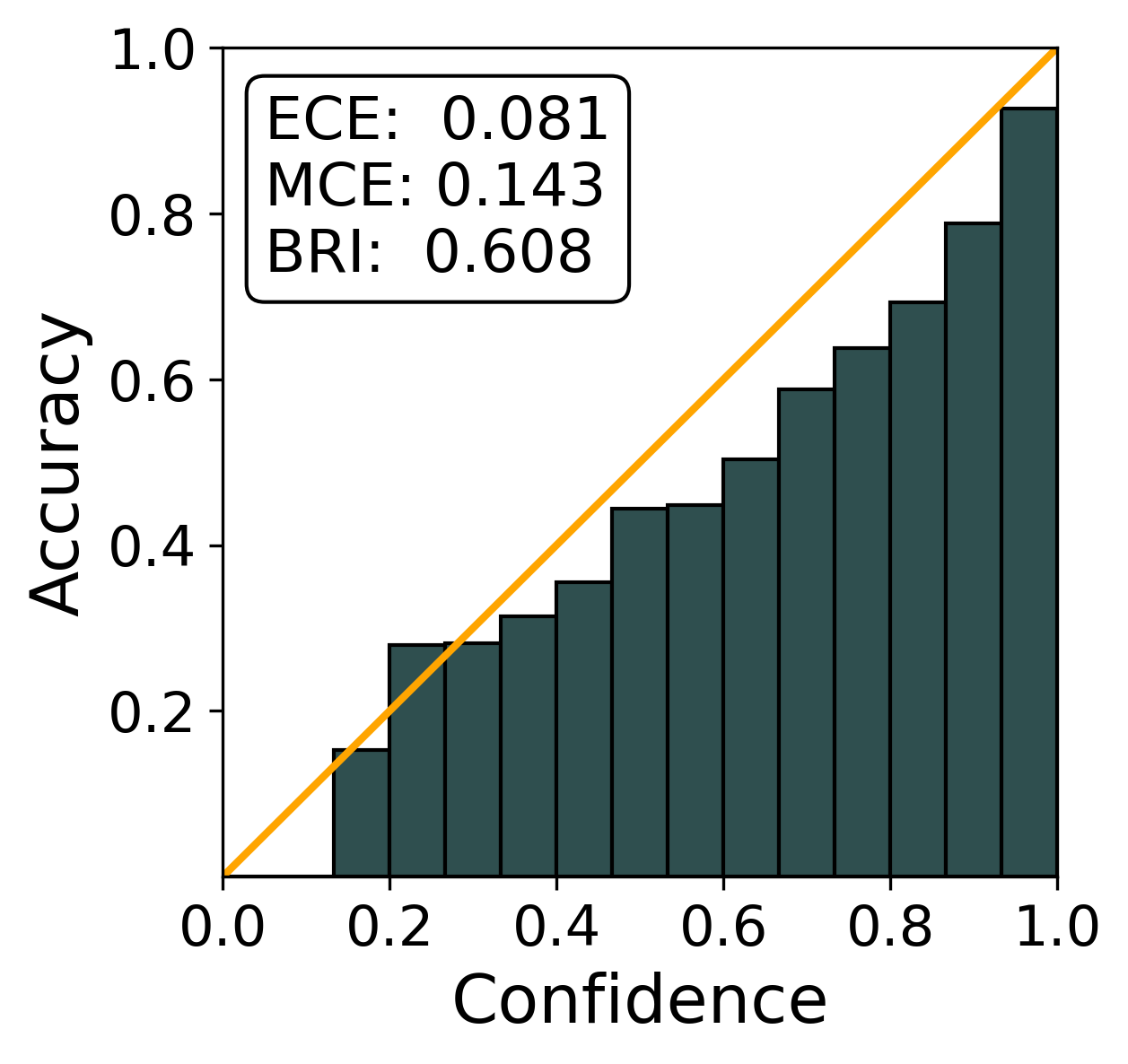} & \includegraphics[width=20mm]{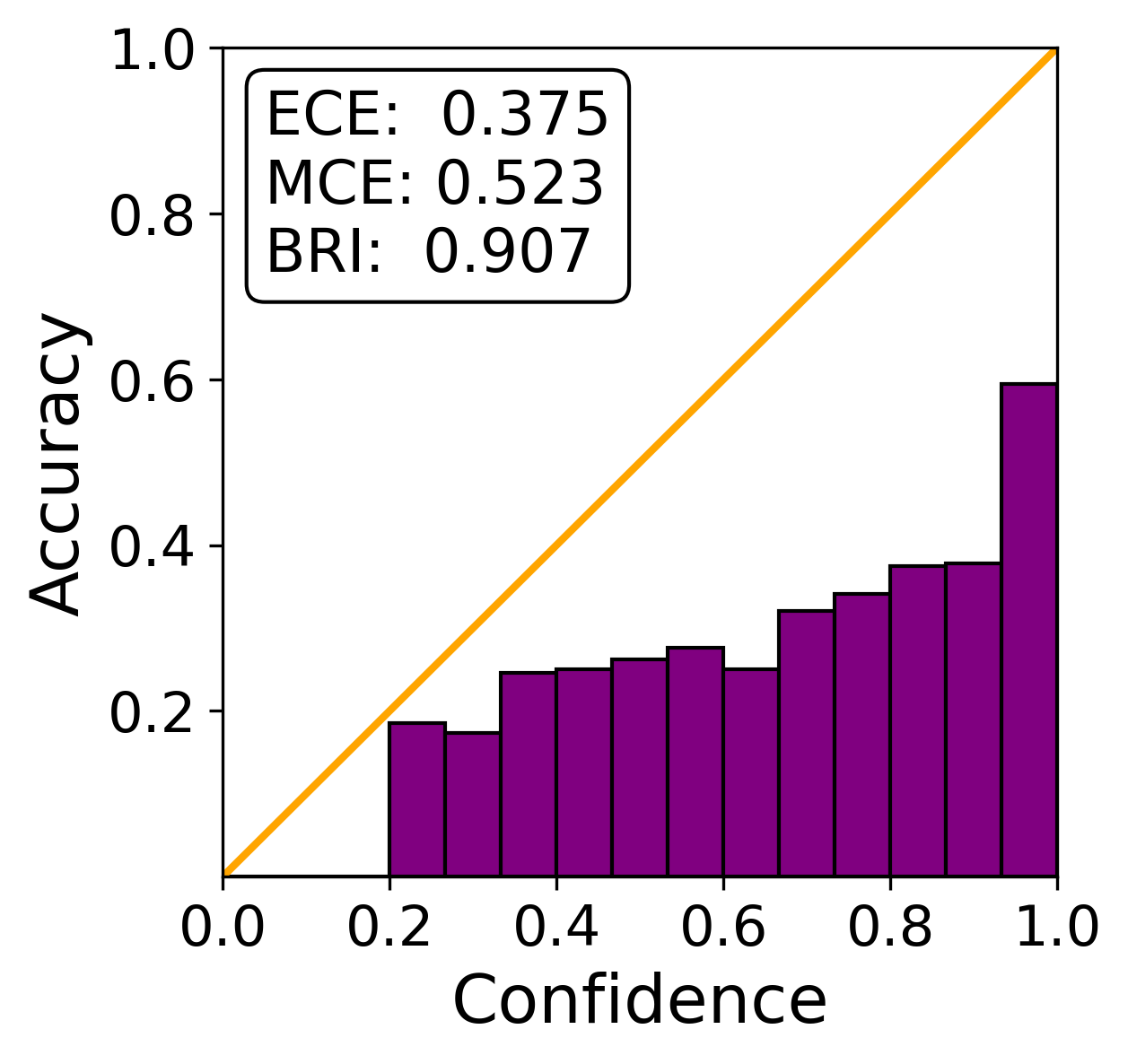} &
    \includegraphics[width=20mm]{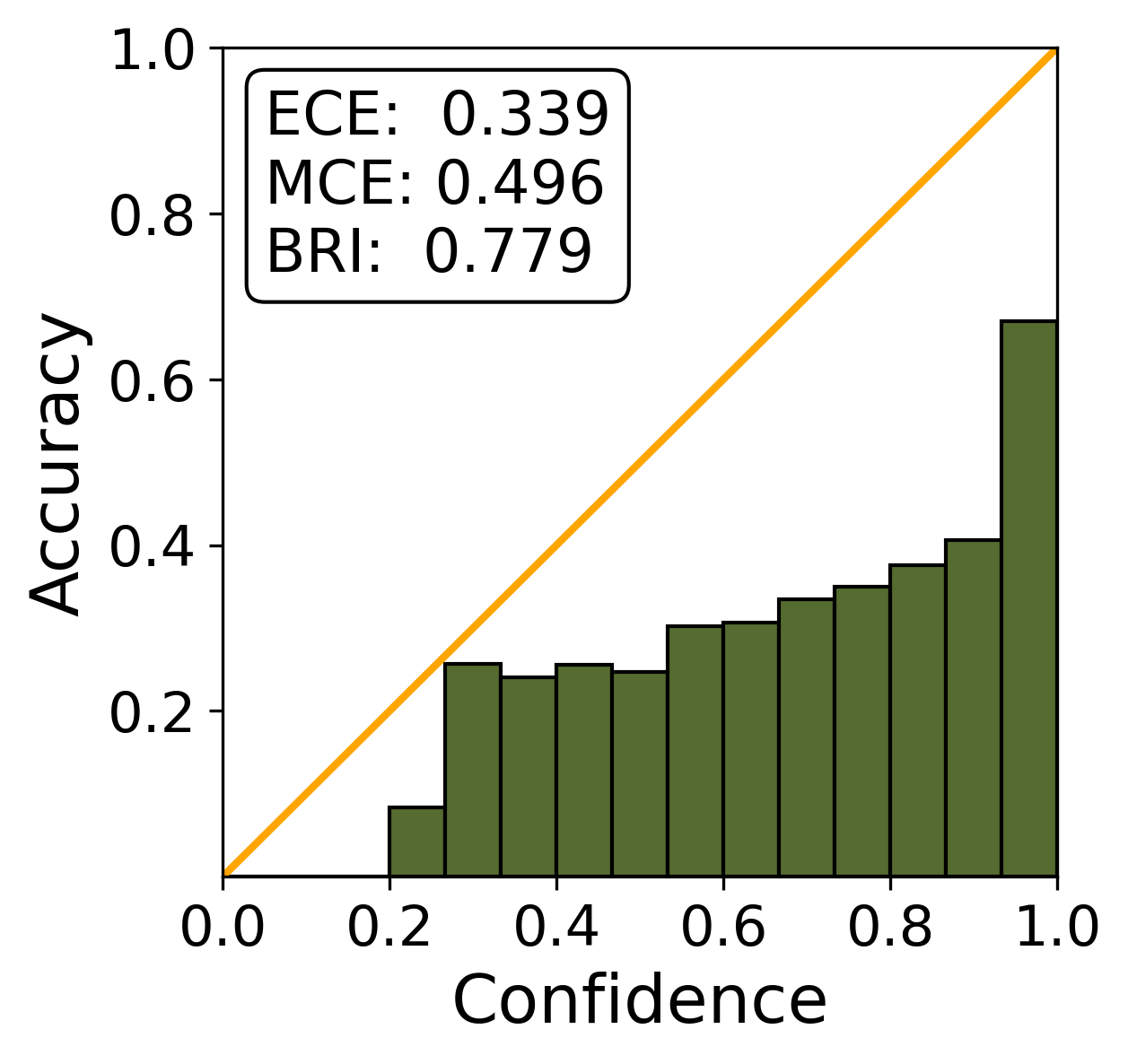} &
    \includegraphics[width=20mm]{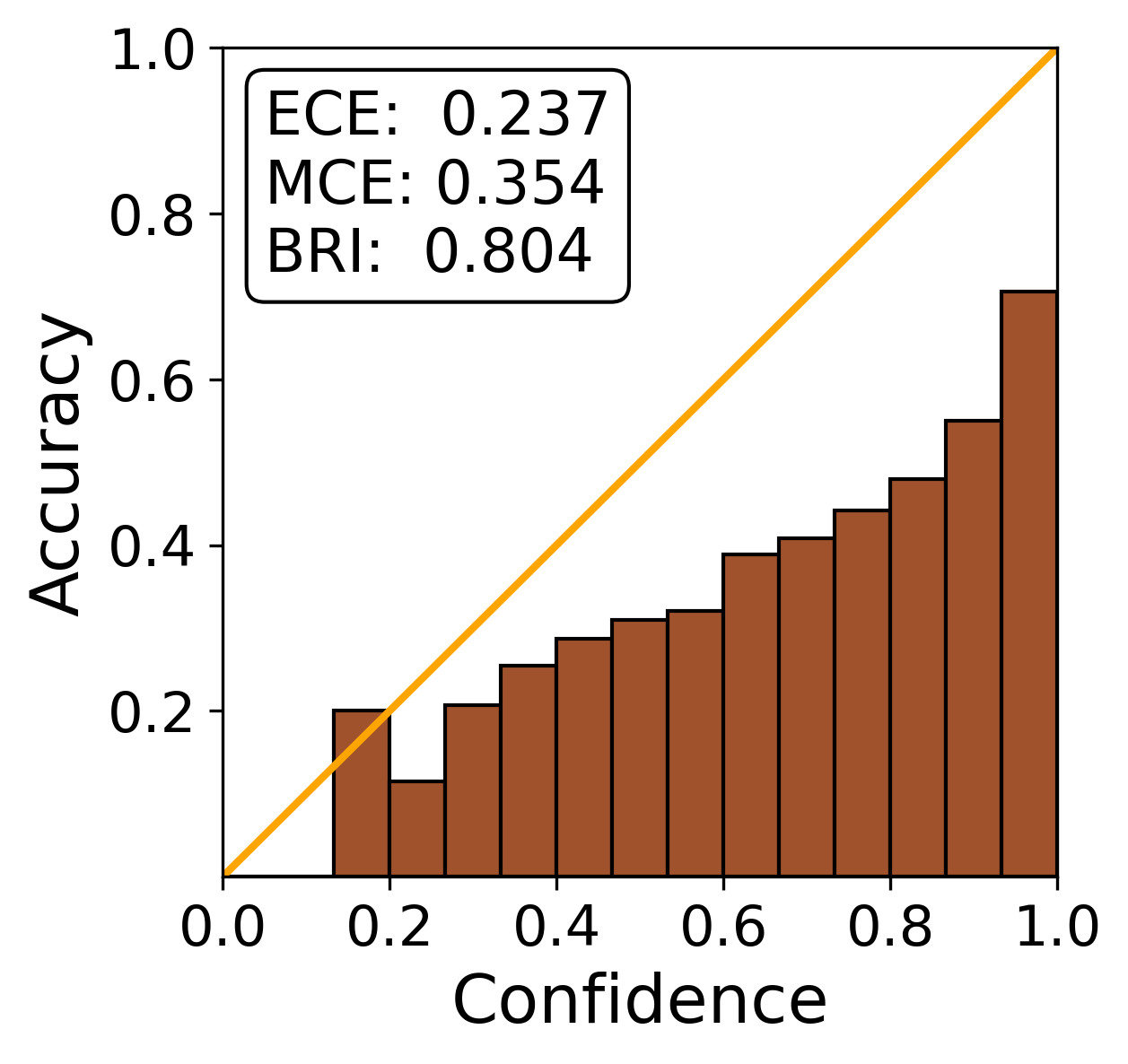} &
    \includegraphics[width=20mm]{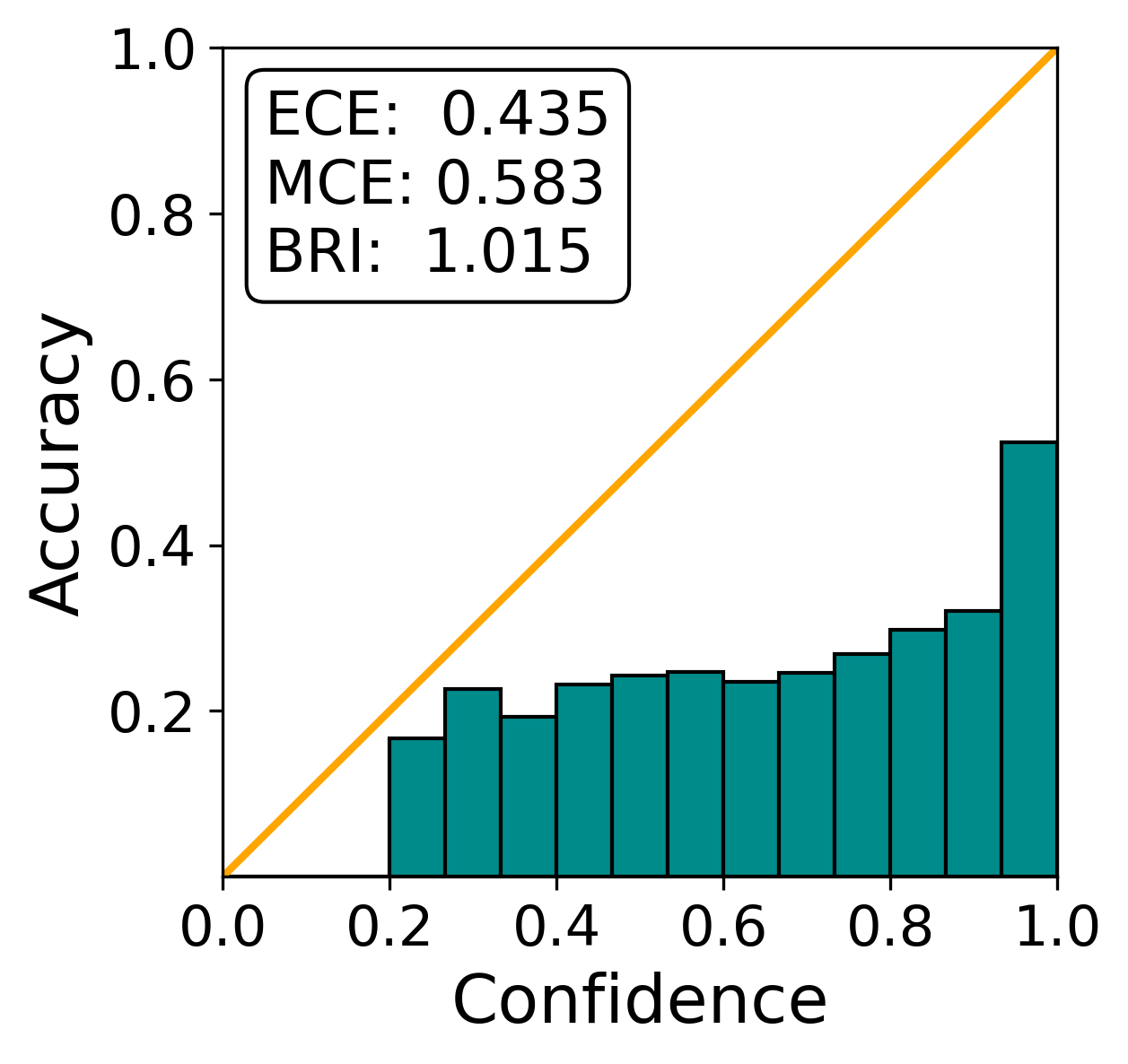}
    \\
     ~~~~pFedHN & ~~~~pFedGP-IP-data & ~~~~pFedGP-IP-compute-pred. & ~~~~~pFedGP-IP-compute-marg. & ~~~~pFedGP-pred. & ~~~~pFedGP-marg.
     \\
    \includegraphics[width=20mm]{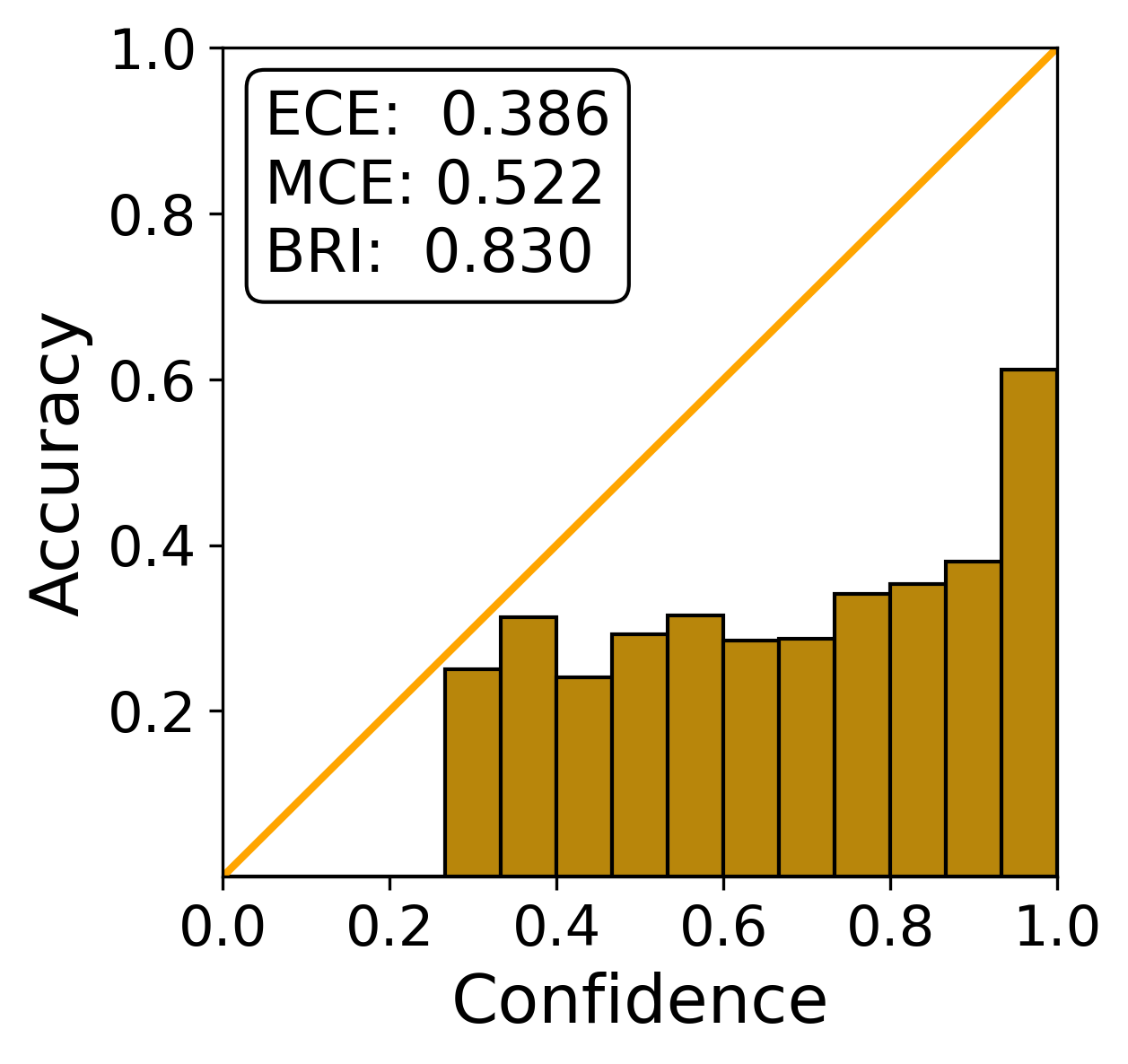} &
    \includegraphics[width=20mm]{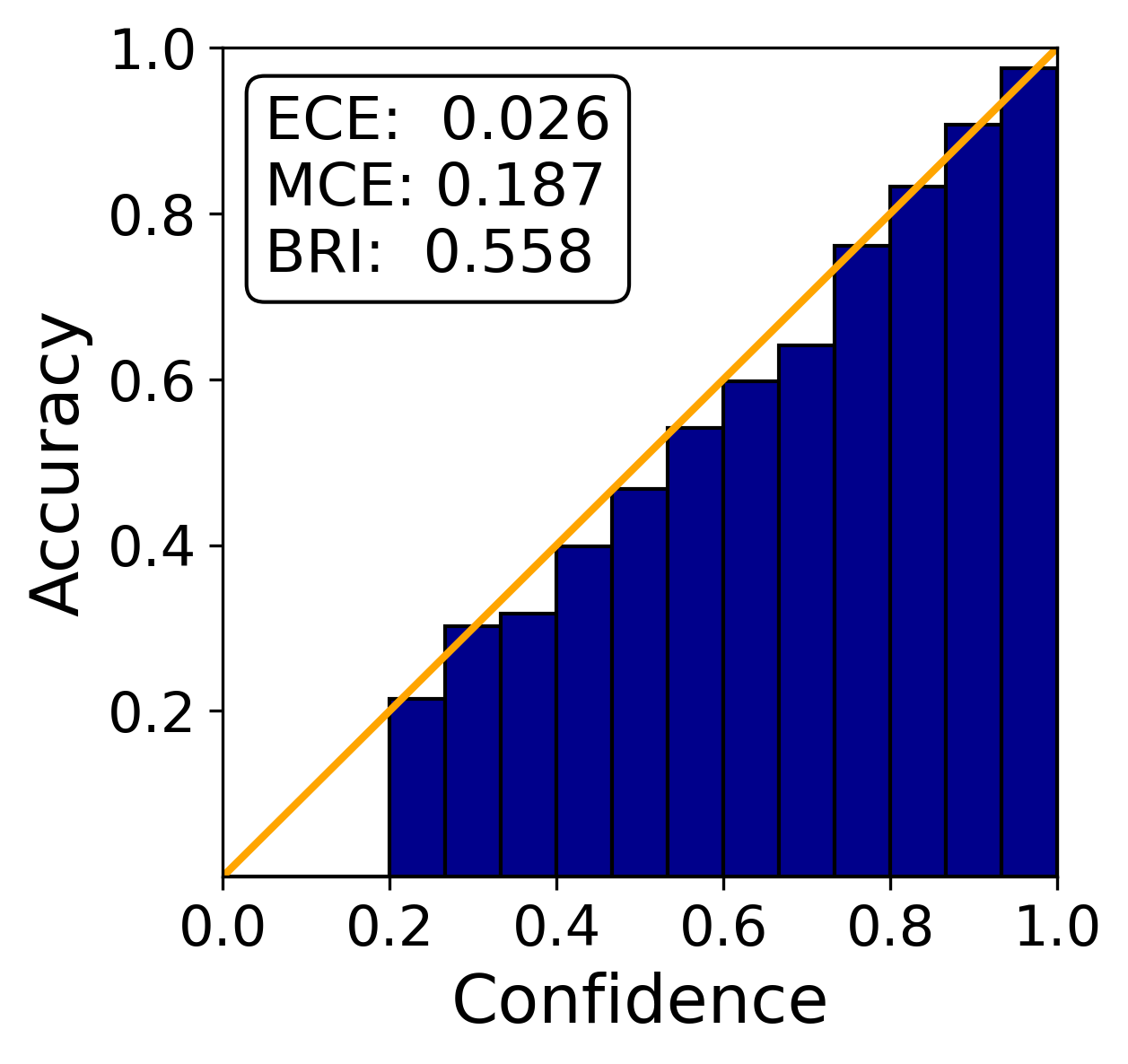} &
    \includegraphics[width=20mm]{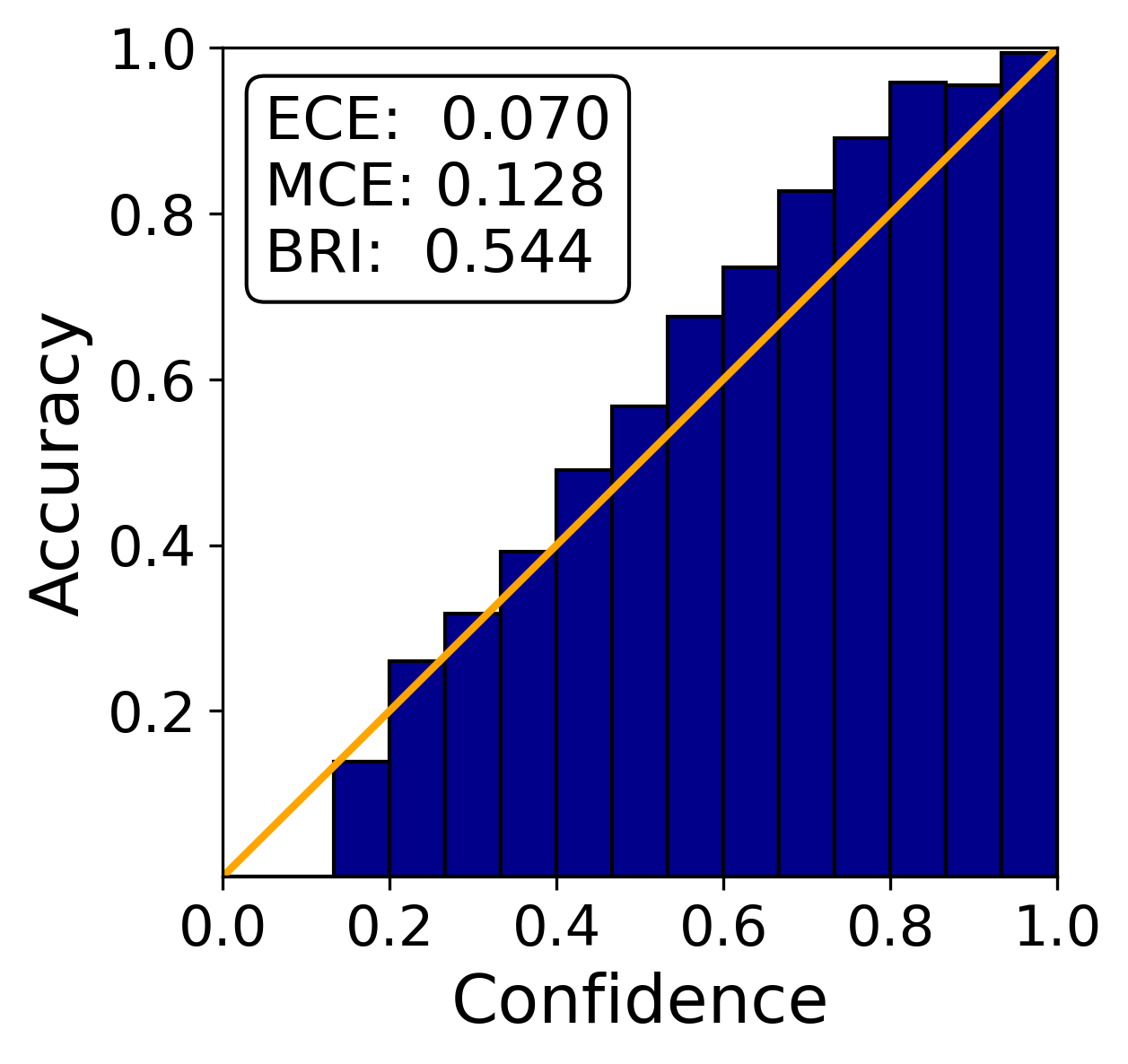} &
    \includegraphics[width=20mm]{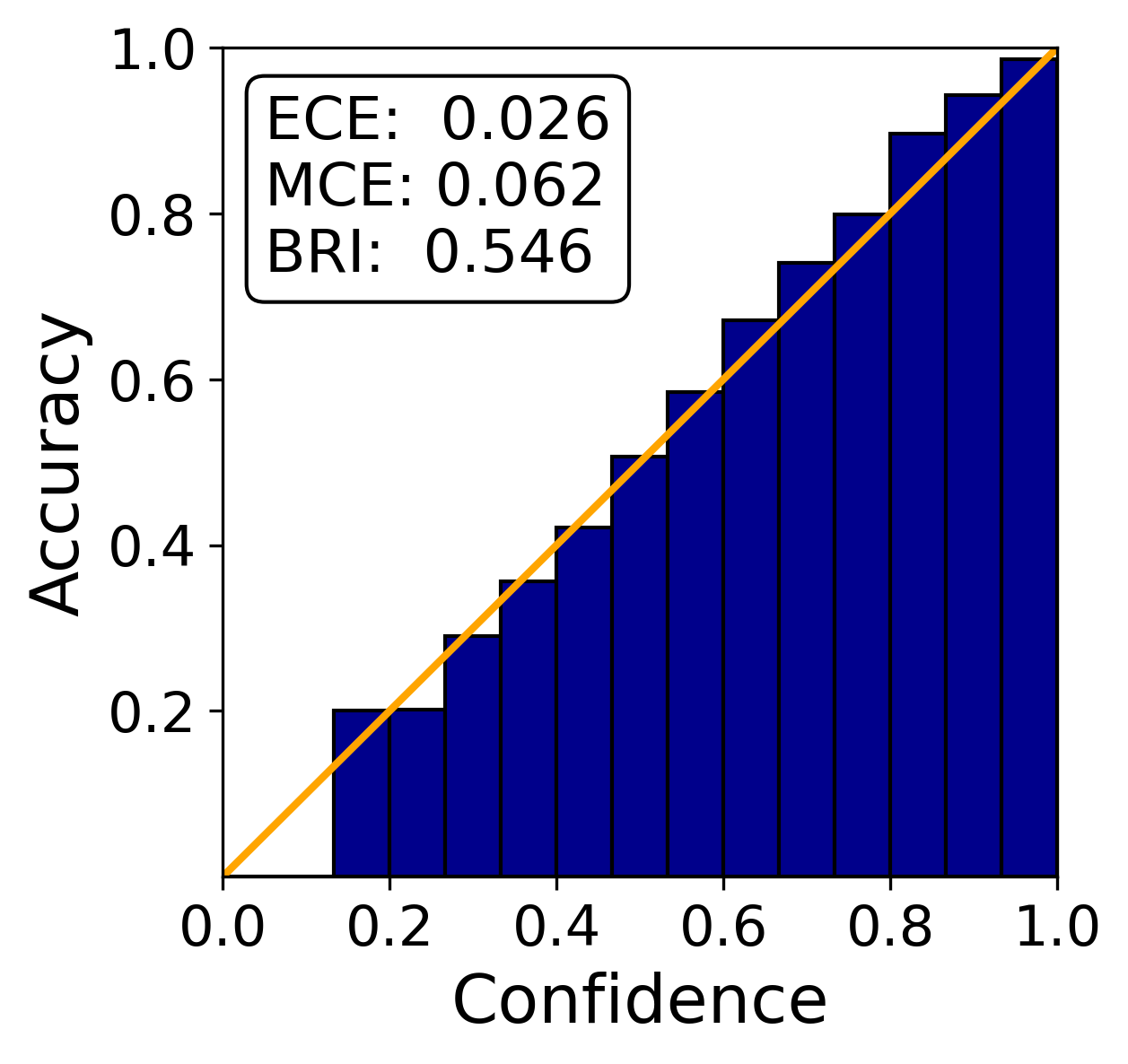} &
    \includegraphics[width=20mm]{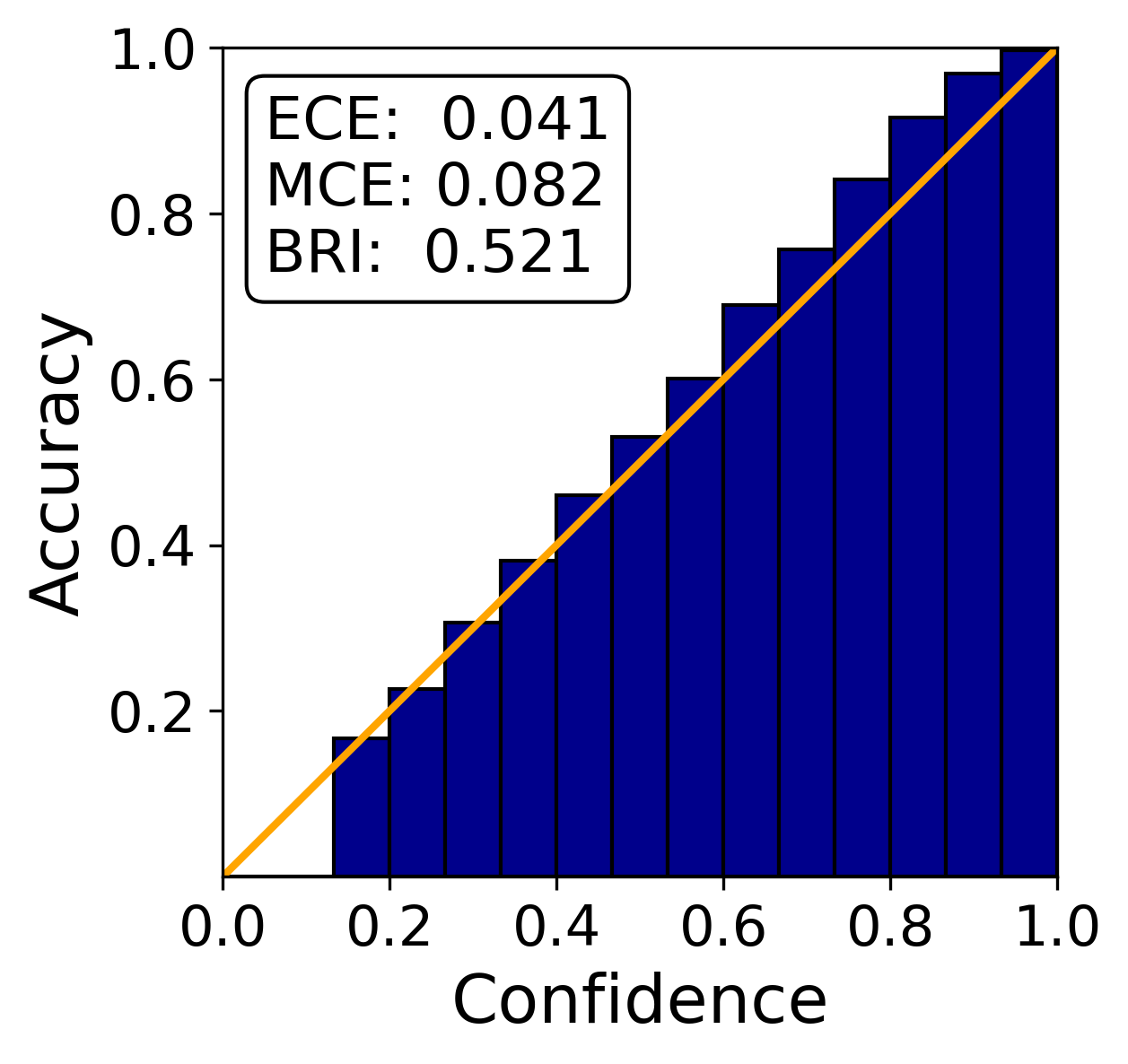} &
    \includegraphics[width=20mm]{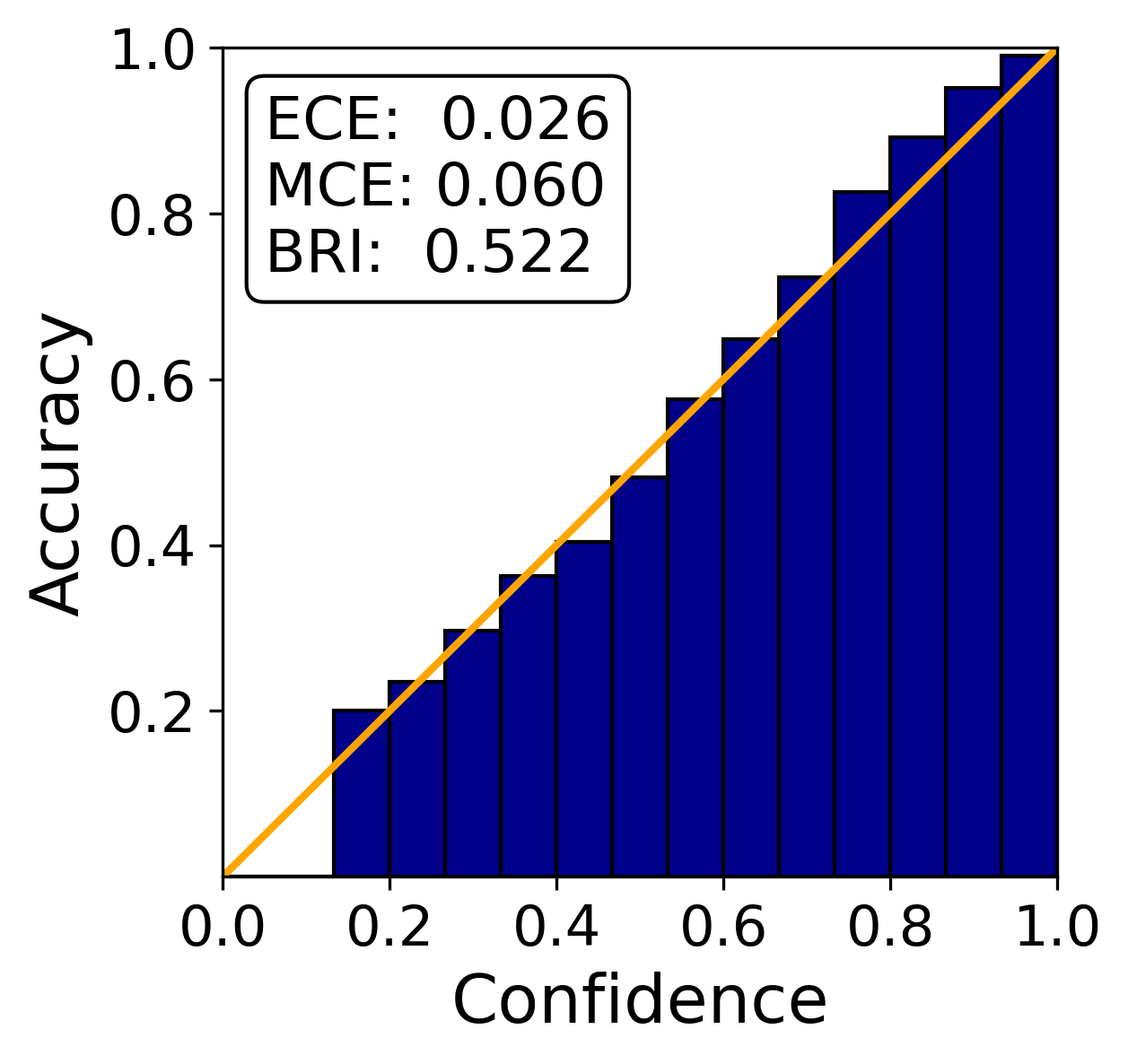}
    \end{tabular}}
    \caption{Reliability diagrams on CIFAR-100 with 100 clients. Default temperature ($t=1$). The last 5 figures are ours.}
    \label{fig:calibration_100_temp1}
\end{figure}

\begin{figure}[!t]
\centering
\tiny
\scalebox{1.0}{
\begin{tabular}{c c c c c c}
    ~~~~FedAvg & ~~~~~~FOLA & ~~~~FedPer & ~~~~LG-FedAvg & ~~~~pFedMe & ~~~~~~FedU\\
    \includegraphics[width=20mm]{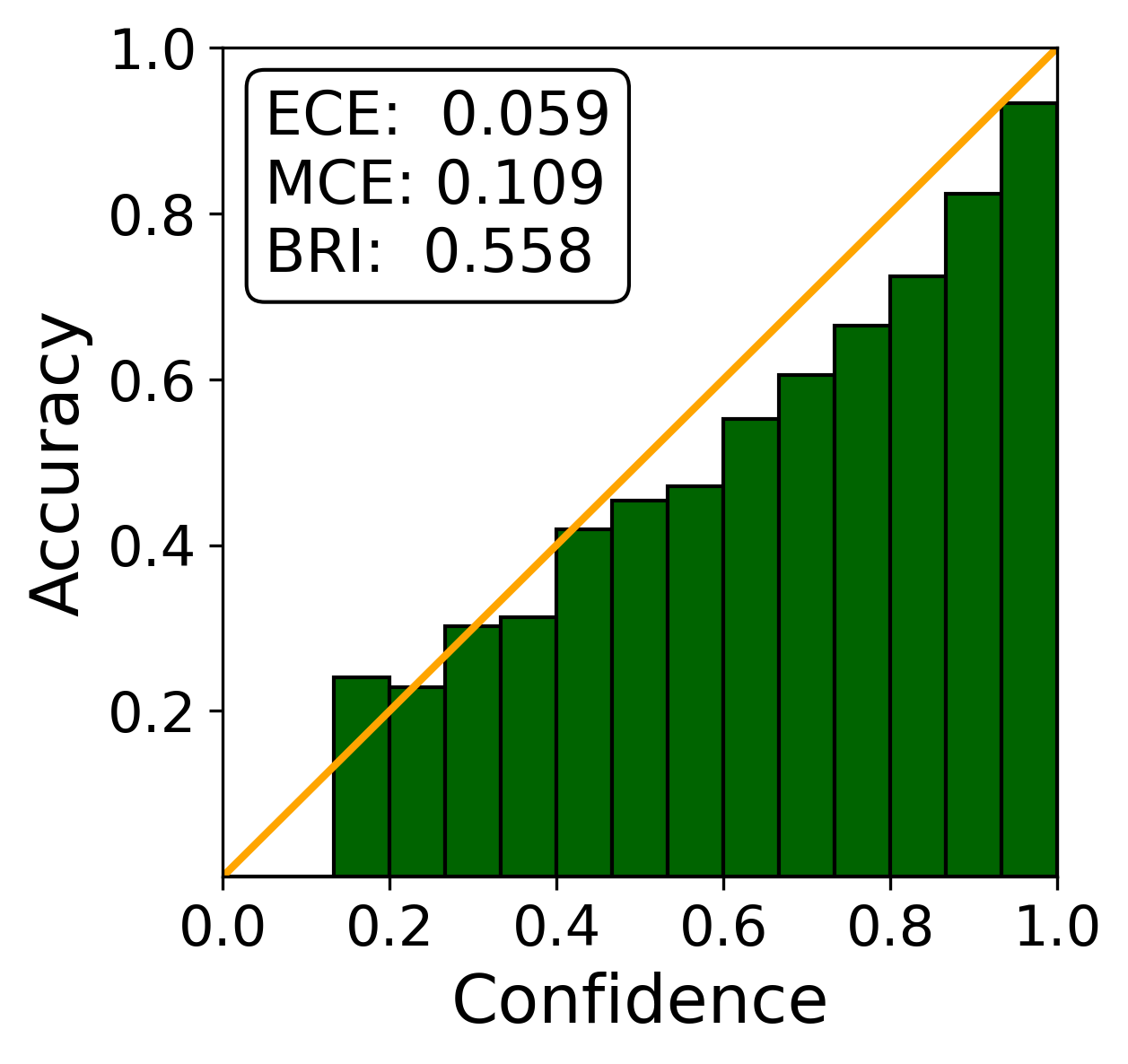} &  \includegraphics[width=20mm]{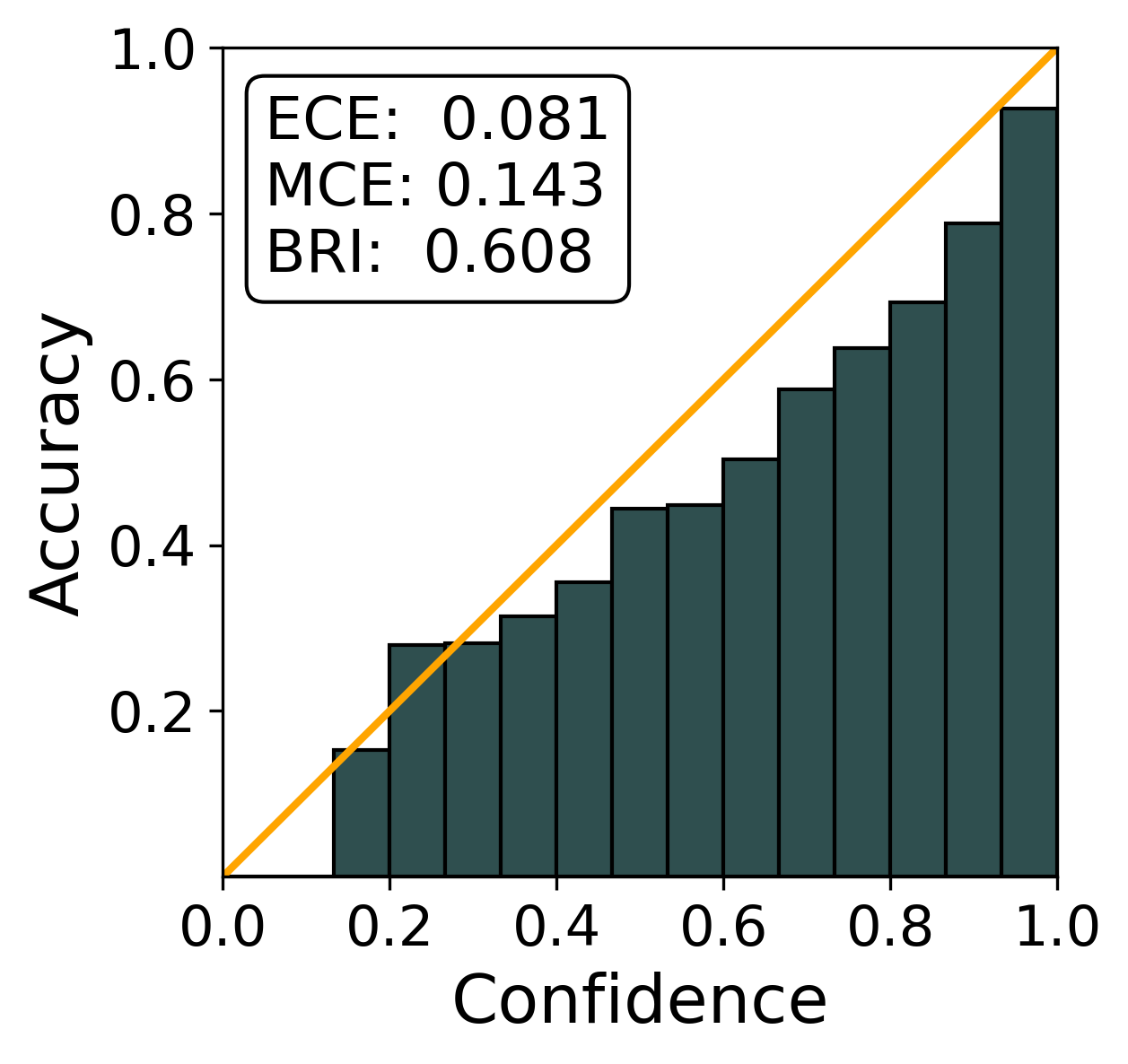} & \includegraphics[width=20mm]{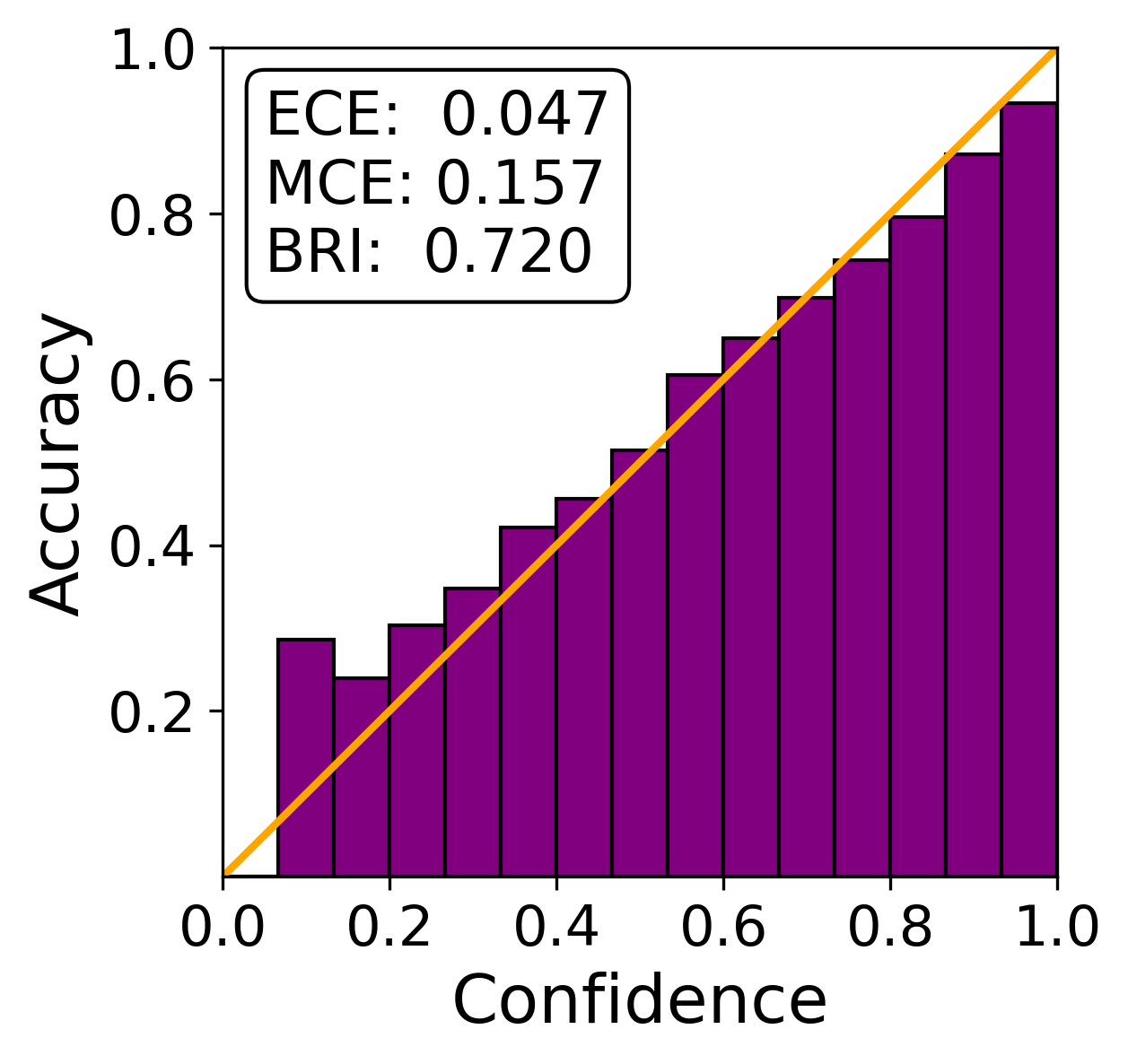} &
    \includegraphics[width=20mm]{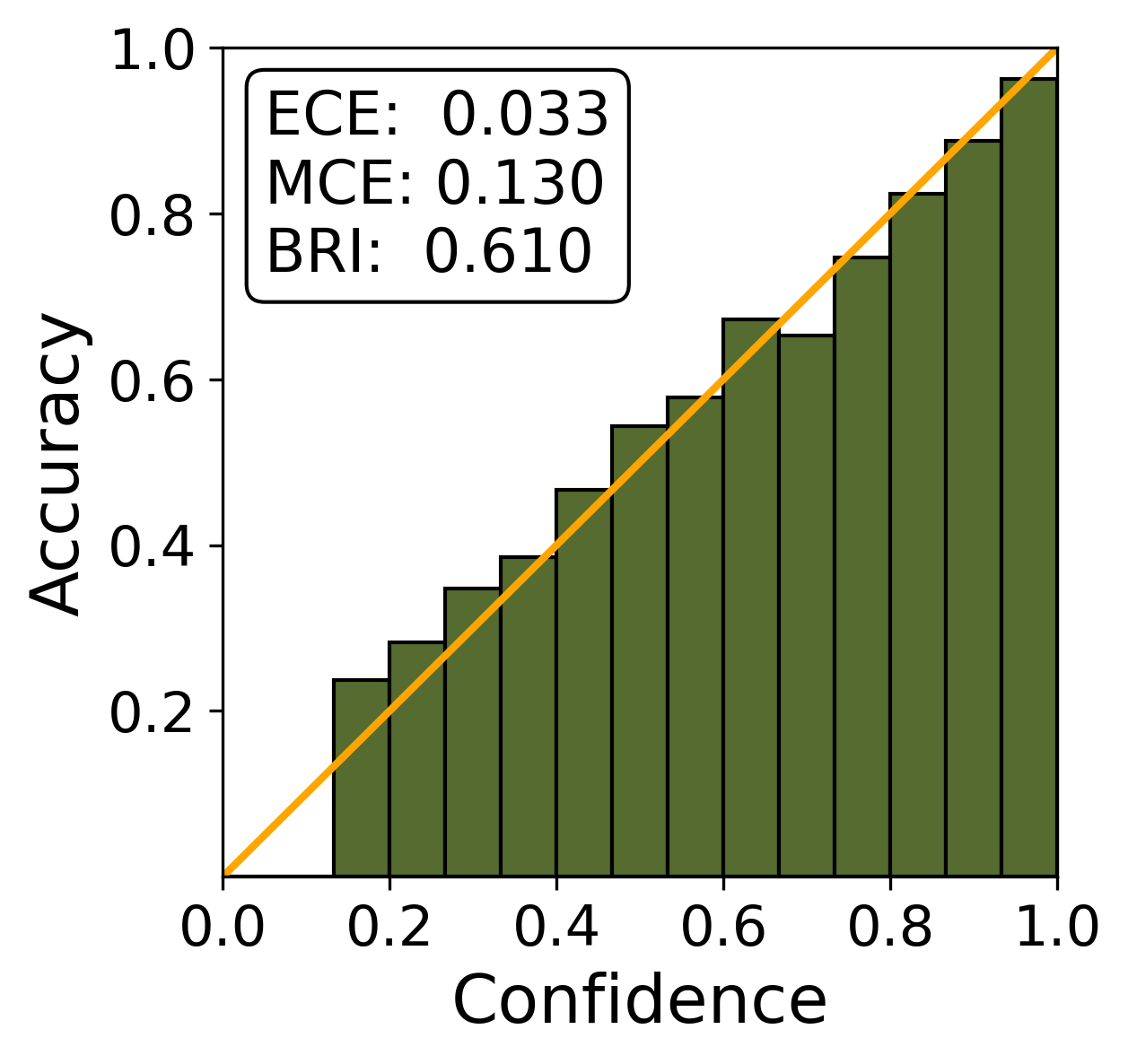} &
    \includegraphics[width=20mm]{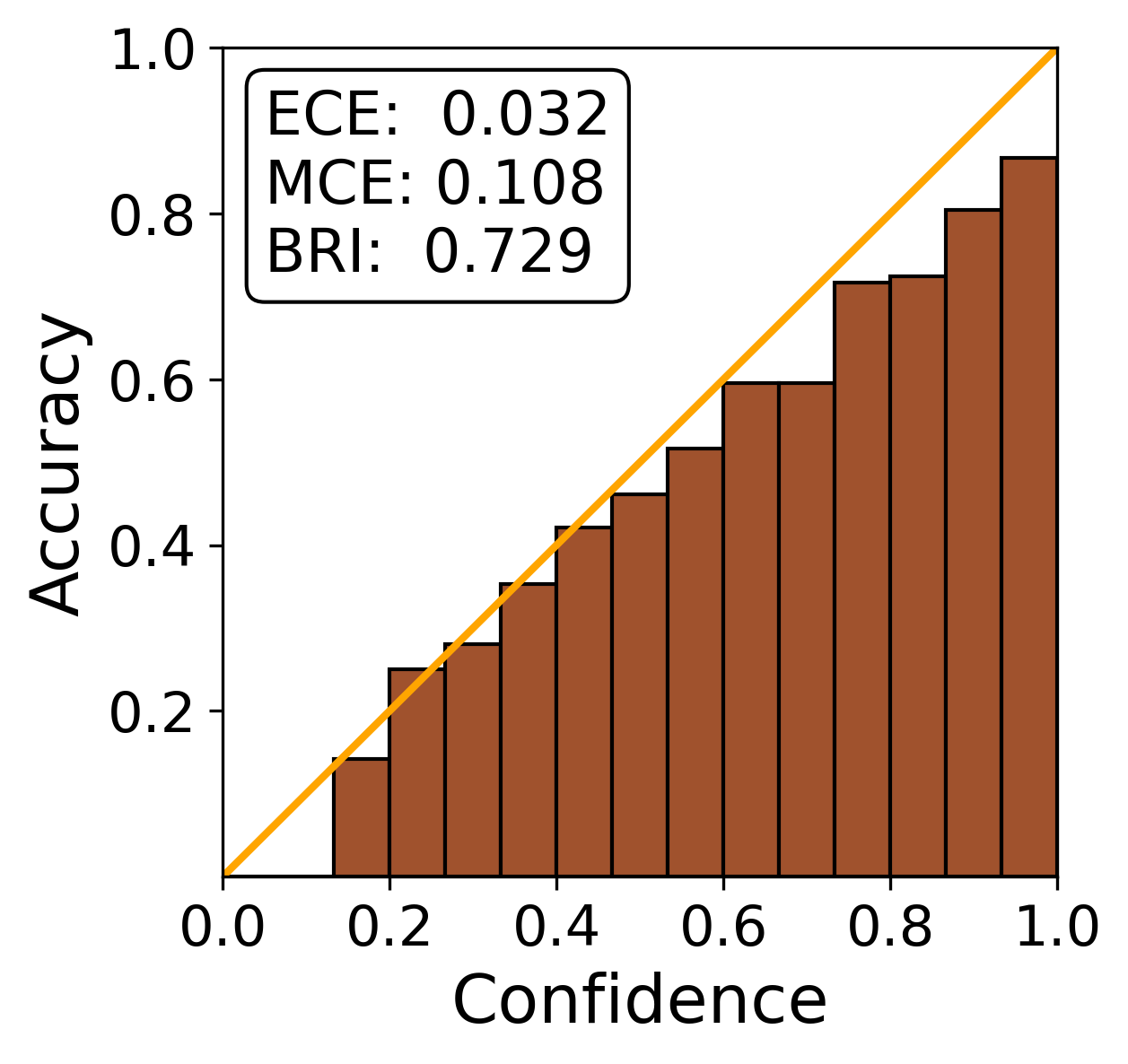} &
    \includegraphics[width=20mm]{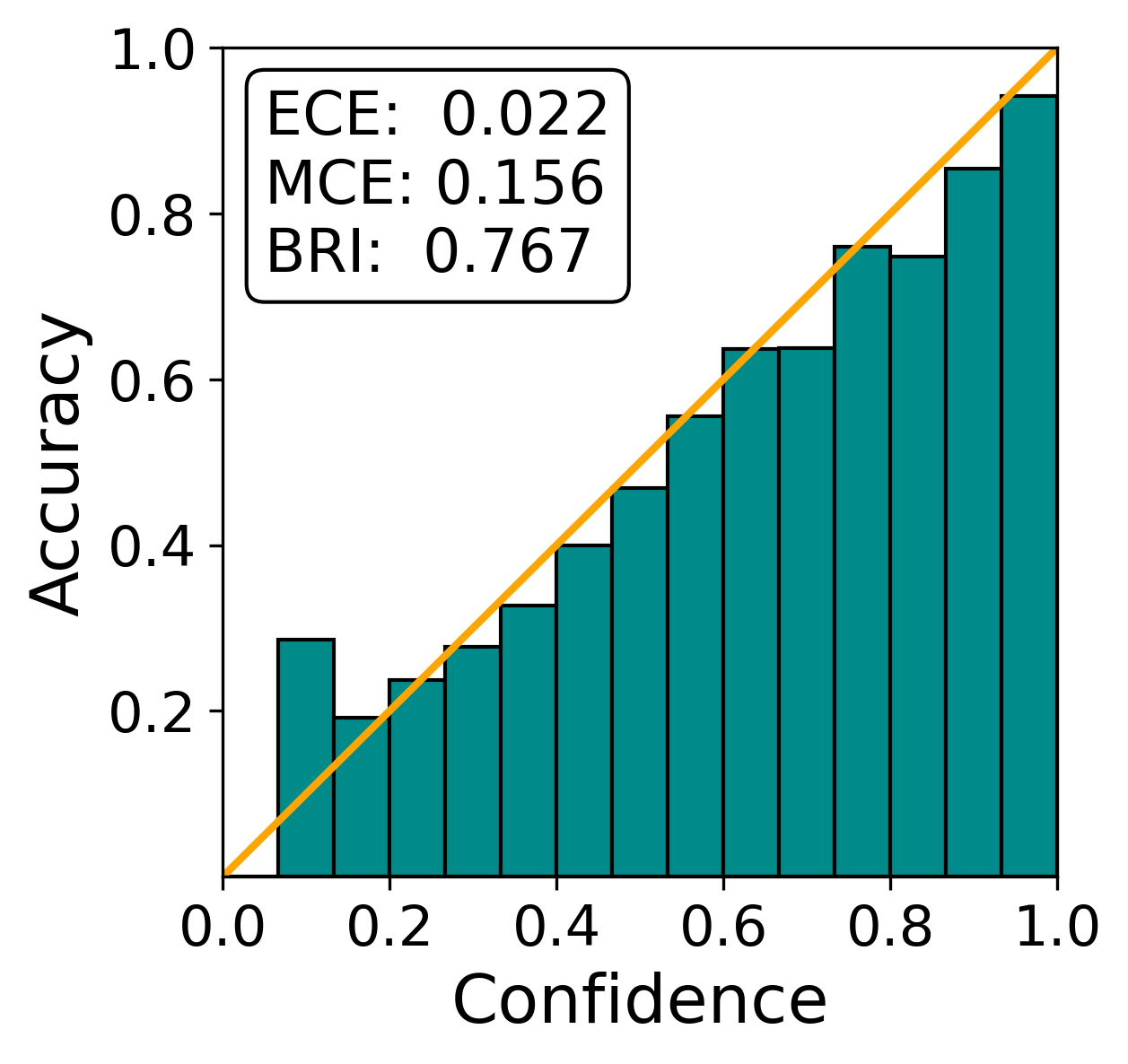}
    \\
     ~~~~pFedHN & ~~~~pFedGP-IP-data & ~~~~pFedGP-IP-compute-pred. & ~~~~~pFedGP-IP-compute-marg. & ~~~~pFedGP-pred. & ~~~~pFedGP-marg.
     \\
    \includegraphics[width=20mm]{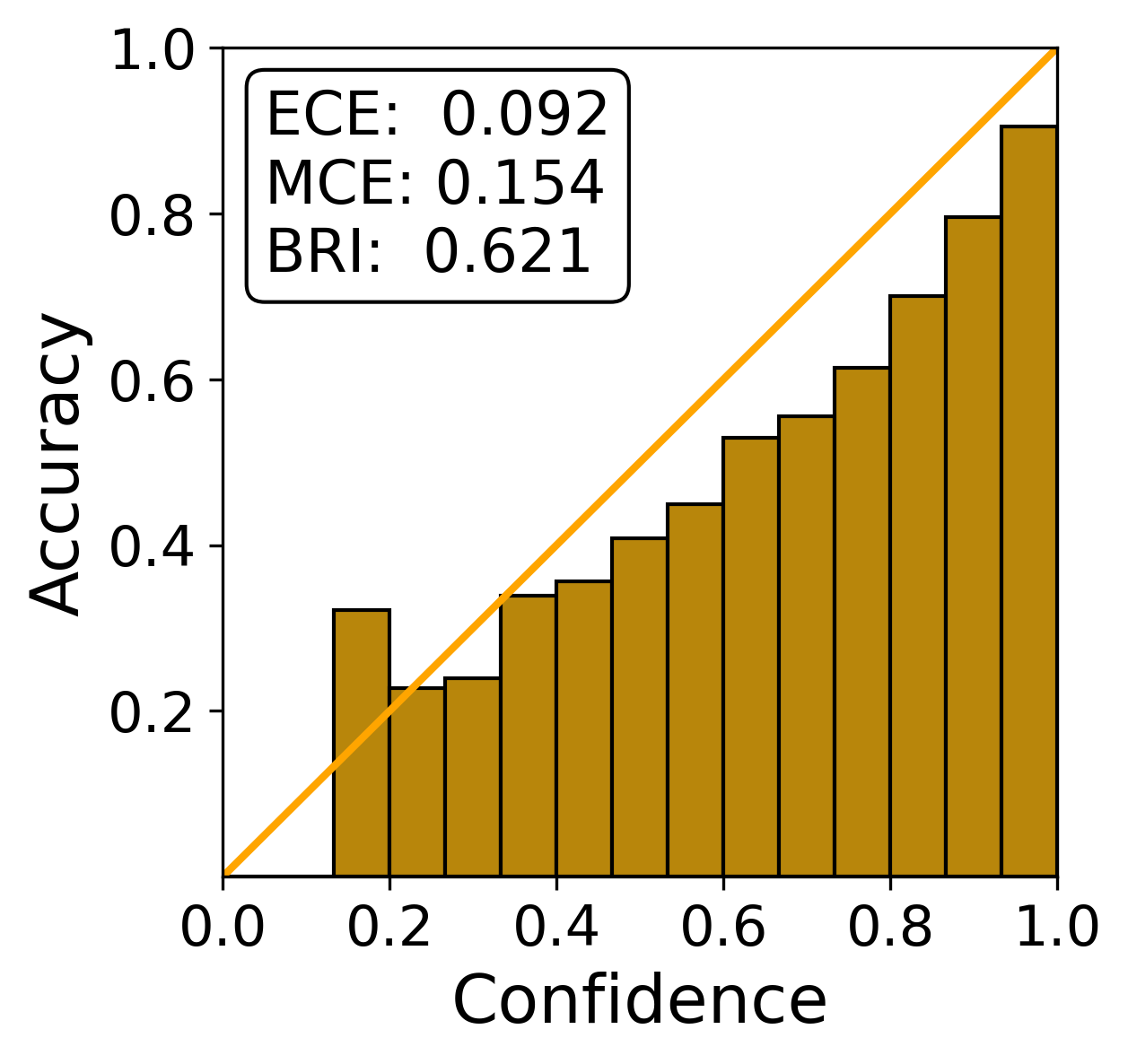} &
    \includegraphics[width=20mm]{figures/calibration/calibration100BestTemp/pFed-GP-IP-data_100clients.png} &
    \includegraphics[width=20mm]{figures/calibration/calibration100BestTemp/pFed-GP-IP-compute_100clients_predictive.png} &
    \includegraphics[width=20mm]{figures/calibration/calibration100BestTemp/pFed-GP-IP-compute_100clients_marginal.png} &
    \includegraphics[width=20mm]{figures/calibration/calibration100BestTemp/pFed-GP_100clients_predictive.png} &
    \includegraphics[width=20mm]{figures/calibration/calibration100BestTemp/pFed-GP_100clients_marginal.png}
    \end{tabular}}
    \caption{Reliability diagrams on CIFAR-100 with 100 clients. Best temperature. The last 5 figures are ours.}
    \label{fig:calibration_100_best_temp}
\end{figure}

Now we present additional reliability diagrams for CIFAR-100 with 50 and 100 clients, with and without temperature scaling (See \Figref{fig:calibration_all} for unified diagrams and Figures \ref{fig:calibration_50_temp1_appendix} - \ref{fig:calibration_100_best_temp} for separate diagrams). For each baseline we applied a grid search over a temperature $t \in \{0.1, 0.2, 0.5, 1.0, 2.0, 5.0, 10.0, 20.0, 50.0, 100.0, 200.0, 500.0, 1000.0\}$, chose the best temperature based on the pre-allocated validation set according to the ECE, and generated the diagram using the test set data. In addition, we present here reliability diagrams obtained by optimizing the marginal likelihood for pFedGP and pFedGP-IP-compute. From the figures, pFedGP does not gain from temperature scaling as baseline methods do since it is a calibrated classifier by design. Although this procedure improves the calibration of baseline methods, we note that finding the right temperature requires having a separate validation set which often can be challenging to obtain for problems in the low data regime.

\clearpage

\end{document}